\documentclass[journal]{IEEEtran}
\ifCLASSINFOpdf
\else
\fi

\usepackage{times}
\usepackage{epsfig}
\usepackage{graphicx}
\usepackage{amsmath}
\usepackage{amssymb}
\usepackage{array}
\usepackage{subfigure}
\usepackage{multirow}
\usepackage{color}
\usepackage{bm}
\usepackage{upgreek}
\usepackage{verbatim}
\usepackage{color}

\begin{document}
%
\title{Fast and High Quality Highlight Removal\\ from A Single Image}
%
%
%

\author{Dongsheng An$^{*}$, Jinli Suo$^{*}$, Xiangyang Ji, Haoqian Wang and Qionghai Dai

\thanks{$*$ indicates equal contribution.}}

\maketitle

\begin{abstract}
Specular reflection exists widely in photography and causes the recorded color deviating from its true value, so fast and high quality highlight removal from a single nature image is of great importance. In spite of the progress in the past decades in highlight removal, achieving wide applicability to the large diversity of nature scenes is quite challenging. To handle this problem, we propose an analytic solution to highlight removal based on an $L_2$ chromaticity definition and corresponding dichromatic model. Specifically, this paper derives a normalized dichromatic model for the pixels with identical diffuse color: a unit circle equation of projection coefficients in two subspaces that are orthogonal to and parallel with the illumination, respectively. 
In the former illumination orthogonal subspace, which is specular-free, we can conduct robust clustering with an explicit criterion to determine the cluster number adaptively. In the latter illumination parallel subspace, a property called pure diffuse pixels distribution rule (PDDR) helps map each specular-influenced pixel to its diffuse component.
In terms of efficiency, the proposed approach involves few complex calculation, and thus can remove highlight from high resolution images fast. Experiments show that this method is of superior performance in various challenging cases.
\end{abstract}

\begin{IEEEkeywords}
Specular reflection, highlight removal, specular and diffuse, $L_2$ normalized dichromatic model, adaptive material clustering.
\end{IEEEkeywords}

%
\IEEEpeerreviewmaketitle

\section{Introduction}
\begin{figure}[t]
\centering
  \subfigure[]{
      \includegraphics[width=0.48\linewidth]{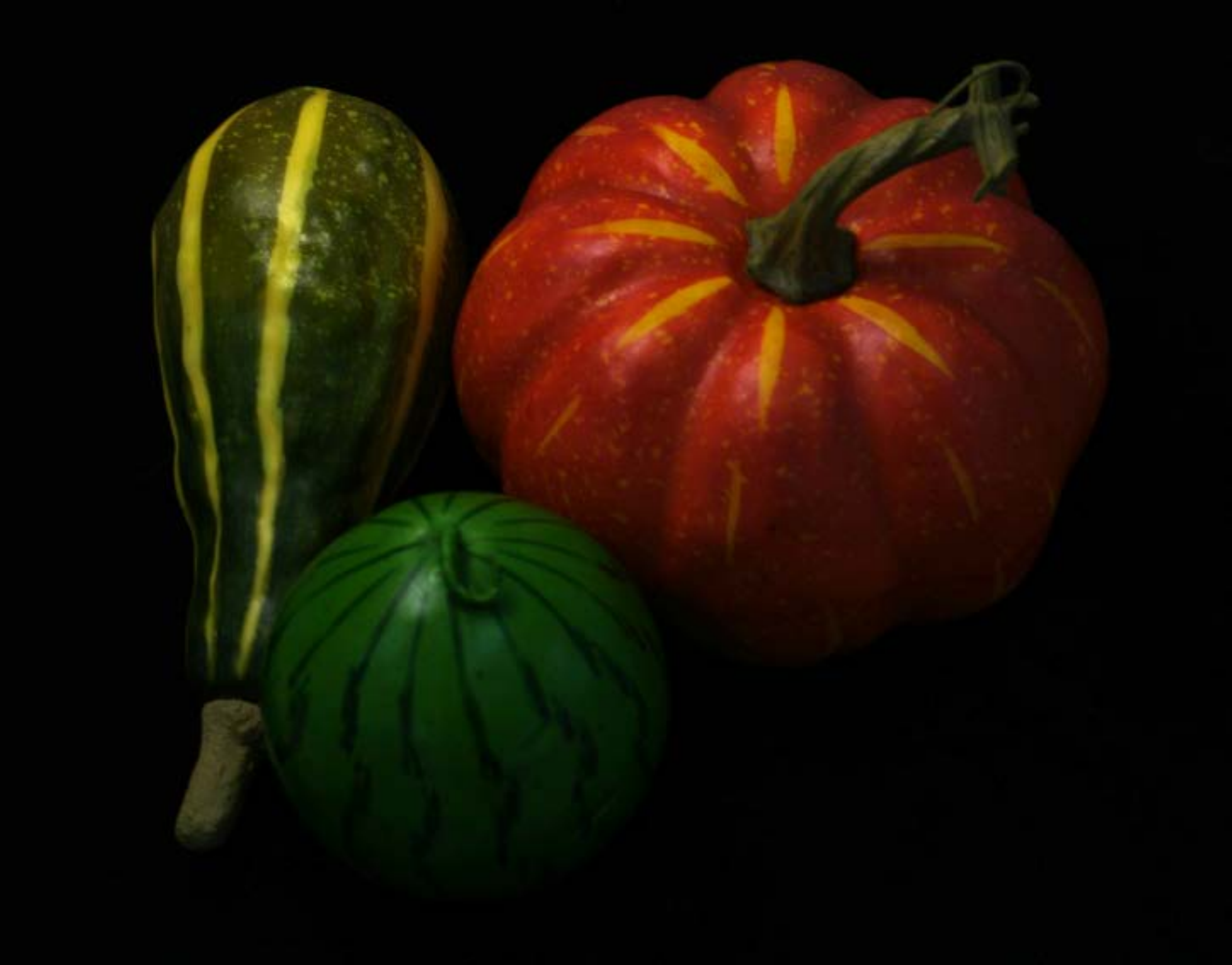}}\hspace{0mm}
  \subfigure[]{
      \includegraphics[width=0.48\linewidth]{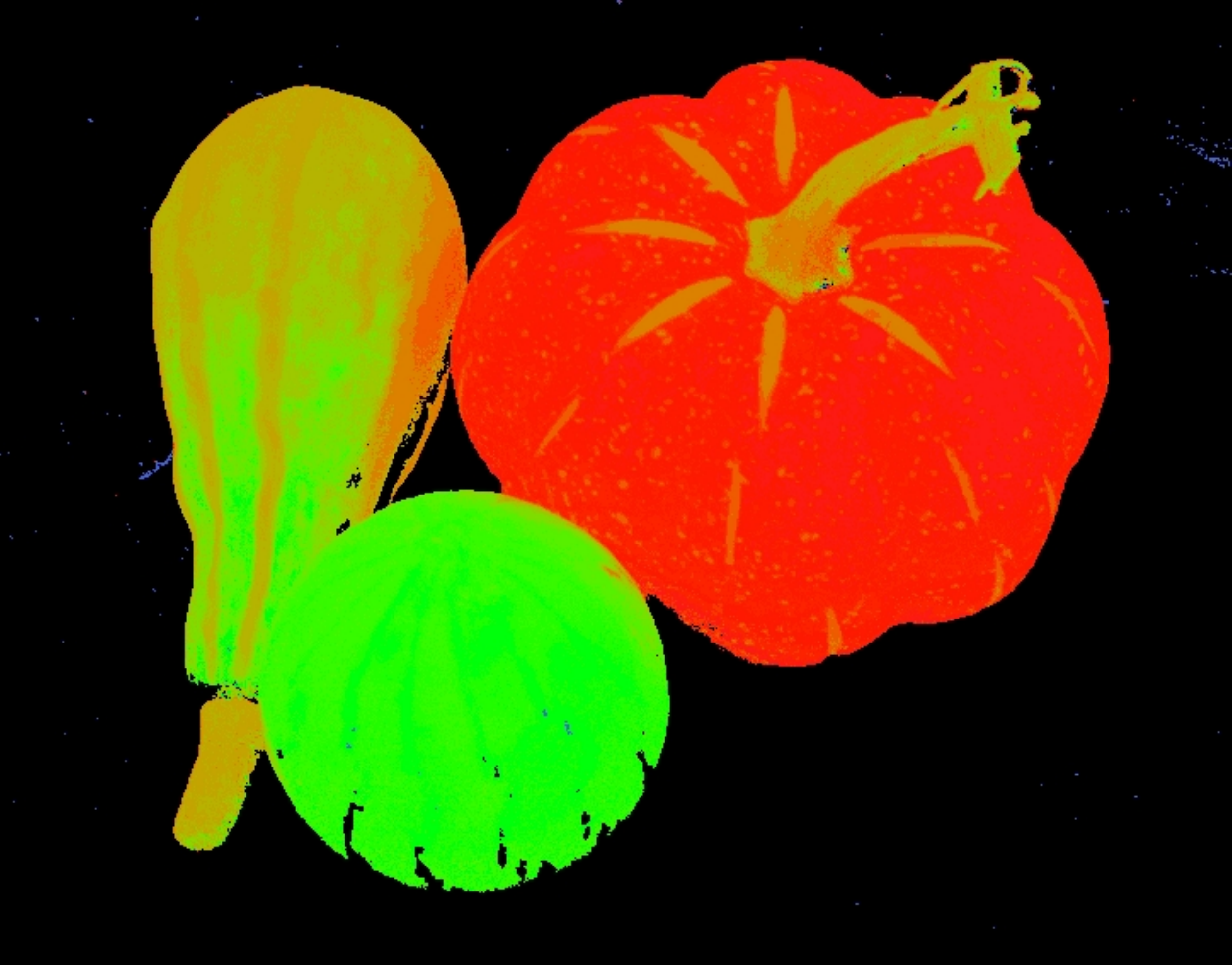}}
  \subfigure[]{
      \includegraphics[width=0.48\linewidth]{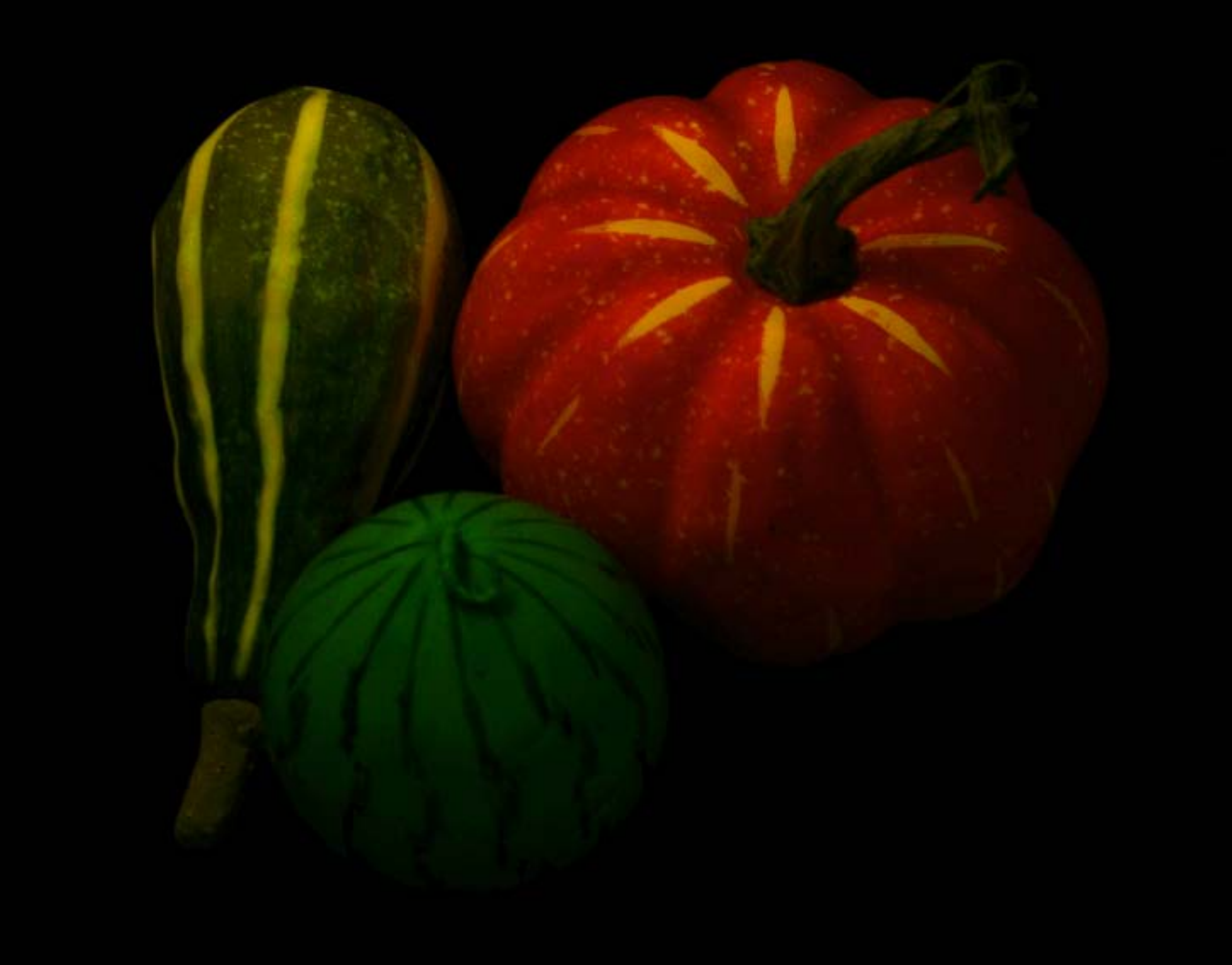}}\hspace{0mm}
  \subfigure[]{
      \includegraphics[width=0.48\linewidth]{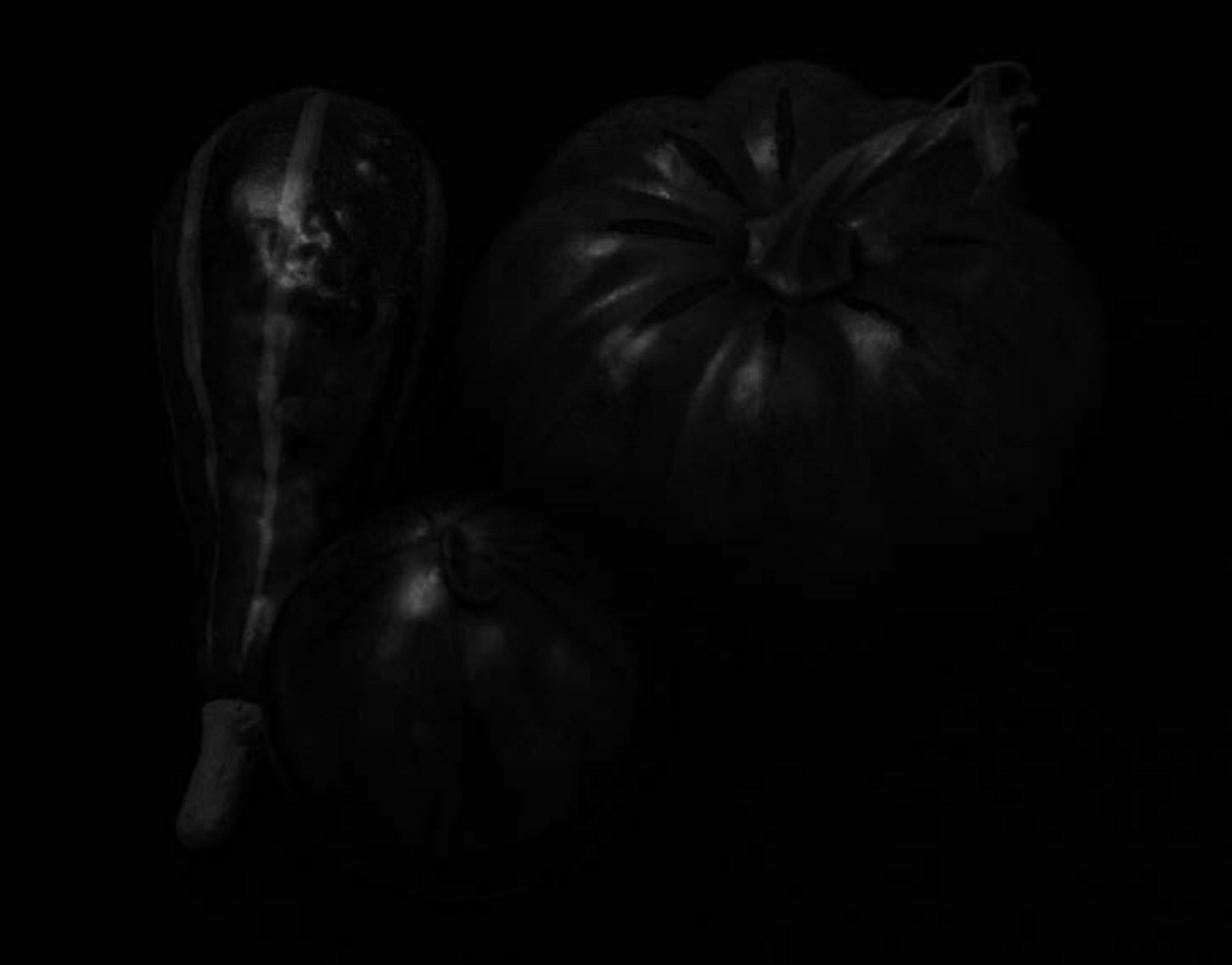}}
  \caption{An exemplar of non-Lambertian scene and its high light removal results by our approach. (a) The input image. (b) Specular robust clustering result. (c)(d) Separated diffuse and specular components, respectively. 
  }
  \label{fig:first_page} 
  \vspace{-2mm}
\end{figure}

Color information describes the scene's reflectance behaviors and plays an important role in various computer vision tasks, such as segmentation, recognition, matching and intrinsic image retrieval.
The image recording the diffuse reflection characterizes the color distribution of the scene, but
the image intensities of widely existing non-Lambertian surfaces, such as the scene displayed in Fig.~\ref{fig:first_page}(a), largely deviate from their true color information in the specular regions.
{There are also work directly based on the specular component, such as shape from specular reflection \cite{Lellmann01}.}
Therefore, separating highlight from diffuse component for the images of non-Lambertian scenes is of crucial importance.

\subsection{Related Works}
Highlight removal has been studied for decades, as reviewed in \cite{Artusi01}.
Physically, the degree of light polarization can be considered as a
strong indicator of specular reflection, while diffuse is considered unpolarized.
Therefore, some polarization based methods with hardware assistance have been proposed, such as \cite{Wolff01, Nayar02, Kim05} and \cite{Umeyama01}. 
Highlight also exhibits varying behaviours under different illumination directions or from different views, so some highlight removal approaches from multiple images are proposed. {Similarly, Feris et al. \cite{Feris01} use a set of images captured in the same point of view but with different flash positions to restore the diffuse component.} With a moving light source in \cite{Sato01}, Sato and Ikeuchi introduce a method exploiting the color signature analysis using the dichromatic model. Similarly, Lin and Shum\cite{Lin02} use linear basic functions to separate diffuse and specular components through two images captured with different light directions. Instead of using multi-illumination inputs, Lin et al.\cite{Lin01} alternatively propose a method using multi-baseline stereo based on the observation that the diffuse component was angle free while the specular component was angle dependent. Statistic techniques can also help highlight removal.

From the threshold to the gradient histograms, which is defined as the difference between the histograms of two different intensities, Chen et al. can \cite{Chen05} reconstruct the specular field successfully, but their method requires more than 200 input images.
Differently, with some statistical properties of nature scenes, Weiss \cite{Weiss01} formulate the recovery of diffuse component as a maximum likelihood estimation problem from an image sequence with the same diffuse component but different specular components. Yang et al.\cite{Qingxiong01} resort to statistical methods to remove specularity from two images with non-overlapping specular highlights. Although being able to successfully remove the specularity, approaches with hardware assistance or from multiple images are much less practical compared to the single image based approaches.

Removing highlights from a single image often needs given illumination, which can be either calibrated or estimated computationally\cite{Tan01}\cite{Drew14}\cite{Lilong01}. {Some work} \cite{Bajcsy01}\cite{Klinker02}\cite{Pesal01} require robust color segmentation for accurate specular detection, which is quite challenging.
{Tan and Ikeuchi} \cite{Tan03} make use of the difference between specular and diffuse pixels in their proposed specular-free image to remove the highlight effects. 
They iteratively replace the chromaticity at a specular position with its neighbor pixel with the maximum diffuse chromaticity until the algorithm converges.
To reduce the high computation cost of such a greedy searching strategy, Yang et al.\cite{Yang01} accelerate this method by introducing bilateral filtering. 
Similarly,
Mallick et al.~\cite{Mallick01} propose a PDE algorithm, which iteratively erodes the specular channel in the SUV color space, as in \cite{Mallick02}.
But these propagation methods cannot handle large area highlight.
Also using the local properties, {Kim} et al. ~\cite{Hyeongwoo01} propose an optimization algorithm utilizing the dark channel prior proposed in ~\cite{Kaiming01}. 
Another optimization based method is proposed by Akashi and Okatani ~\cite{Akashi14}, who formulate the separation as a non negative matrix decomposition problem. This method is sensitive to the initial values and one must run this method several times to get the most reasonable result. Besides, both the optimization based methods are very slow.
Researchers also spend efforts on highlight removal techniques based on material clustering. Tan et al.\cite{Tan02}, Shen and Zheng \cite{Shen13} both propose to conduct specular-independent material clustering and relate the specular intensity to its specular-free counterpart.
The strategy that clusters materials first and then recovers the intrinsic diffuse colors in each cluster is promising.
In the clustering step, the specular-free image in \cite{Shen13} is channel-independent and cannot discriminate colors like [1 2 3], [2 1 3] and [3 2 1].
In the step identifying the pure diffuse pixels of each cluster, both methods either involve some approximations or make strong assumptions, and fail in some cases, such as approaching-white material and strong specularity.

\subsection{Our Approach}
This paper focuses on highlight removal from a single image and targets for wide applicability to the large variety of nature scenes. 
Besides, this approach is also designed to be memory saving and of low running cost to handle high resolution images fast.

For a non-Lambertian surface, its reflectance can be represented by a linear combination of diffuse and specular components. Then, the highlight removal naturally falls into a signal separation problem.
Adopting a two-step strategy similar to \cite{Tan02}, we decompose the highlight removal procedure into two subproblems: (i) discriminating intrinsic colors (i.e., diffuse reflectance) of the pixels, either affected or unaffected by specular highlight; (ii) finding the pure diffuse pixels and then recovering the diffuse component in each cluster. 
Accurate identification of pixels with the same material is a nontrivial problem due to the influence of specularity.  
Within each cluster, distinguishing the pure diffuse pixels from those affected by specularity is also difficult.

To avoid the influences of specularity, material clustering in the specular-free subspace is favourable. With the known illumination, we directly project the image intensities into two subspaces, orthogonal to and parallel with the illumination direction, respectively.
{It is noteworthy that all the previous methods are based on the $L_1$ chromaticity definition, and there exist some approximations in the succeeding derivation of the highlight removal methods. In this paper, we propose an $L_2$ chromaticity definition and obtain an analytical highlight removal solution.}
From the newly defined chromaticity definition and the corresponding $L_2$ normalized dichromatic model, we derive an explicit analytical expression---a unit circle equation---between the parallel and orthogonal components of the pixels with the same chromaticity. Guided by this expression, an error term is defined to measure the color uniformity among the pixels and adaptively determine the number of diffuse colors in the clustering.
Fig.~\ref{fig:first_page}(b) displays the clustering results of the scene in Fig.~\ref{fig:first_page}(a). We can see that the clustering preserves the 
smoothly varying reflectance on the squash, and inter-reflection between them.
To find the pure diffuse pixels in each cluster, a strictly defined property under the $L_2$ normalized dichromatic model can be derived: within each cluster, the pixels with different specular strengths form a circular arc lying on the plane spanned by the illumination and the illumination-orthogonal directions. The pixels with increasing specular strengths move along the arc monotonously, and the pure diffuse pixels locate at one end. Here we name this pattern the pure diffuse pixels distribution rule (PDDR). { This rule is similar to the model proposed by Finlayson and Drew in \cite{Graham01}, where the pixels of the same material lie on a straight line. However, this model needs at least 4 channels.}
Besides, we can easily determine the specular strength for each pixel and recover the diffuse intensity.

Although bearing some similarity to the two-step strategy in \cite{Tan02} and \cite{Shen13}, our approach largely differentiates from these two work in finding the pure diffuse pixels and demonstrates higher performance.
Only based on the common assumption that there exist pure diffuse pixels for each material in the scene, our PDDR property makes the detection of these pixels very easy. Besides, without any approximation in the whole derivation, the detected pure diffuse pixels and the highlight removal results are of high accuracy.
Oppositely, in order to find the pixels unaffected by specular highlight, both \cite{Tan02} and \cite{Shen13} make some additional assumptions.
Tan et al. \cite{Tan02} assume that the 1st PCA basis of diffuse reflectance is wavelength free and result in unpleasant color bias and high sensitivity to noise.
Requiring calibrating the RGB response curves also limits its practicability.
In \cite{Shen13}, the criterion for recovering the diffuse color would fail for materials whose color approaches white.
Besides, in \cite{Shen13} the assumption that the number of specular affected pixels in each cluster does not exceed a threshold makes this method fail in handling scenes with widespread highlight.

In summary, the proposed approach has advantages over previous approaches in multiple aspects.

\begin{enumerate}
\item Due to the new defined $L_2$ chromaticity, $L_2$ normalized dichromatic model and the PDDR property, we can achieve robust scene adaptive material clustering and accurate recovery of the diffuse colors, even when they are {similar} to the illumination.
\item Making use of the global structure instead of the local information, the proposed method can handle images with large area or strong specularity.
\item Without complex optimization, our method can remove highlight {fast for high resolution images}. 
\end{enumerate}



\section{The $L_2$ Normalized Dichromatic Model}
In this paper, we normalize the widely used dichromatic model \cite{Shafer01} by $L_2$ norm and correspondingly derive an orthogonal decomposition strategy for the surface appearance.

For a color camera, the imaging process can be formulated as:
\begin{equation}\label{eqs:Imaging}
{\bf{I}}_c({\bf{x}})=\omega_r({\bf{x}})\int_{\Lambda}r({\bf{x}},\lambda)e(\lambda)q_c(\lambda)d\lambda
+\omega_l({\bf{x}})\int_{\Lambda}e(\lambda)q_c(\lambda)d\lambda.
\end{equation}Here ${\bf{I}}_c({\bf{x}})$ is the intensity of channel $c$ at pixel $\bf{x}$, with $c$ indexing the camera channel and ${\bf{x}}=\{x,y\}$ representing the 2D location. On the righthand side of the equation, the two terms are respectively the diffuse and specular components, with $\omega_r({\bf{x}})$ and $\omega_l({\bf{x}})$ being the corresponding strengths. In each term, $\lambda$ denotes the wavelength with range being $\Lambda$, while $r(\lambda)$, $e(\lambda)$ and $q_c(\lambda)$ represent the surface reflectance, the illumination spectrum and the camera's spectral response of channel $c$, respectively.

\begin{figure}[t]
  \centering
  \includegraphics[width=0.98\linewidth]{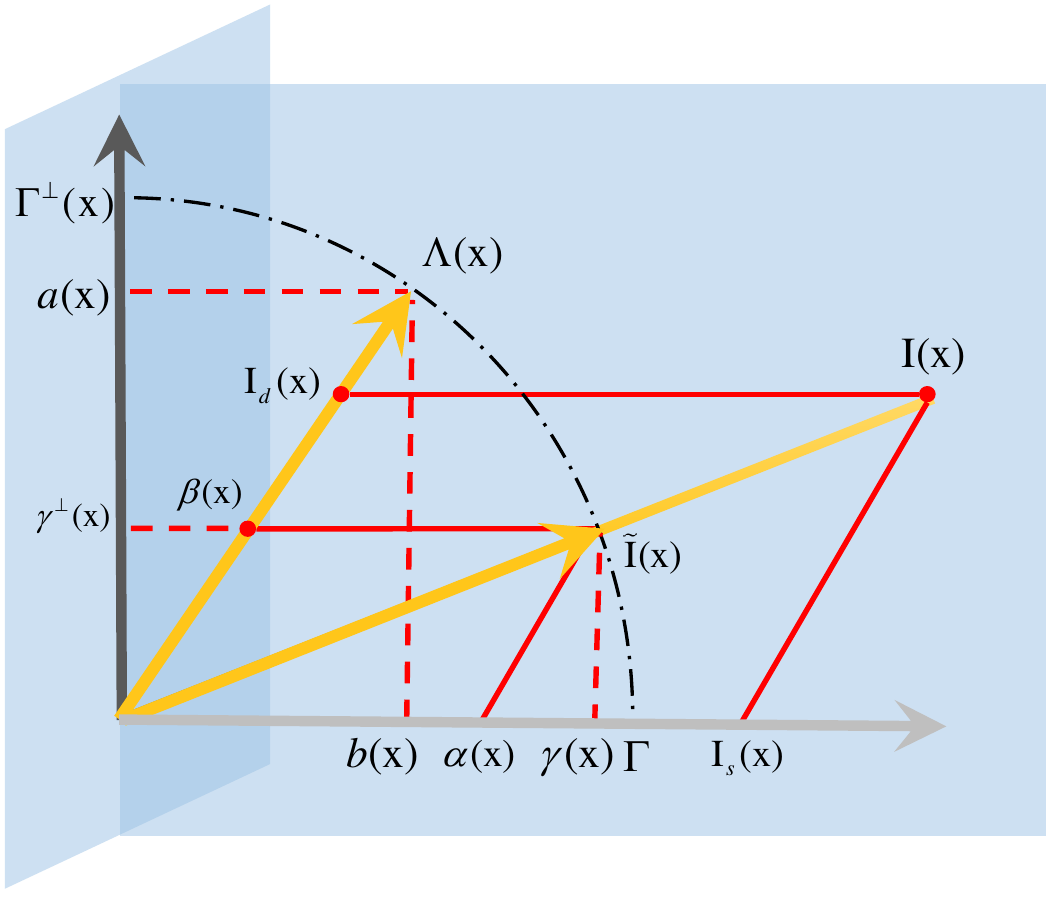}
  \caption{The illustration and notations of the proposed $L_2$ normalized dichromatic model.}
  \label{fig:notation} 
\end{figure}

Denoting  the pure diffuse component and  specular highlight as ${\bf{I}}_d({\rm{\bf{x}}})=\omega_r({\bf{x}})\int_{\Lambda}r({\bf{x}},\lambda)e(\lambda)q_c(\lambda)d\lambda$ and ${\bf{I}}_s({\rm{\bf{x}}})=\omega_l({\bf{x}})\int_{\Lambda}e(\lambda)q_c(\lambda)d\lambda$, Eq.~\ref{eqs:Imaging} can be represented by
\begin{equation}\label{eqs:Iseparation1}
{\bf{I}}({\rm{\bf{x}}})={\bf{I}}_d({\rm{\bf{x}}})+{\bf{I}}_s({\rm{\bf{x}}}).
\end{equation}
The diffuse component represents the inherent color of the surface and the specular highlight implies the color of illumination, as illustrated in Fig.~\ref{fig:notation}.
Different from the previous definition of color chromaticity $\widetilde{\bf{I}}({\bf{x}})=\frac{{\bf{I}}({\bf{x}})}{\sum_{c\in\{r,g,b\}} {{\bf{I}}_c({\bf{x}})}}$ as used in \cite{Tan01}\cite{Hyeongwoo01}\cite{Tan02}\cite{Shen13}\cite{Cong01}, we propose a $L_2$ definition as
\begin{equation}\label{eqs:Chromaticity}
{\widetilde{\bf{I}}}({\bf x})={{\bf{I}}({\rm{\bf{x}}})}/{\|{\bf{I}}({\rm{\bf{x}}})\|_F}.
\end{equation}In this equation, $\|{\bf{I}}({\rm{\bf{x}}})\|_F=\sqrt{\sum_{c\in\{r,g,b\}} {\bf{I}}_c({\rm{\bf{x}}})^2}$, with ${\bf{I}}_c({\rm{\bf{x}}})$ being the intensity of the $c$th channel.

Similarly, we denote $\Lambda$ and $\Gamma$ as the chromaticity of diffuse component and specular highlight respectively:
\begin{eqnarray}\label{eqs:gl}
{{\Lambda}}({\bf x})={{\bf{I}_d}({\rm{\bf{x}}})}/{\|{\bf{I}_d}({\rm{\bf{x}}})\|_F}\\\nonumber~~~~~~~~
{{\Gamma}}({\bf x})={{\bf{I}_s}({\rm{\bf{x}}})}/{\|{\bf{I}_s}({\rm{\bf{x}}})\|_F}.
\end{eqnarray}Based on the above definition and assuming that all the pixels are illuminated by the identical illumination color (i.e., $\Gamma$ is position-independent), the normalized reflectance ${\widetilde{\bf{I}}}({\bf x})$ can be written as
\begin{equation}\label{eqs:uni_Refle}
{\widetilde{\bf{I}}}({\bf x})=\alpha({\bf{x}}){\Lambda}({\bf{x}})+\beta({\bf{x}})\Gamma.
\end{equation}Here the coefficients $\alpha({\bf{x}})=\|{\bf{I}}_d({\bf{x}})\|_F/\|{\bf{I}}({\bf{x}})\|_F$ and $\beta({\bf{x}})=\|{\bf{I}}_s({\bf{x}})\|_F/\|{\bf{I}}({\bf{x}})\|_F$.


For the chromaticity of diffuse component ${\Lambda}({\bf{x}})$, it can be projected into two subspaces, one is parallel with while the other is orthogonal to the illumination direction $\Gamma$, as illustrated by the two planes in Fig.~\ref{fig:notation}. The projection procedure can be conducted according to the orthogonal projection algorithm proposed by Chang \cite{Chein01} as
\begin{equation}\label{eqs:d_decom}
{\Lambda}({\bf{x}})=a({\bf{x}}){\Gamma}^{\bot}({\bf{x}})+b({\bf{x}}){\Gamma},
\end{equation}in which ${\Gamma}^T{\Gamma}^{\bot}({\bf{x}})=0,\|{\Gamma}^{\bot}({\bf{x}})\|_F^2=1$, $\|{\Gamma}\|_F^2=1$ and the items $a({\bf{x}})$ and $b({\bf{x}})$ are the projection coefficients along two directions, respectively.
Since $\Lambda({\bf x})$ is normalized, we can easily derive that
\begin{equation}
a({\bf{x}})^2+b({\bf{x}})^2=1.
\end{equation}

\begin{figure*}[t]
\begin{center}
   \includegraphics[width=1.0\linewidth]{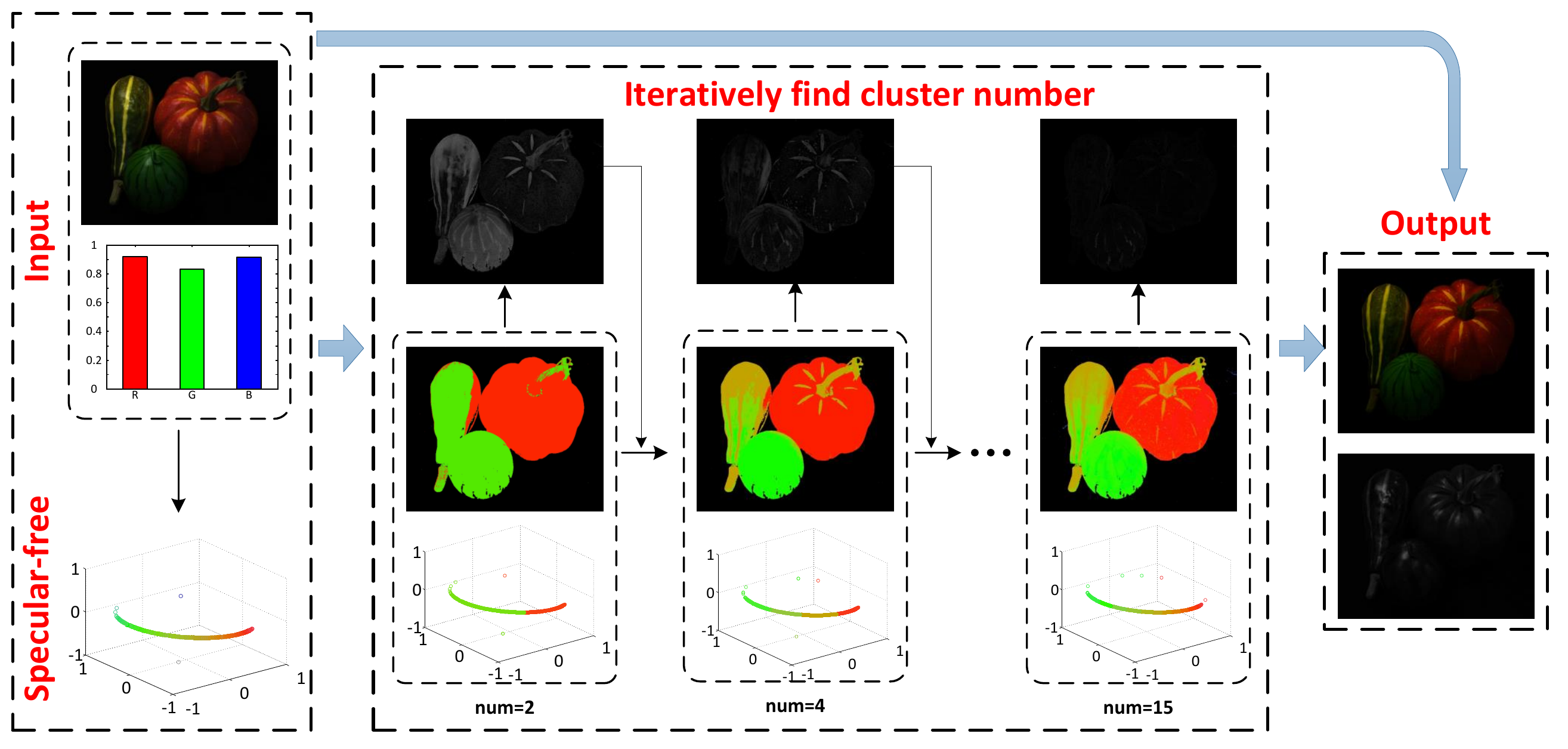}
\end{center}
   \caption{The frame of our highlight removal approach via adaptive material clustering. {\bf Left column:} The upper block shows the input normalized image and illumination, and the lower block shows their projection in the specular-free space. {\bf Middle column:} The procedure of adaptive clustering, with the clustering results in the specular-free space at bottom, corresponding labeling in the original image in the
middle, and the fitting error to the analytical model Eq.~(\ref{eqs:coef_eqs}) on the top. {\bf Right column: } The separated diffuse and specular components.}
\label{fig:framework}
\end{figure*}

Substituting Eq.~(\ref{eqs:d_decom}) into Eq.~(\ref{eqs:uni_Refle}), we can obtain
\begin{equation}\label{eqs:uni_refle}
{\widetilde{\bf{I}}}({\bf x})=\alpha({\bf{x}})a({\bf{x}}){\Gamma}^{\bot}({\bf{x}})+(\alpha({\bf{x}})b({\bf{x}})+\beta({\bf{x}})){\Gamma},
\end{equation}which can be further simplified into
\begin{equation}\label{eqs:uni_refle1}
{\widetilde{\bf{I}}}(x)=\gamma^{\bot}({\bf{x}}){\Gamma}^{\bot}({\bf{x}})+\gamma({\bf{x}}){\Gamma},
\end{equation}by setting $\gamma^{\bot}({\bf{x}})=\alpha({\bf{x}})a({\bf{x}})$ and $\gamma({\bf{x}})=(\alpha({\bf{x}})b({\bf{x}})+\beta({\bf{x}}))$.
Since ${\widetilde{\bf{I}}}({\bf x})$, ${\Gamma}^{\bot}({\bf{x}})$ and ${\Gamma}$ are normalized, 
we have
\begin{equation}\label{eqs:coef_eqs}
\gamma^{\bot}({\bf{x}})^2+\gamma({\bf{x}})^2=1.
\end{equation}From Fig.~\ref{fig:notation}, adopting the proposed $L_2$ chromatic definition, all the reflectance with the identical diffuse chromaticity satisfy this equation, i.e., lie on the dot-slash unit circle. 

\section{Scene Adaptive Highlight Removal}
According to the $L_2$ normalized dichromatic model, the direction $\Gamma^{\bot}({\bf{x}})$ is solely determined by the diffuse chromaticity $\Lambda({\bf x})$ and can be used to discriminate different materials. Under the guidance of PDDR property, the coefficient $\gamma({\bf{x}})$ corresponding to the direction $\Gamma$ can be used to find the pure diffuse pixels and remove specular highlight in each cluster. Summing up these properties in both directions, we propose a highlight removal framework, as illustrated in Fig.~\ref{fig:framework}, using the image in Fig.~\ref{fig:first_page}(a) as a running example.

Given the input specular contaminated image and corresponding illumination (shown in the top-left block), we firstly 
get the normalized specular-free component $\Gamma^{\bot}({\bf{x}})$ corresponding to each pixel, as shown in the bottom-left block. Afterwards, we iteratively conduct specular-free clustering using K-Means by increasing the number of clusters until converge, with the convergence criterion defined from the fitting error to the analytical model in Eq.~(\ref{eqs:coef_eqs}), as visualized in the top row, middle column of Fig.~\ref{fig:framework}. The evolution of the clustering results is displayed in the lower two rows of middle column in Fig.~\ref{fig:framework}: the upper one in the image space and the lower one in the parameter space. Thereafter, we separate the specular and diffuse components at each pixel, with the results shown on the rightmost column. The following subsections will detail the two successive steps: material clustering and diffuse component recovery.

\subsection{Material Clustering}

\begin{figure*}[t]
\centering
  \vspace{-2mm}
  \includegraphics[width=0.98\linewidth,height=0.99\linewidth]{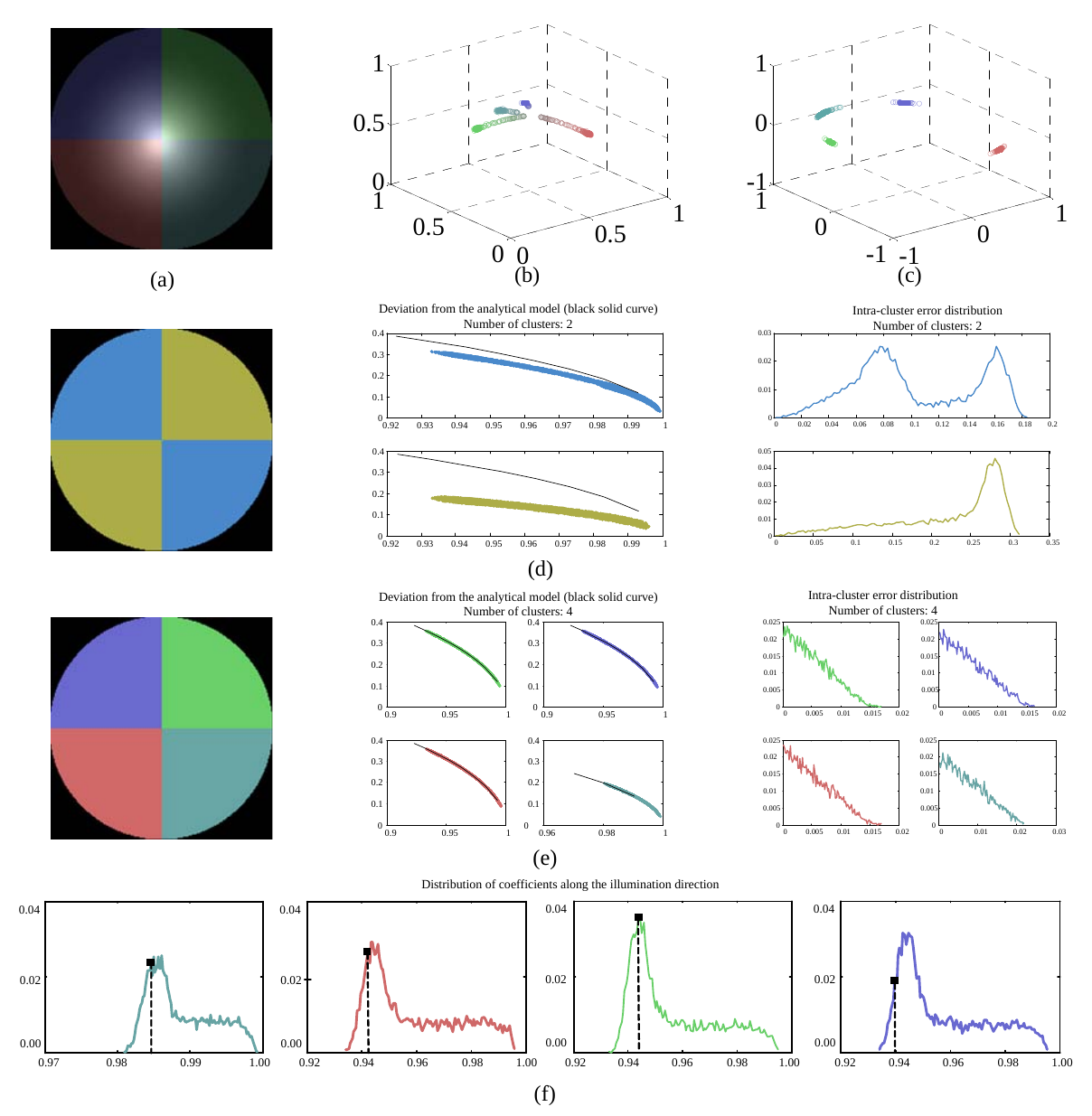}
  \vspace{-3mm}
  \caption{Description of our automatic clustering scheme on a synthetic image.  (a) The synthetic input image. (b)(c) The normalized reflectance in RGB space and that in the specular-free space, respectively. (d) From left to right: the clustering result with incorrect class number (i.e., 2), corresponding deviation from the unit circle description and statistics of the fitting errors. (e) The clustering results, the distribution of the pixels in the unit circle spaces and the corresponding fitting errors with the correct cluster number (i.e., 4). (f) The identification of the diffuse component from the histogram of specular coefficients $\gamma$s. The line colors correspond to the four clusters in (e) and solid dots denote the results.}
  \label{fig:hr_remove} 
\end{figure*}

For a specularity-influenced image, such as the one displayed in  Fig.~\ref{fig:hr_remove}(a), it is difficult to locate the pixels with the same diffuse reflectance in the original space because the influences of specularity, as shown in Fig.~\ref{fig:hr_remove}(b).
To avoid the affects from the specularity, we cluster the materials in the illumination orthogonal subspace ${\Gamma}^{\bot}({\bf{x}})$, in which the pixels with the same diffuse color but different specular strengths cluster well, as shown in Fig.~\ref{fig:hr_remove}(c), { which corresponds to the normalized data along ${\Gamma}^{\bot}({\bf{x}})$ in Fig.~\ref{fig:notation}}. Without loss of generalization, the proposed algorithm determines the number of clusters adaptively.

We start from a cluster number no larger than the true material types, and increase it successively until it reaches the correct number. Usually, we can set the initial cluster number as 1 for safe. In each iteration, we firstly project the chromaticity of the input image into the illumination orthogonal subspace for clustering. Then for each cluster, we replace ${\Gamma}^{\bot}({\bf{x}})$ with the cluster center ${\bf{C}}$, and re-project the normalized reflectance ${\widetilde{\bf{I}}}({\bf x})$ into ${\bf{C}}$ and $\Gamma$ to get coefficients $\gamma^{\bot}$ and $\gamma$. According to Eq.~(\ref{eqs:coef_eqs}), if the sum of squares of the coefficients is close to 1, i.e., lying on a unit circle, ${\bf{C}}$ represents this cluster well. On the contrary, if the pixels with coefficients deviating from the unit circle are more than the given threshold (e.g., 10\%), we suppose that the pre-set cluster number is incorrect. We use the total fitting error to the unit circle as the clustering precision, and set the threshold to be 0.1 empirically. After checking all the clusters, we increase the cluster by the number of clusters not meeting the precision criteria and go into the next iteration. Fig.~\ref{fig:hr_remove}(d) shows the clustering result and the corresponding error of Fig.~\ref{fig:hr_remove}(a) with an incorrect cluster number. The result obviously deviates from the dichromatic model.

We terminate the iteration when the cluster number stops increasing, and the correct clustering result of Fig.~\ref{fig:hr_remove}(a) is visualized in the leftmost image in Fig.~\ref{fig:hr_remove}(e). The corresponding fitting error to the proposed unit circle model and its distribution are also displayed. Experimentally, the algorithm converges within 5 iterations for most scenes.

\begin{figure*}[p]
\centering
\begin{minipage}[ht]{0.02\textwidth}
      \vspace{-10mm}
      (a)\vspace{35mm}
      (b)\vspace{48mm}
      (c)\vspace{48mm}
      (d)\vspace{48mm}
      (e)
 \end{minipage}\hspace{3mm}
\begin{minipage}[ht]{0.14\textwidth}
      \includegraphics[width=0.98\linewidth]{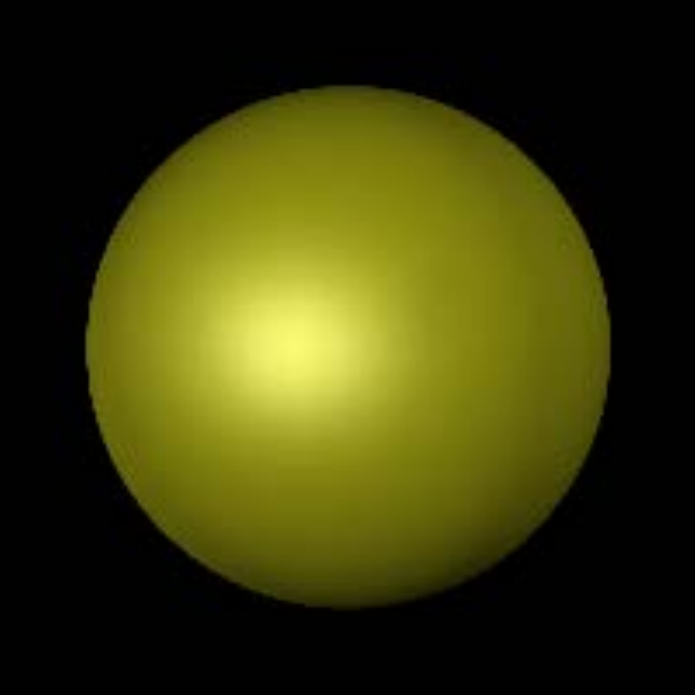}\\\vspace{-3mm}
      \includegraphics[width=0.98\linewidth]{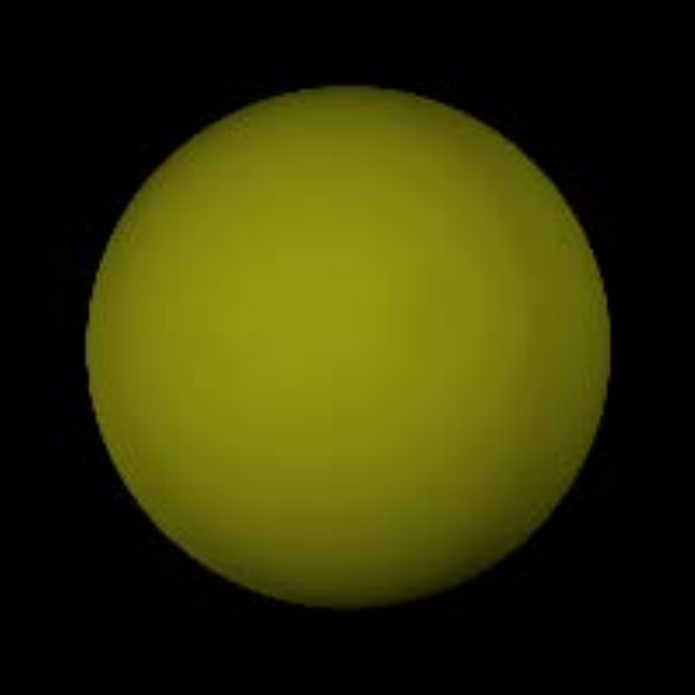}\\
      \includegraphics[width=0.98\linewidth]{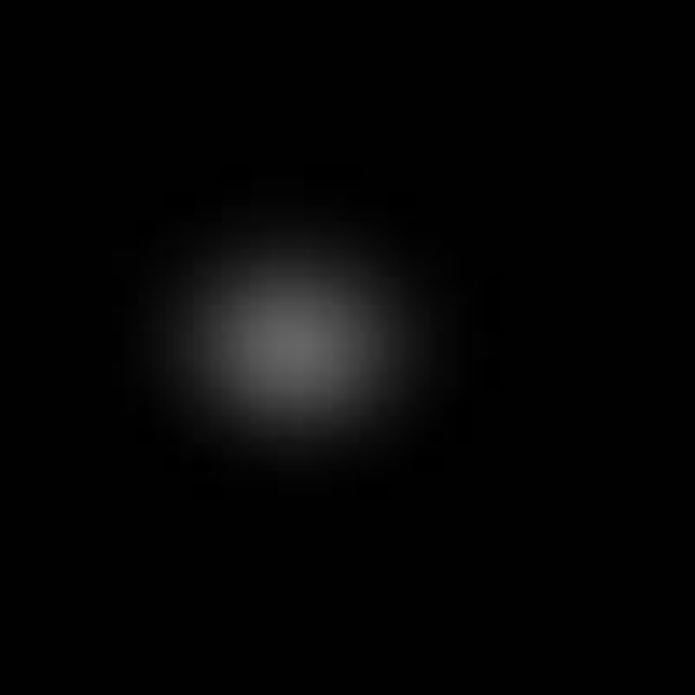}\\\vspace{-3mm}
      \includegraphics[width=0.98\linewidth]{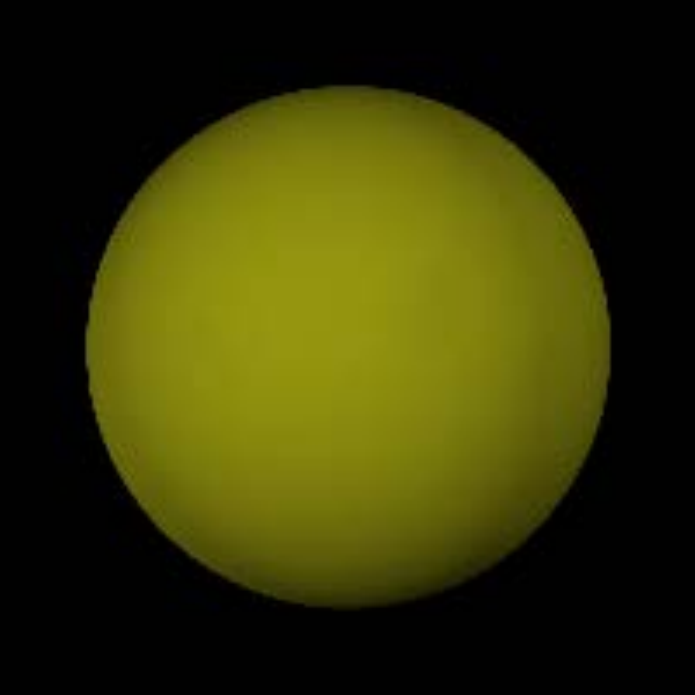}\\
      \includegraphics[width=0.98\linewidth]{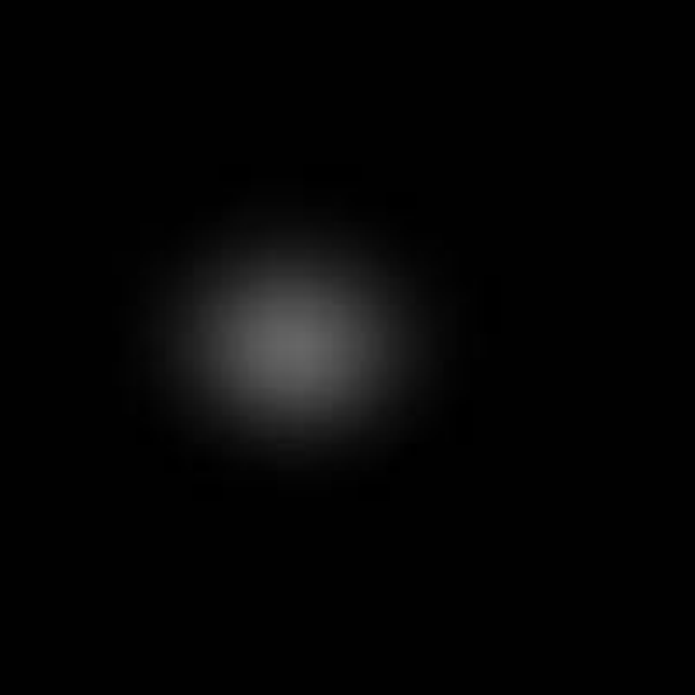}\\\vspace{-3mm}
      \includegraphics[width=0.98\linewidth]{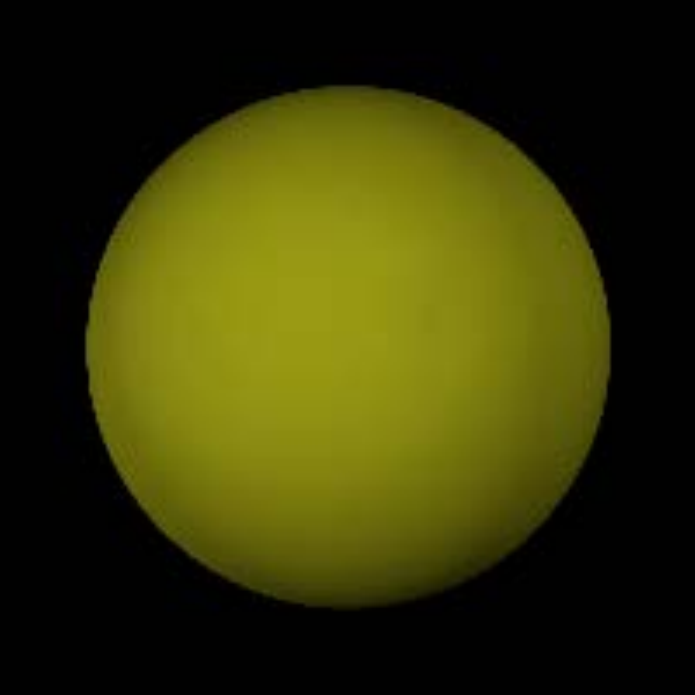}\\
      \includegraphics[width=0.98\linewidth]{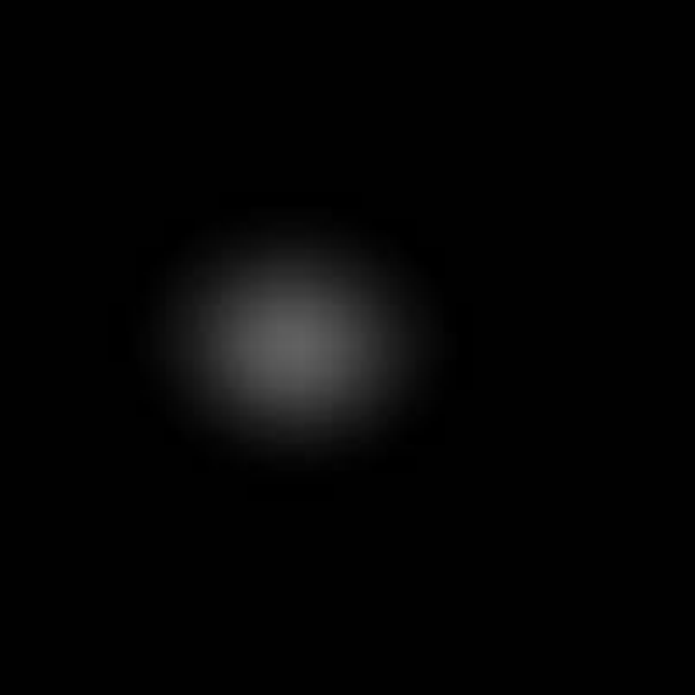}\\\vspace{-3mm}
      \includegraphics[width=0.98\linewidth]{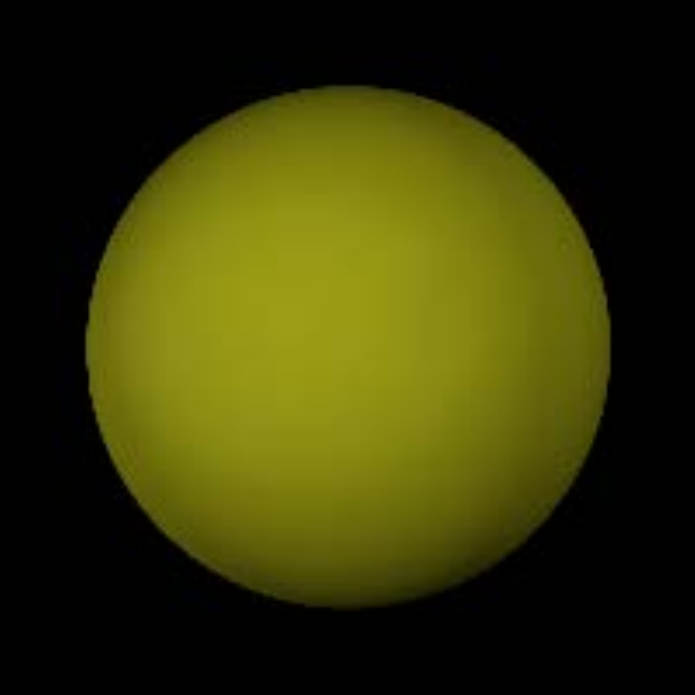}\\
      \includegraphics[width=0.98\linewidth]{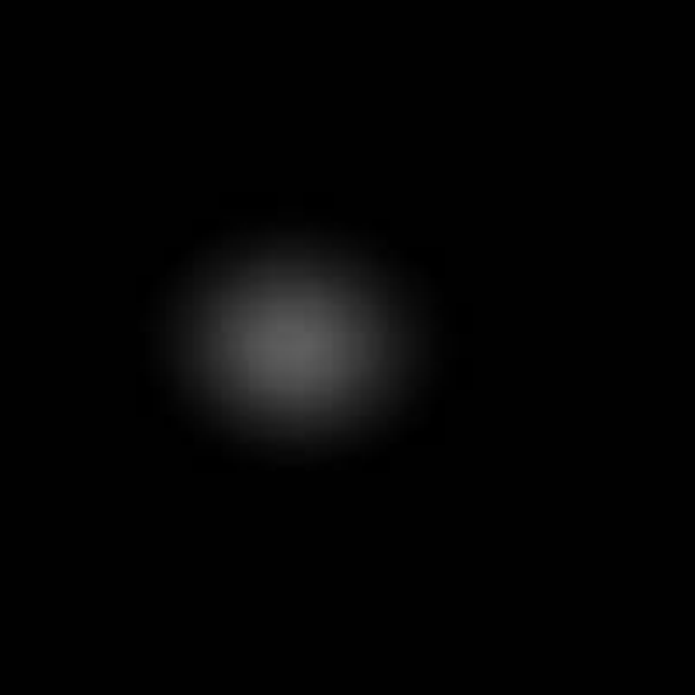}
 \end{minipage}
\begin{minipage}[ht]{0.14\textwidth}
      \includegraphics[width=0.98\linewidth]{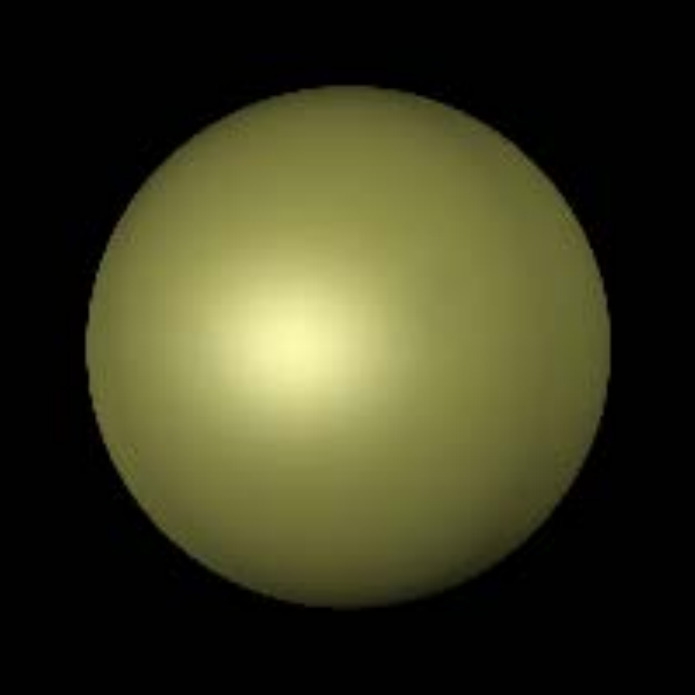}\\\vspace{-3mm}
      \includegraphics[width=0.98\linewidth]{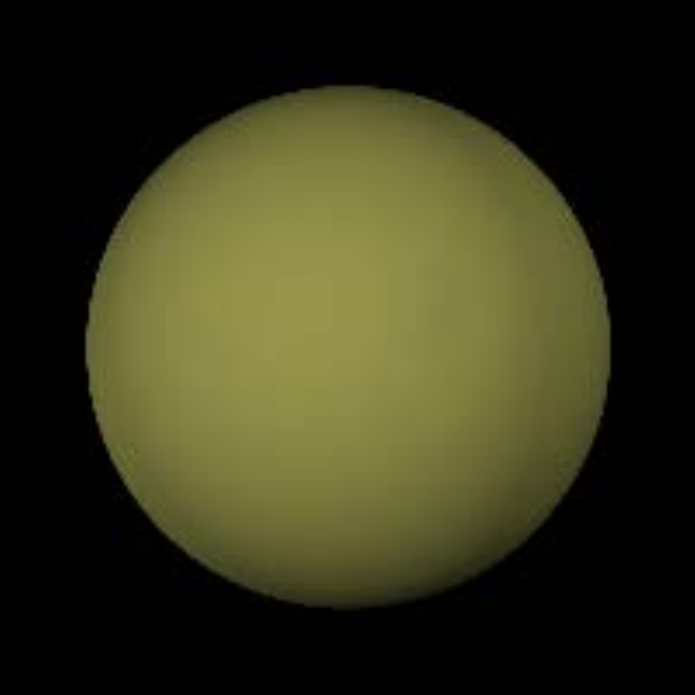}\\
      \includegraphics[width=0.98\linewidth]{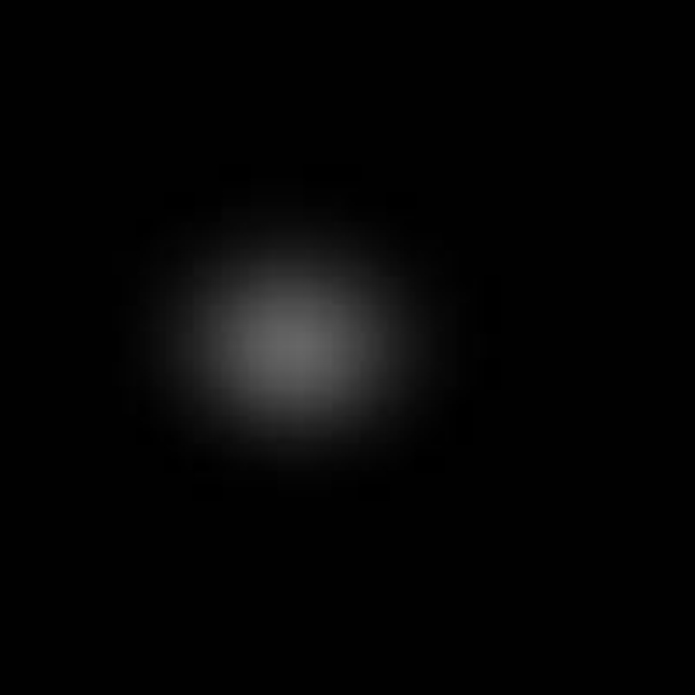}\\\vspace{-3mm}
      \includegraphics[width=0.98\linewidth]{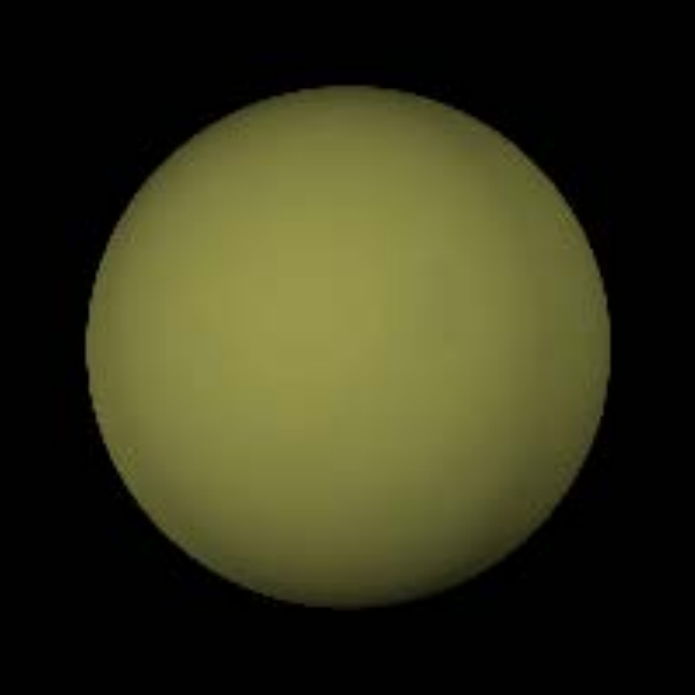}\\
      \includegraphics[width=0.98\linewidth]{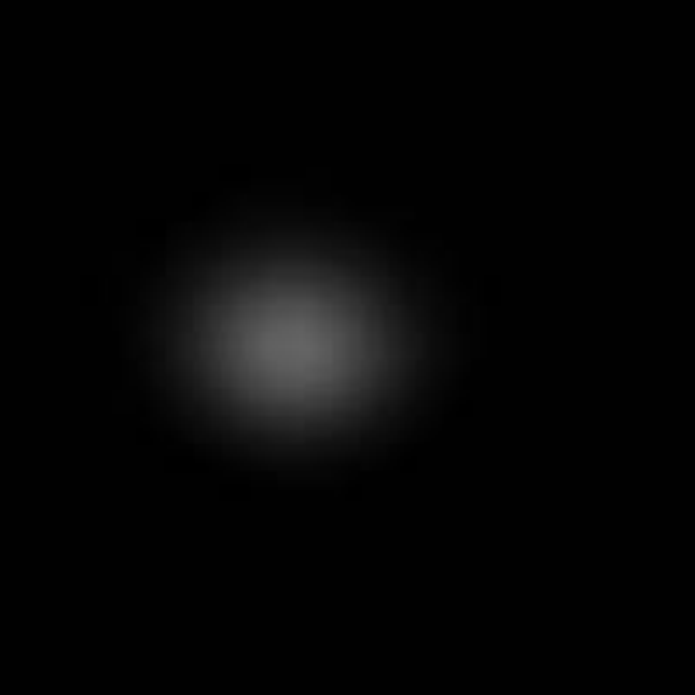}\\\vspace{-3mm}
      \includegraphics[width=0.98\linewidth]{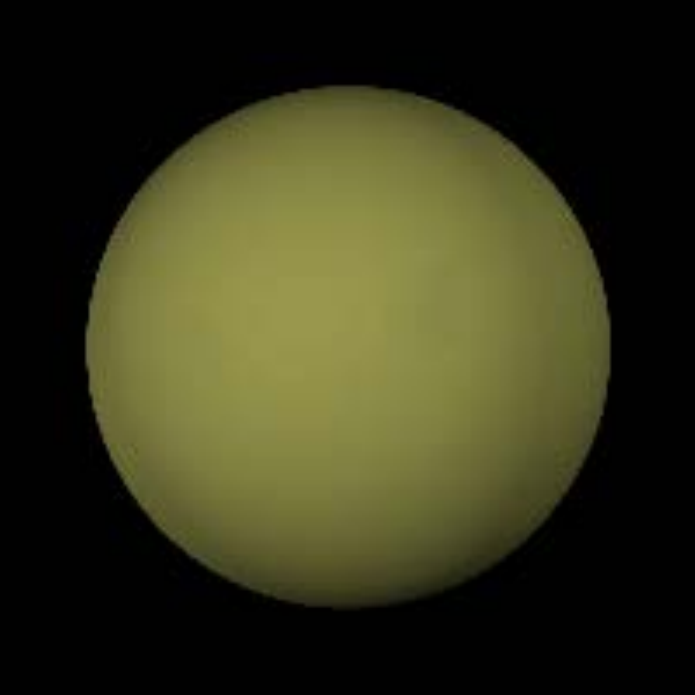}\\
      \includegraphics[width=0.98\linewidth]{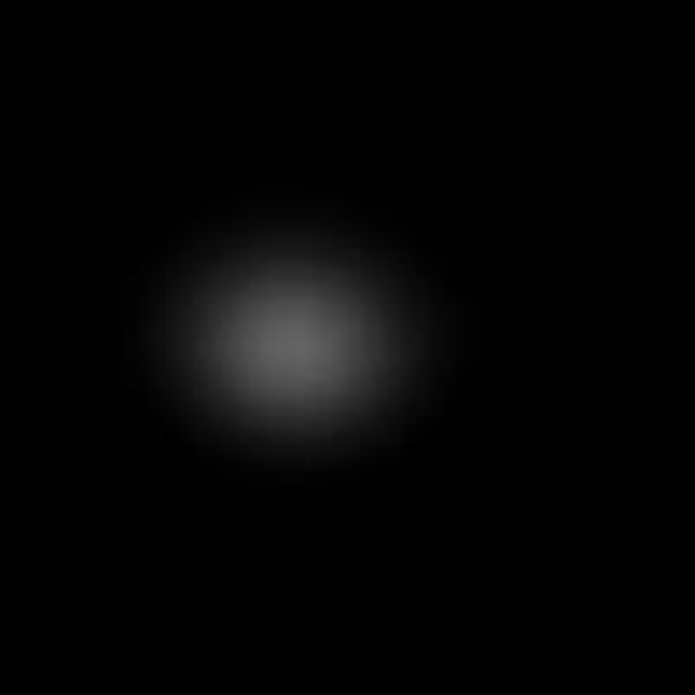}\\\vspace{-3mm}
      \includegraphics[width=0.98\linewidth]{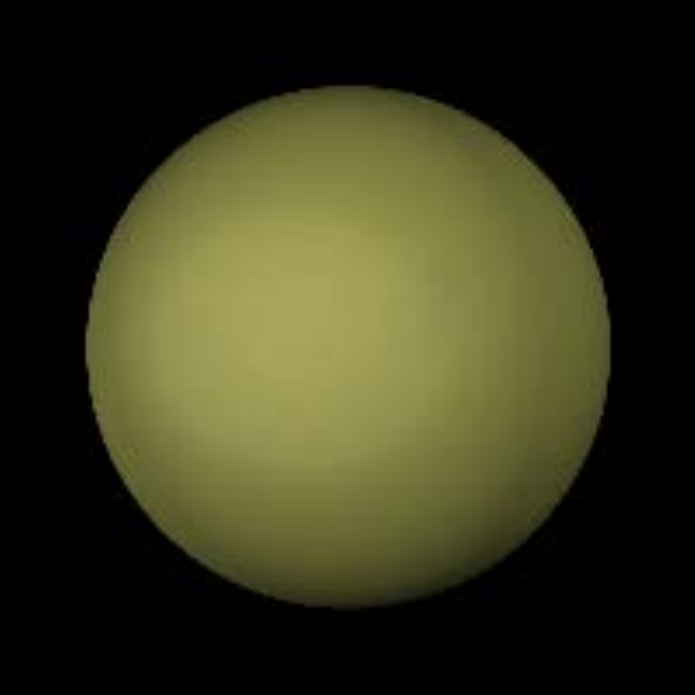}\\
      \includegraphics[width=0.98\linewidth]{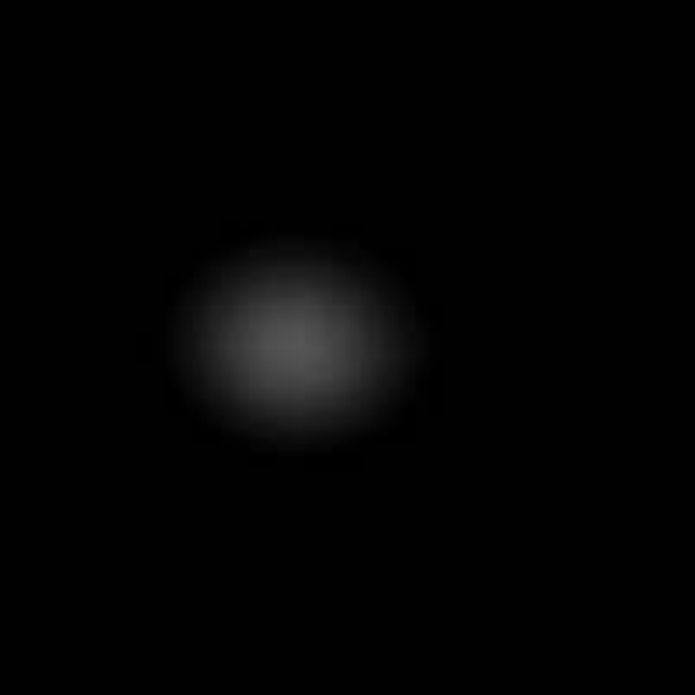}
 \end{minipage}
\begin{minipage}[ht]{0.14\textwidth}
      \includegraphics[width=0.98\linewidth]{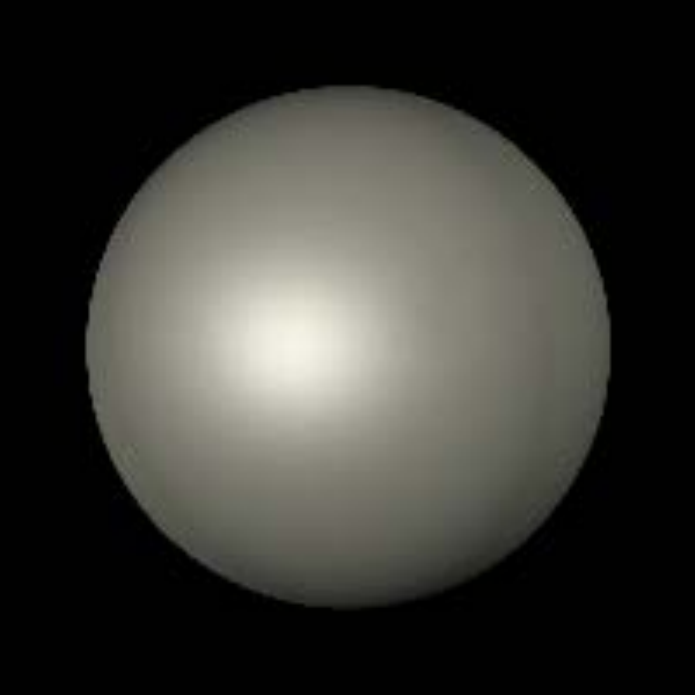}\\\vspace{-3mm}
      \includegraphics[width=0.98\linewidth]{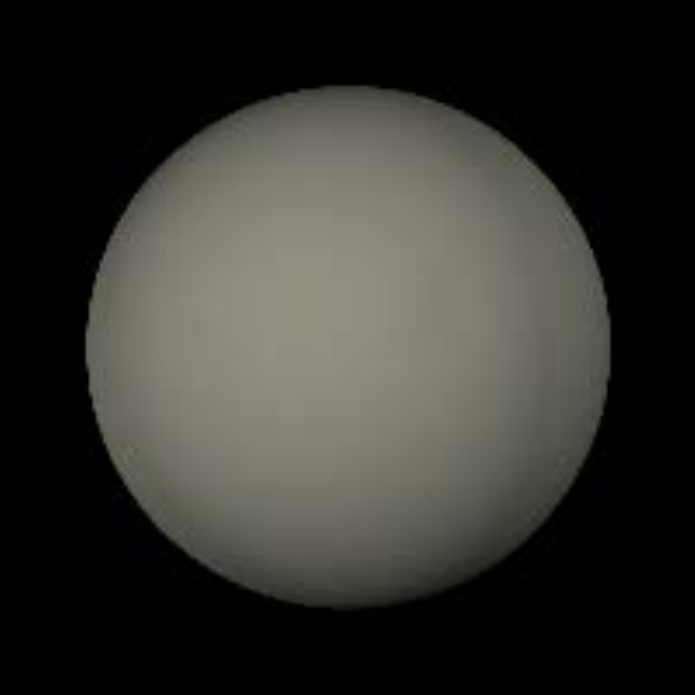}\\
      \includegraphics[width=0.98\linewidth]{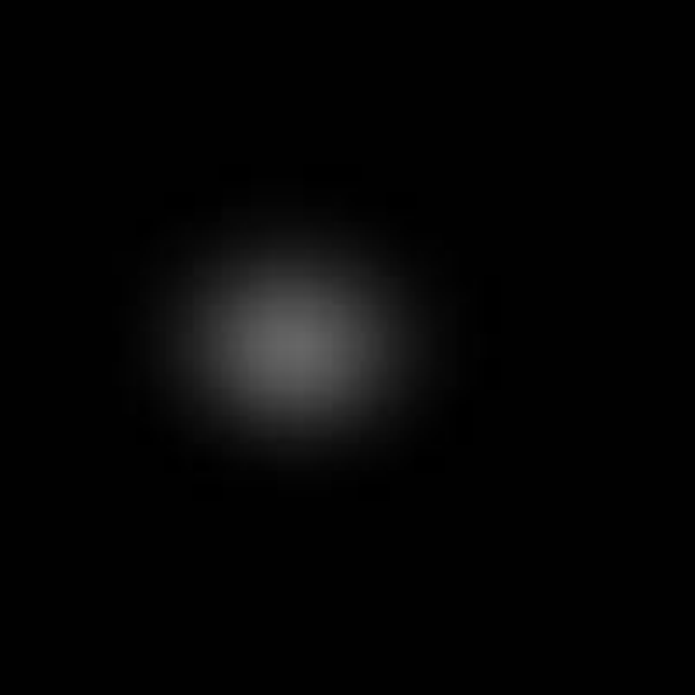}\\\vspace{-3mm}
      \includegraphics[width=0.98\linewidth]{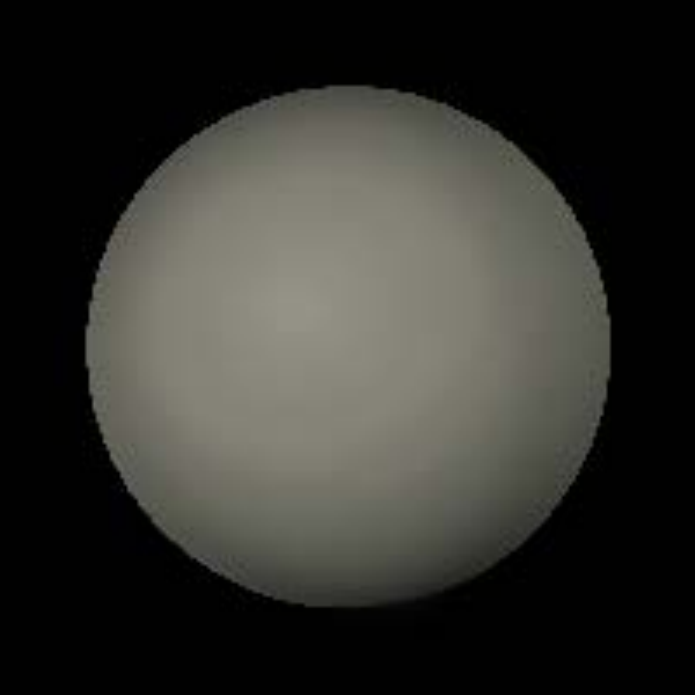}\\
      \includegraphics[width=0.98\linewidth]{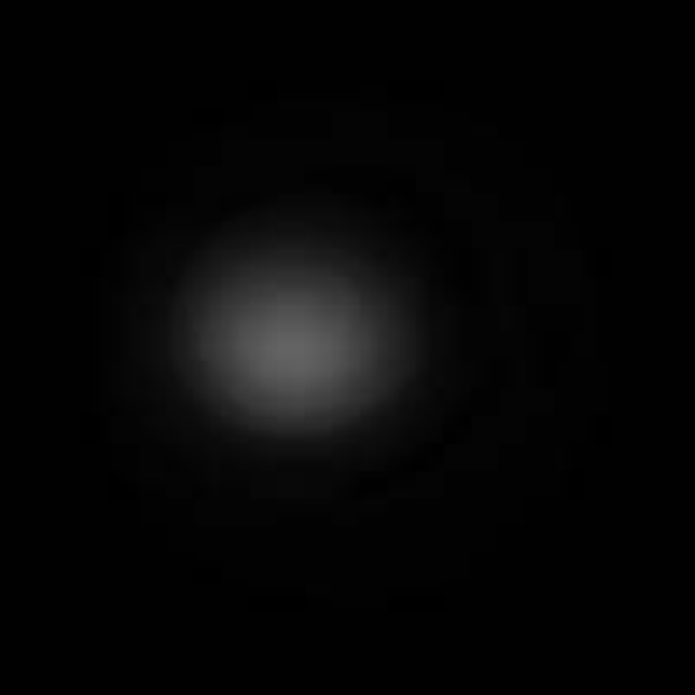}\\\vspace{-3mm}
      \includegraphics[width=0.98\linewidth]{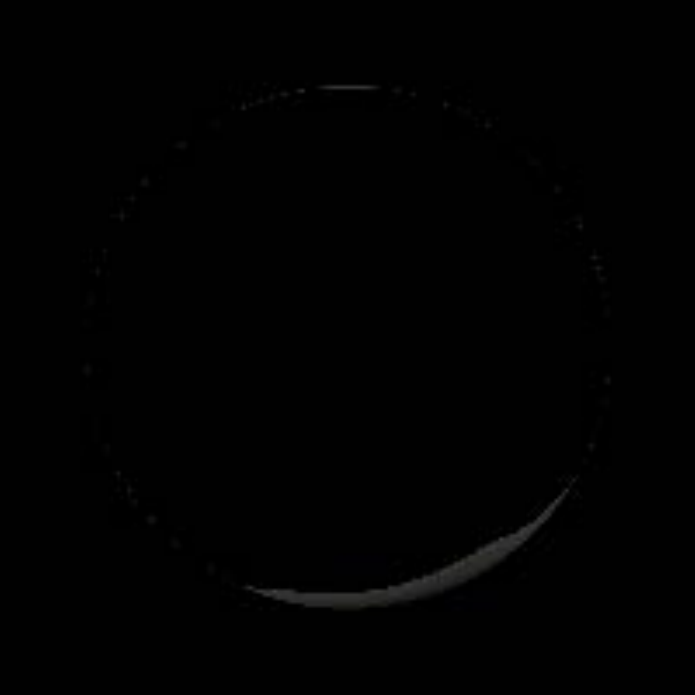}\\
      \includegraphics[width=0.98\linewidth]{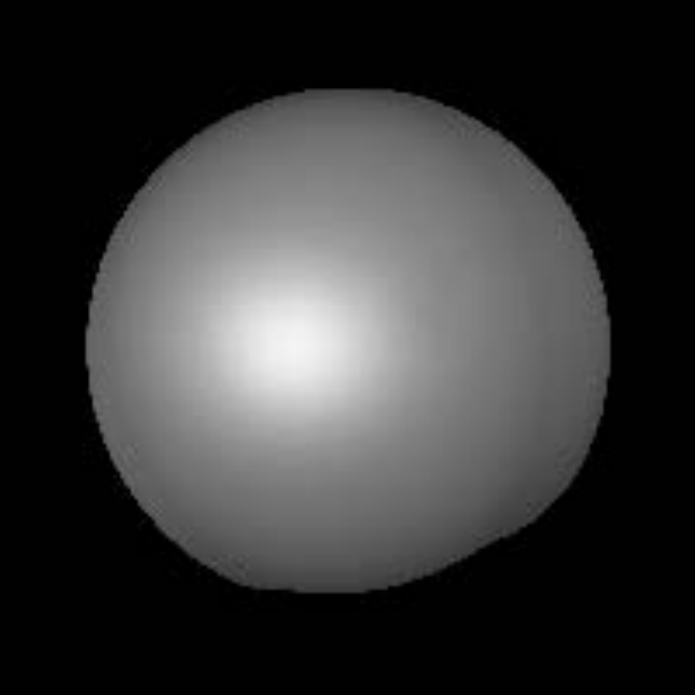}\\\vspace{-3mm}
      \includegraphics[width=0.98\linewidth]{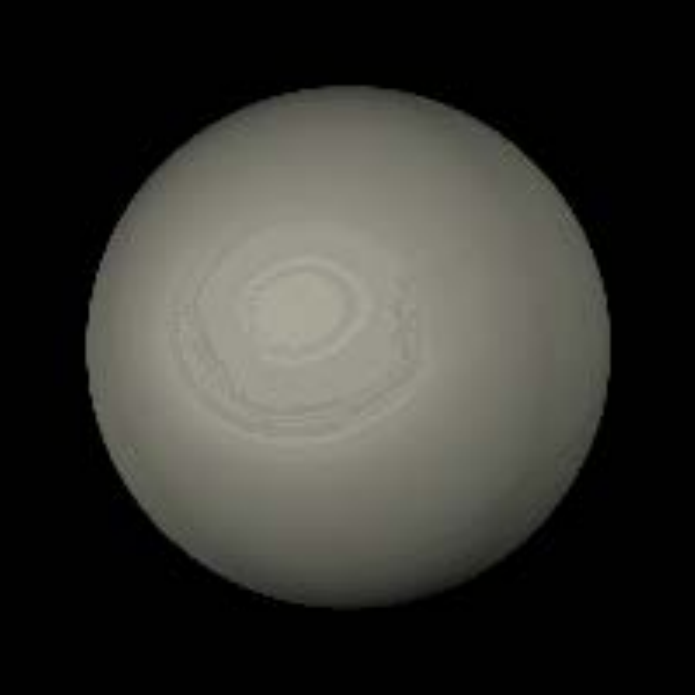}\\
      \includegraphics[width=0.98\linewidth]{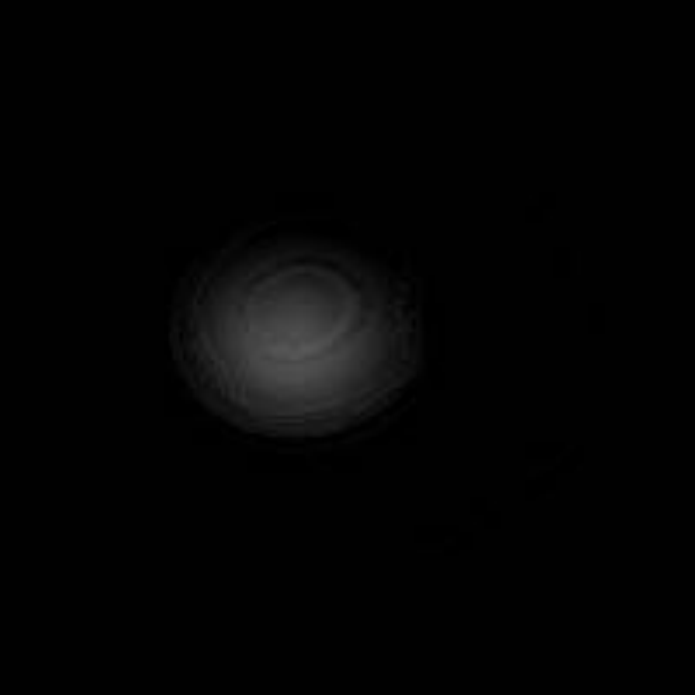}
 \end{minipage}
\begin{minipage}[ht]{0.14\textwidth}
      \includegraphics[width=0.98\linewidth]{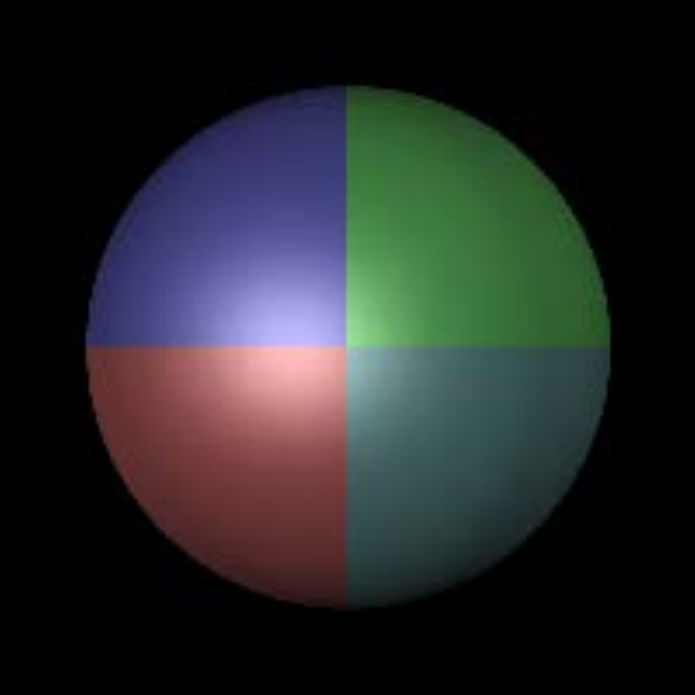}\\\vspace{-3mm}
      \includegraphics[width=0.98\linewidth]{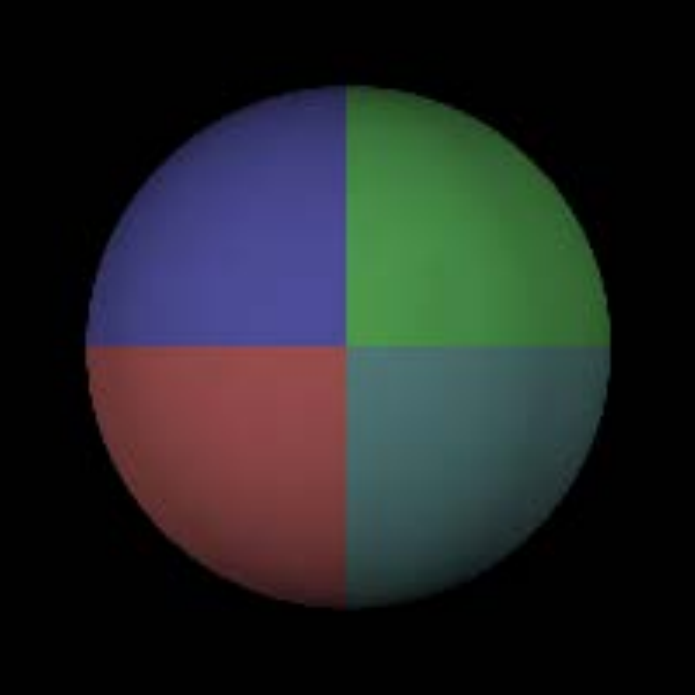}\\
      \includegraphics[width=0.98\linewidth]{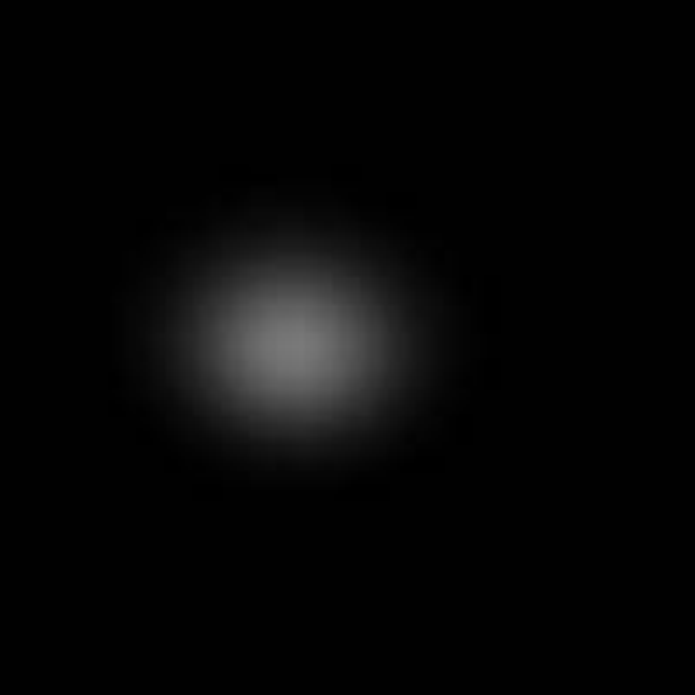}\\\vspace{-3mm}
      \includegraphics[width=0.98\linewidth]{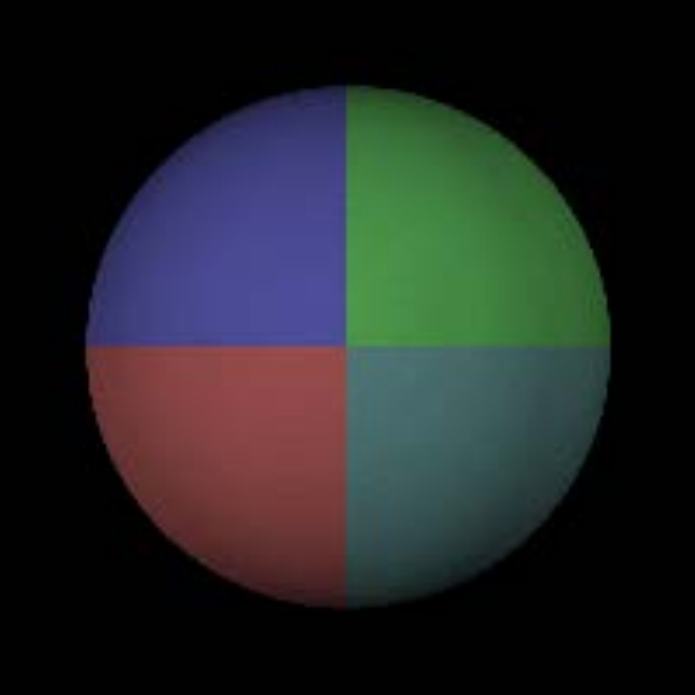}\\
      \includegraphics[width=0.98\linewidth]{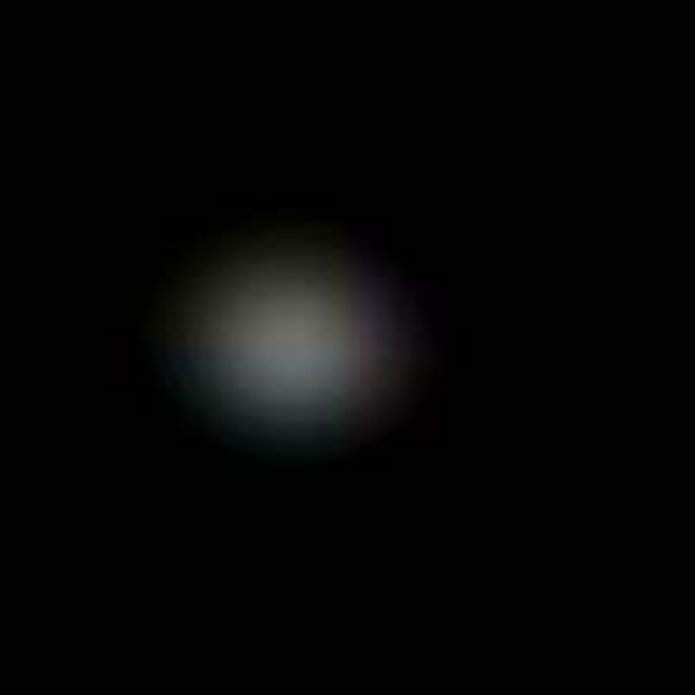}\\\vspace{-3mm}
      \includegraphics[width=0.98\linewidth]{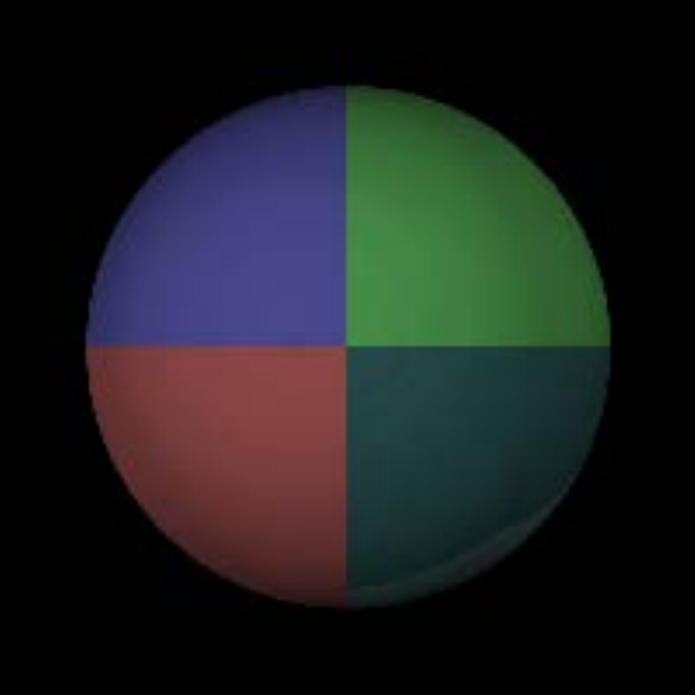}\\
      \includegraphics[width=0.98\linewidth]{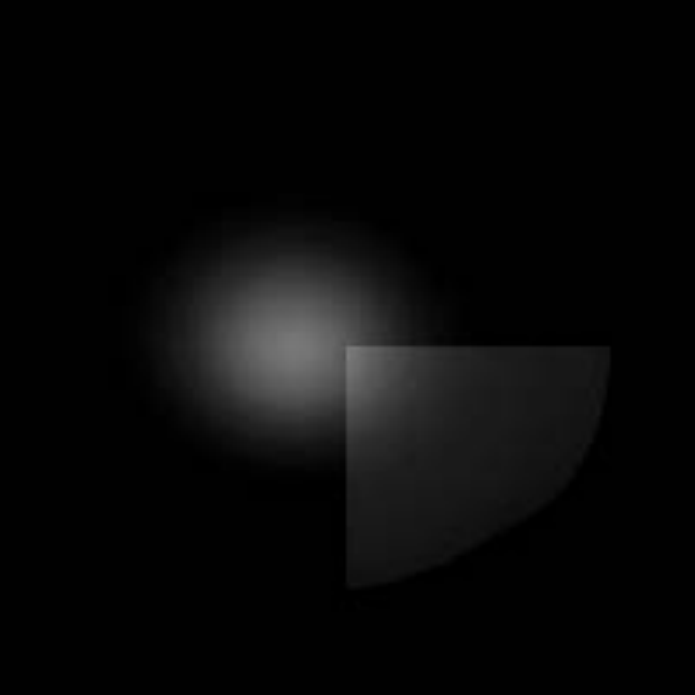}\\\vspace{-3mm}
      \includegraphics[width=0.98\linewidth]{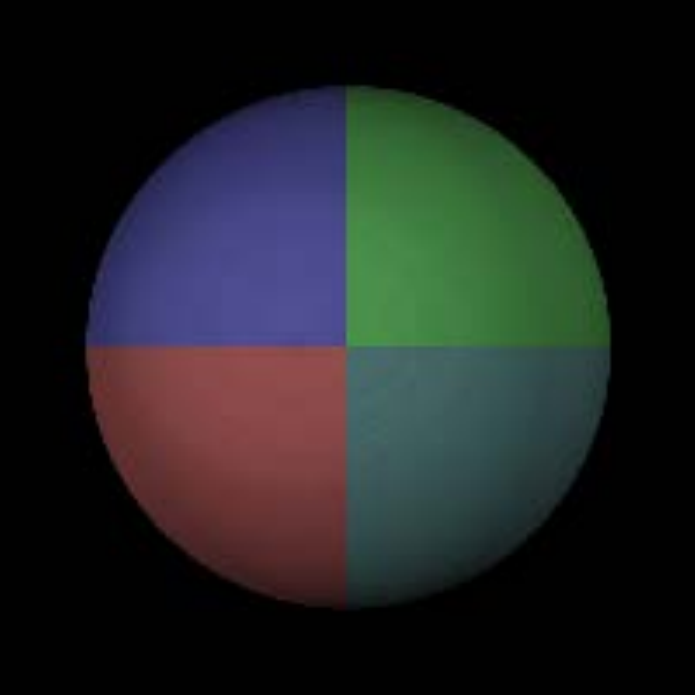}\\
      \includegraphics[width=0.98\linewidth]{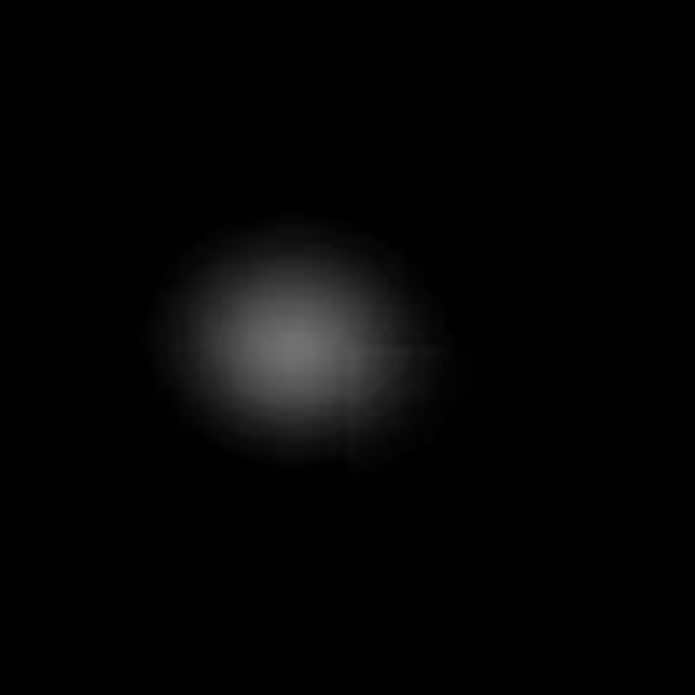}
 \end{minipage}
\begin{minipage}[ht]{0.187\textwidth}
      \includegraphics[width=0.98\linewidth]{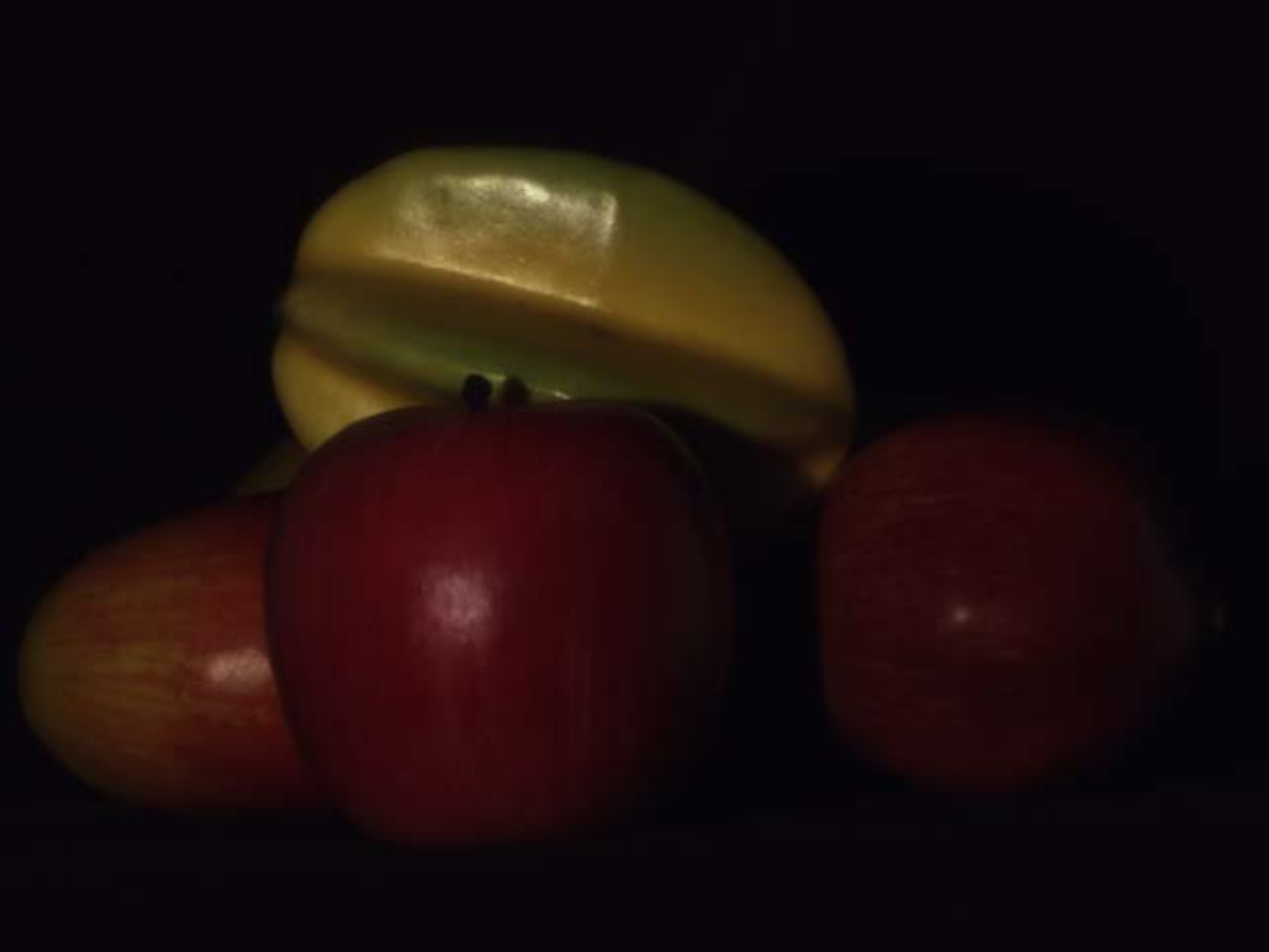}\\\vspace{-3mm}
      \includegraphics[width=0.98\linewidth]{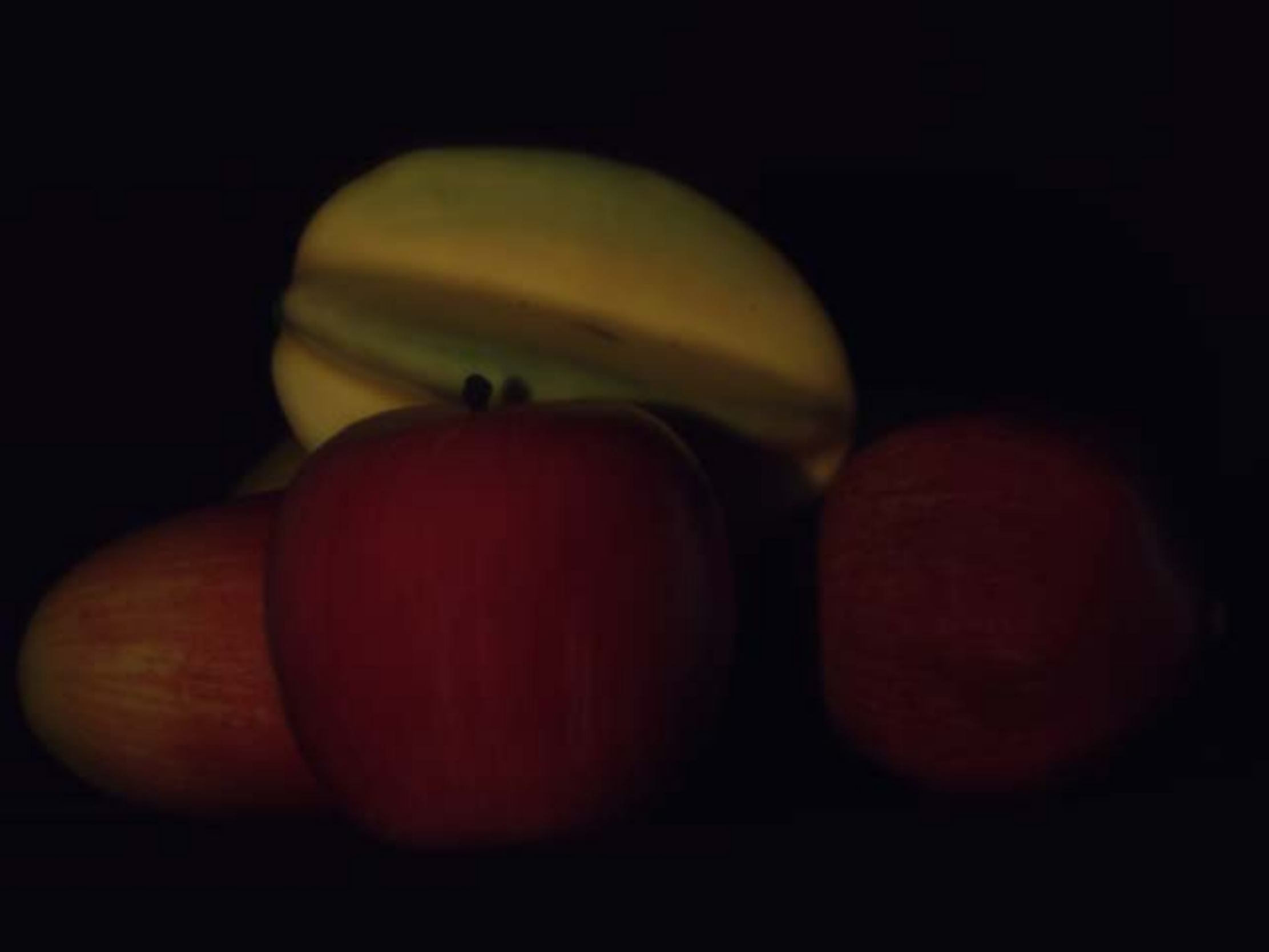}\\
      \includegraphics[width=0.98\linewidth]{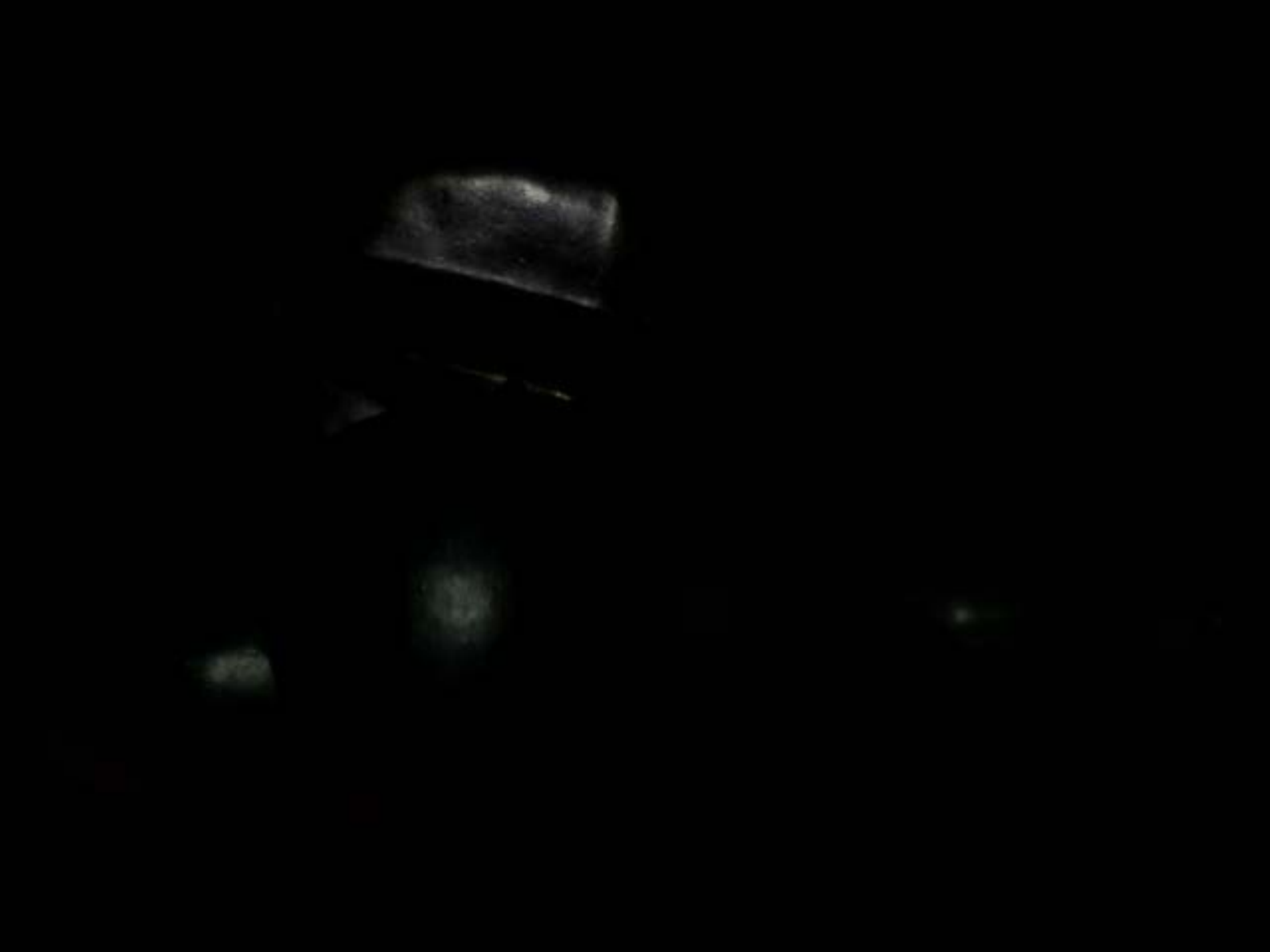}\\\vspace{-3mm}
      \includegraphics[width=0.98\linewidth]{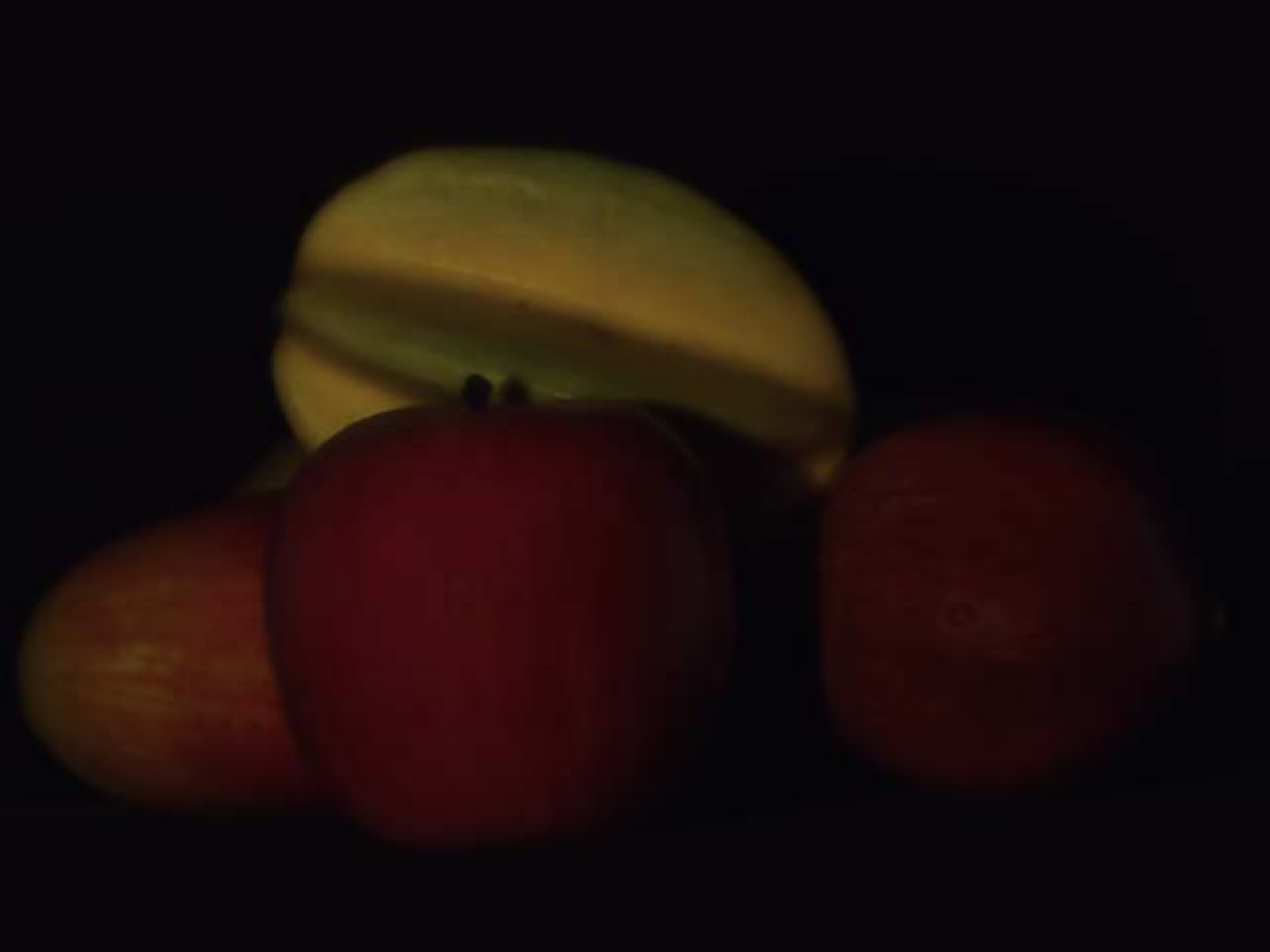}\\
      \includegraphics[width=0.98\linewidth]{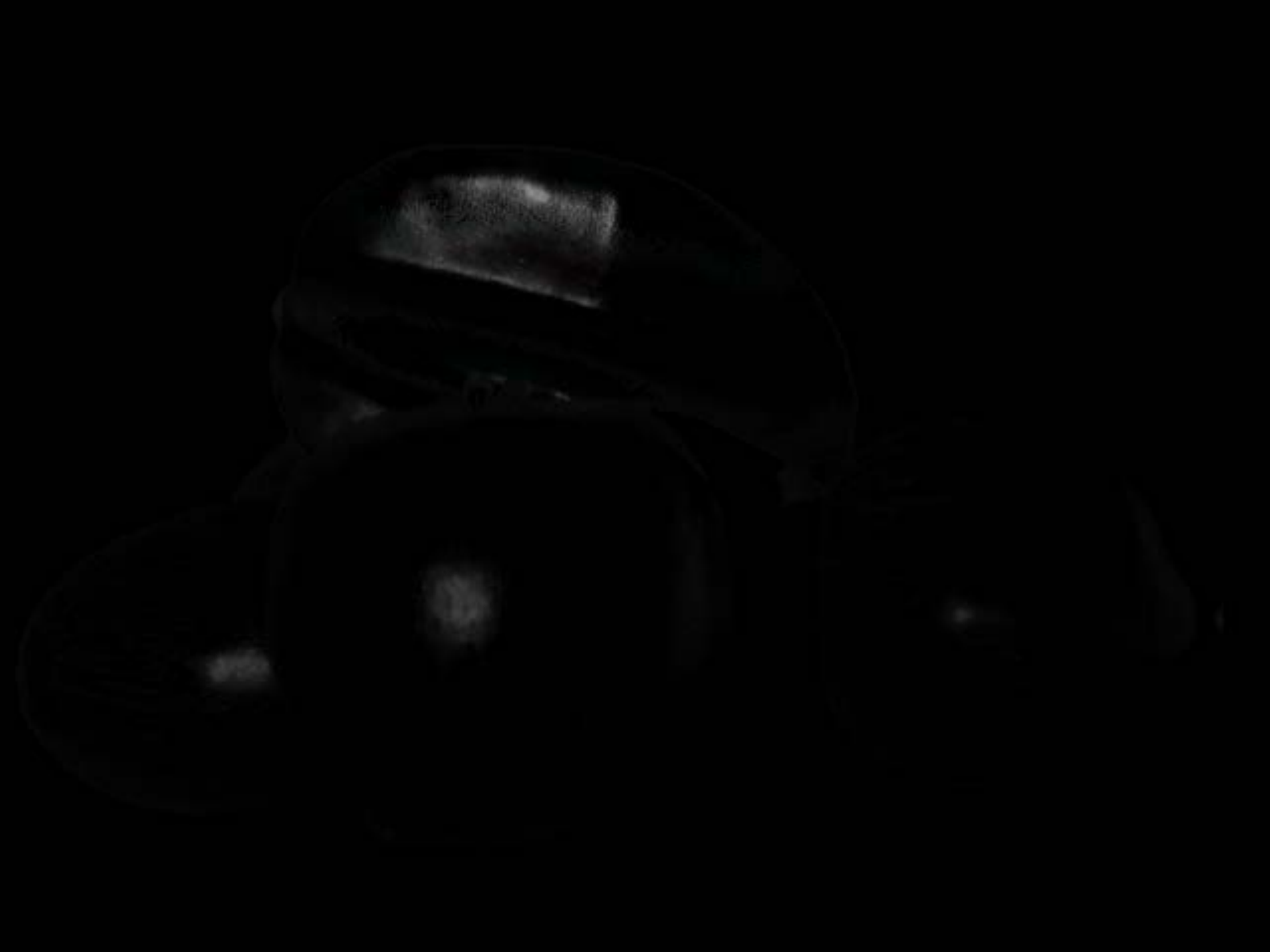}\\\vspace{-3mm}
      \includegraphics[width=0.98\linewidth]{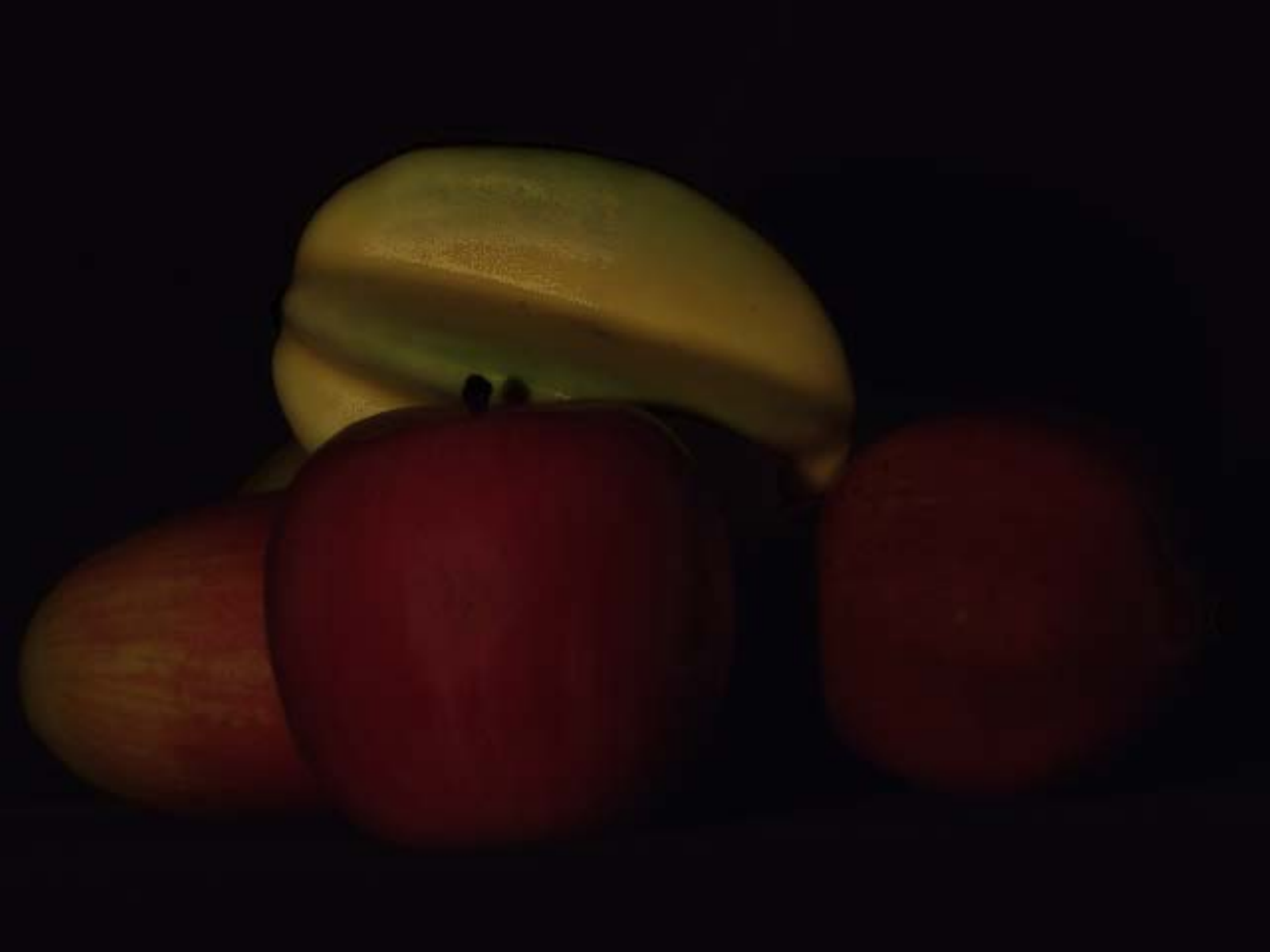}\\
      \includegraphics[width=0.98\linewidth]{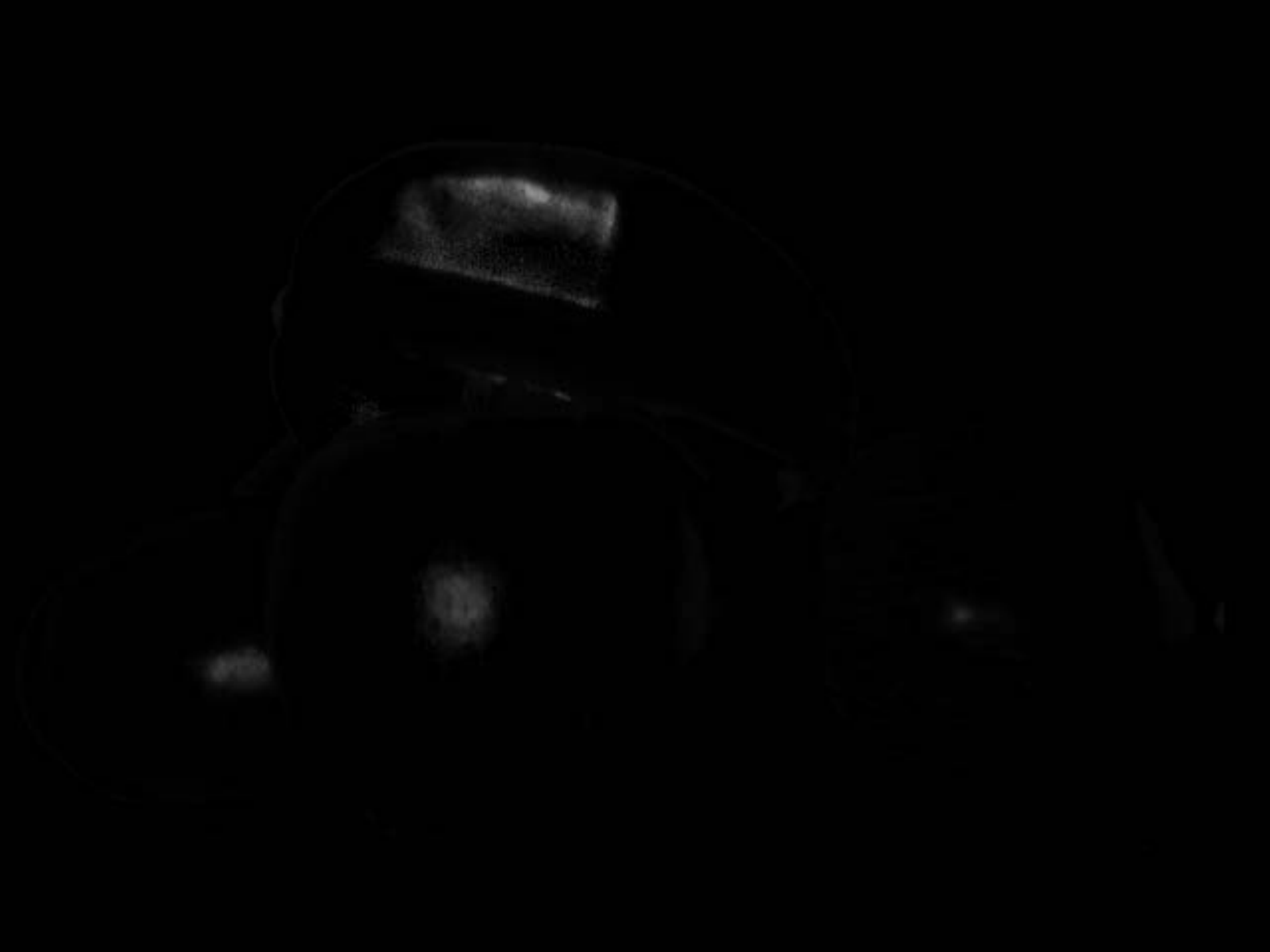}\\\vspace{-3mm}
      \includegraphics[width=0.98\linewidth]{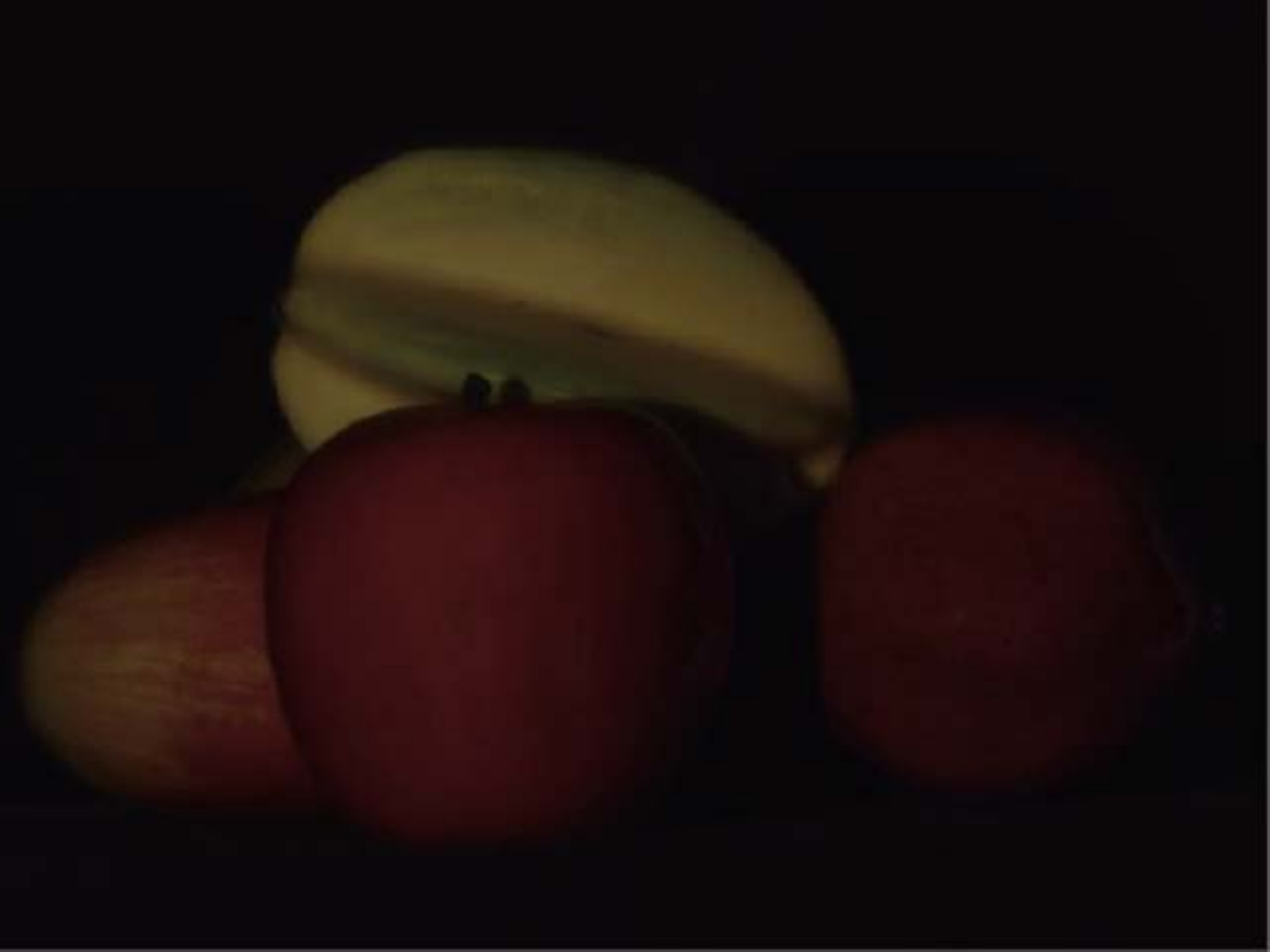}\\
      \includegraphics[width=0.98\linewidth]{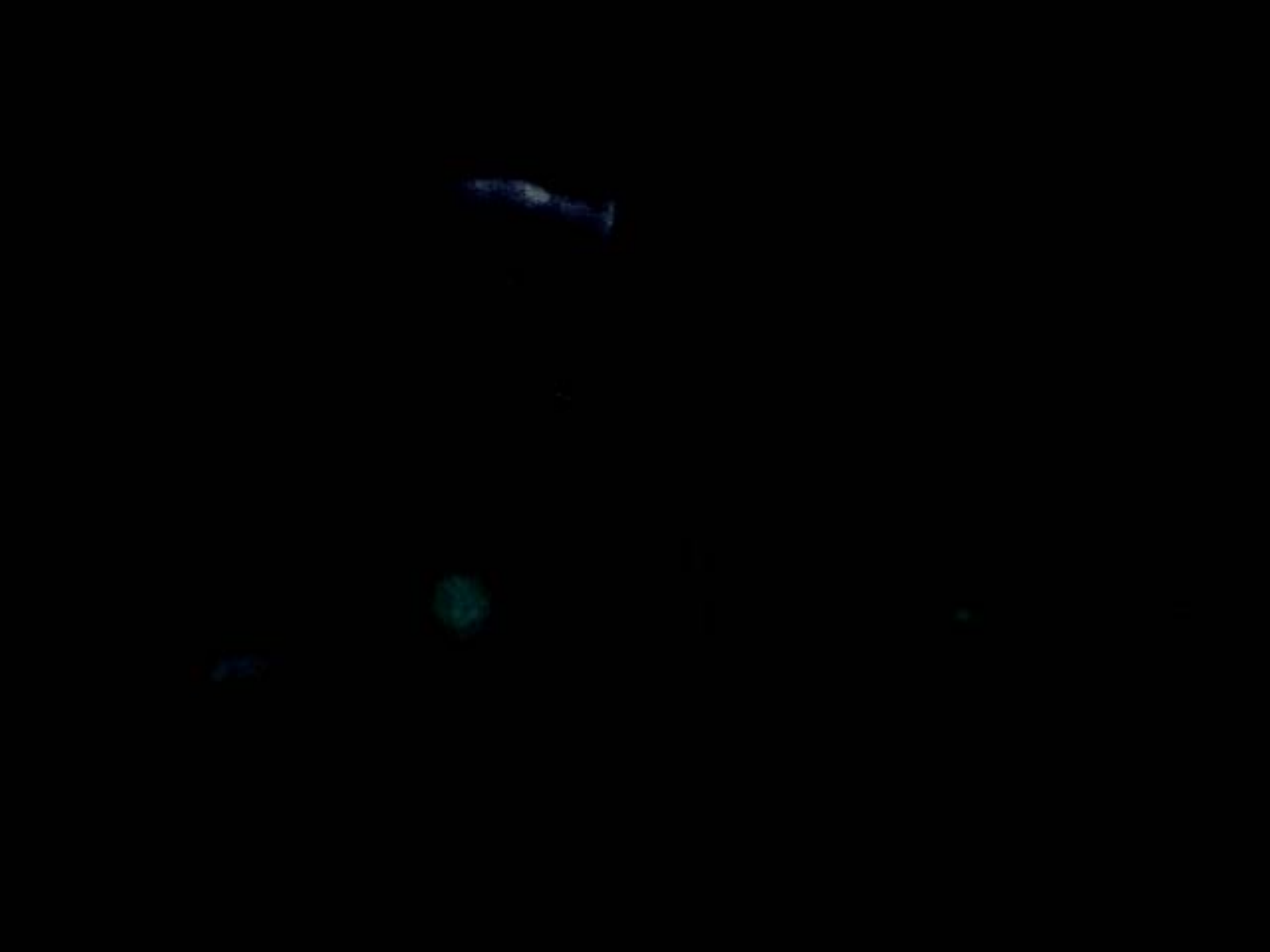}
 \end{minipage}
\begin{minipage}[ht]{0.156\textwidth}
      \includegraphics[width=0.98\linewidth]{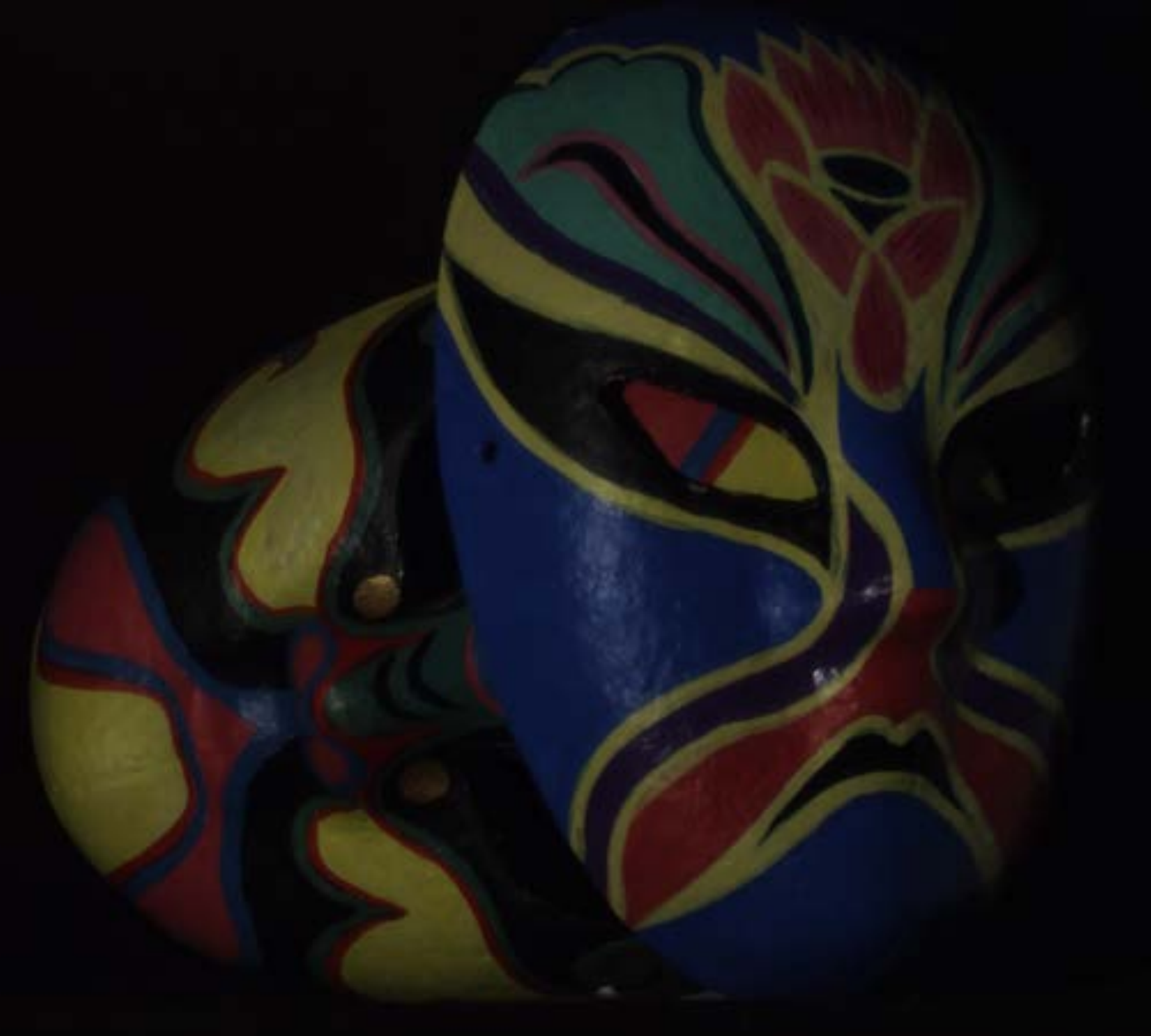}\\\vspace{-3mm}
      \includegraphics[width=0.98\linewidth]{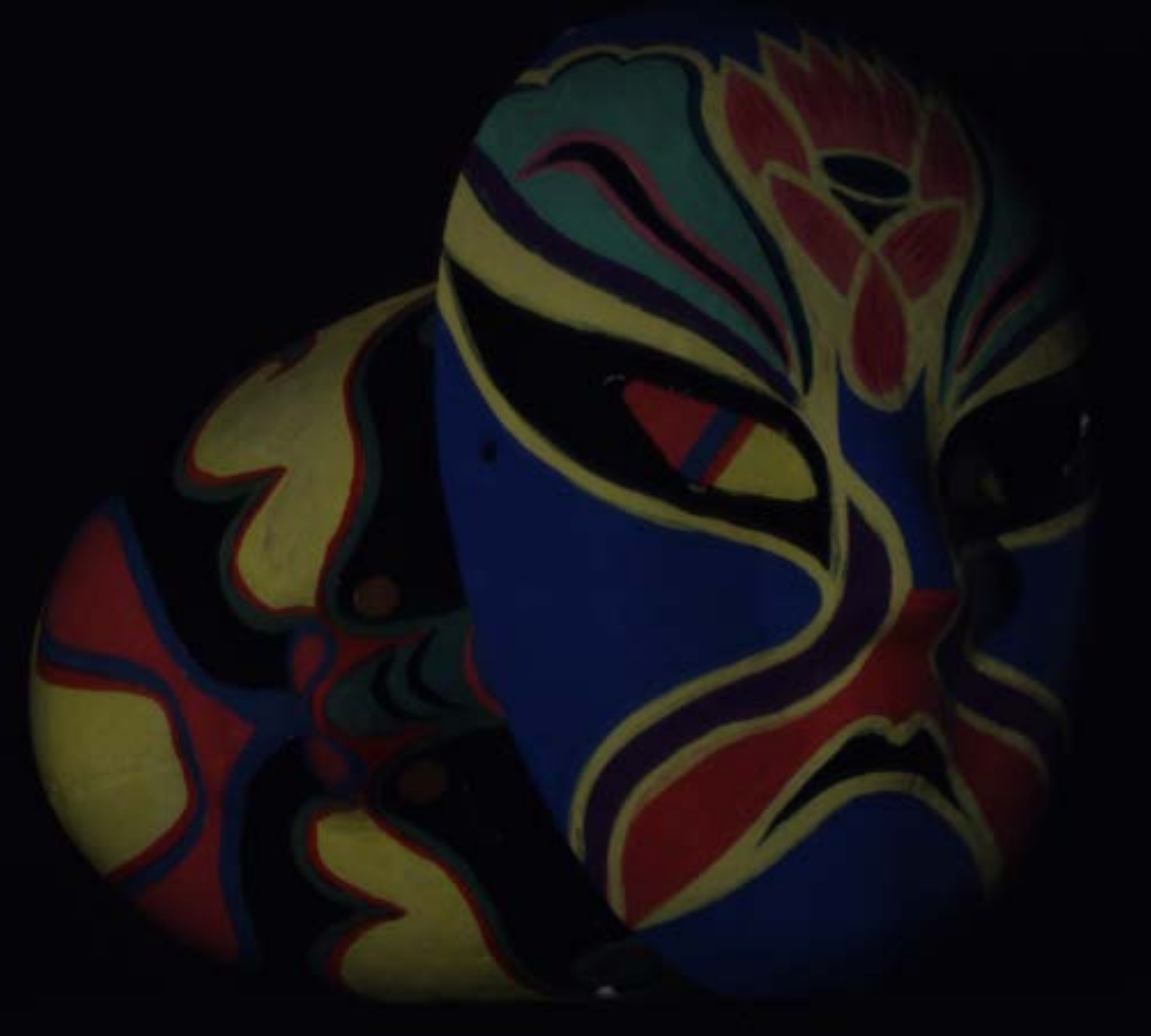}\\
      \includegraphics[width=0.98\linewidth]{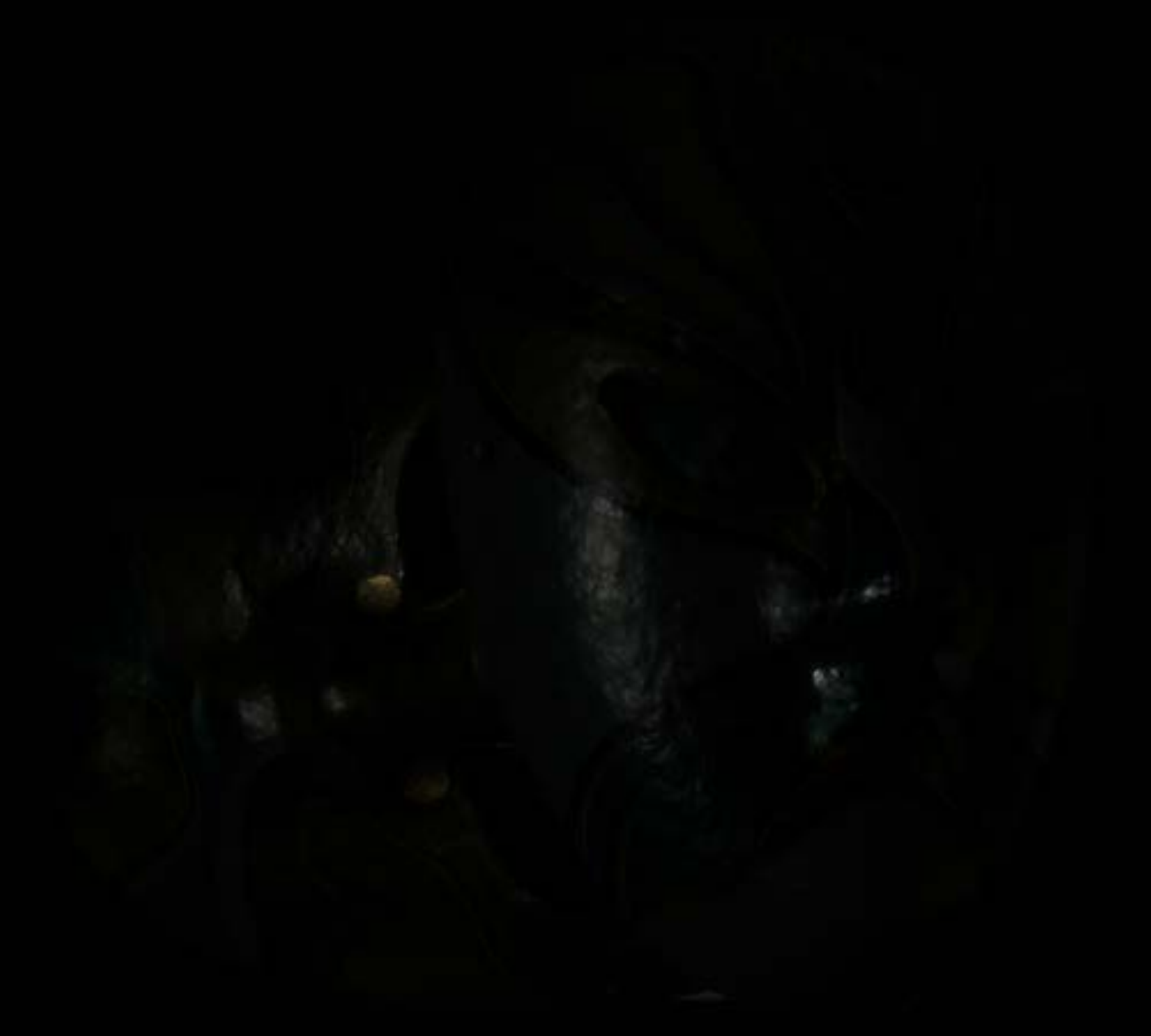}\\\vspace{-3mm}
      \includegraphics[width=0.98\linewidth]{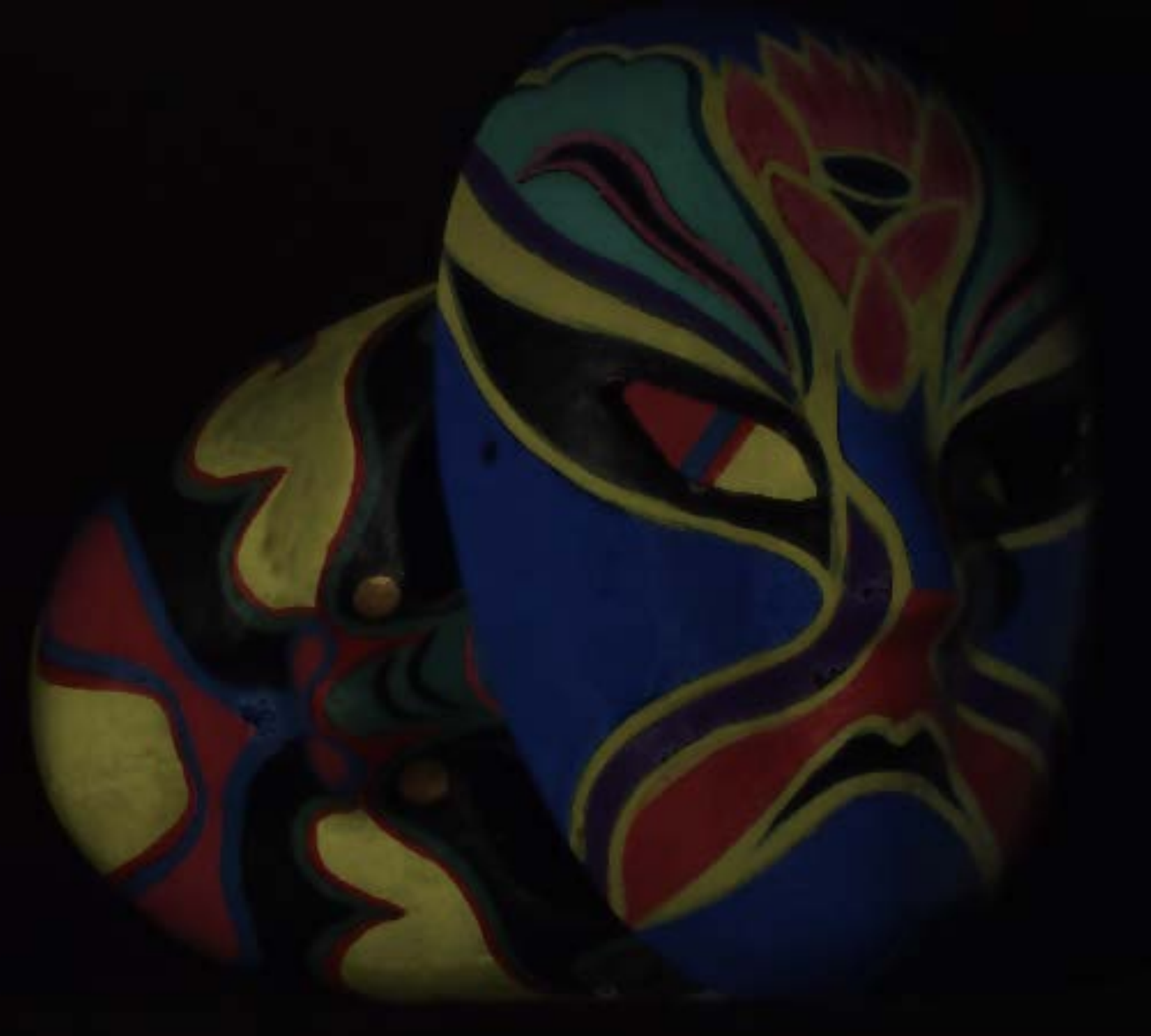}\\
      \includegraphics[width=0.98\linewidth]{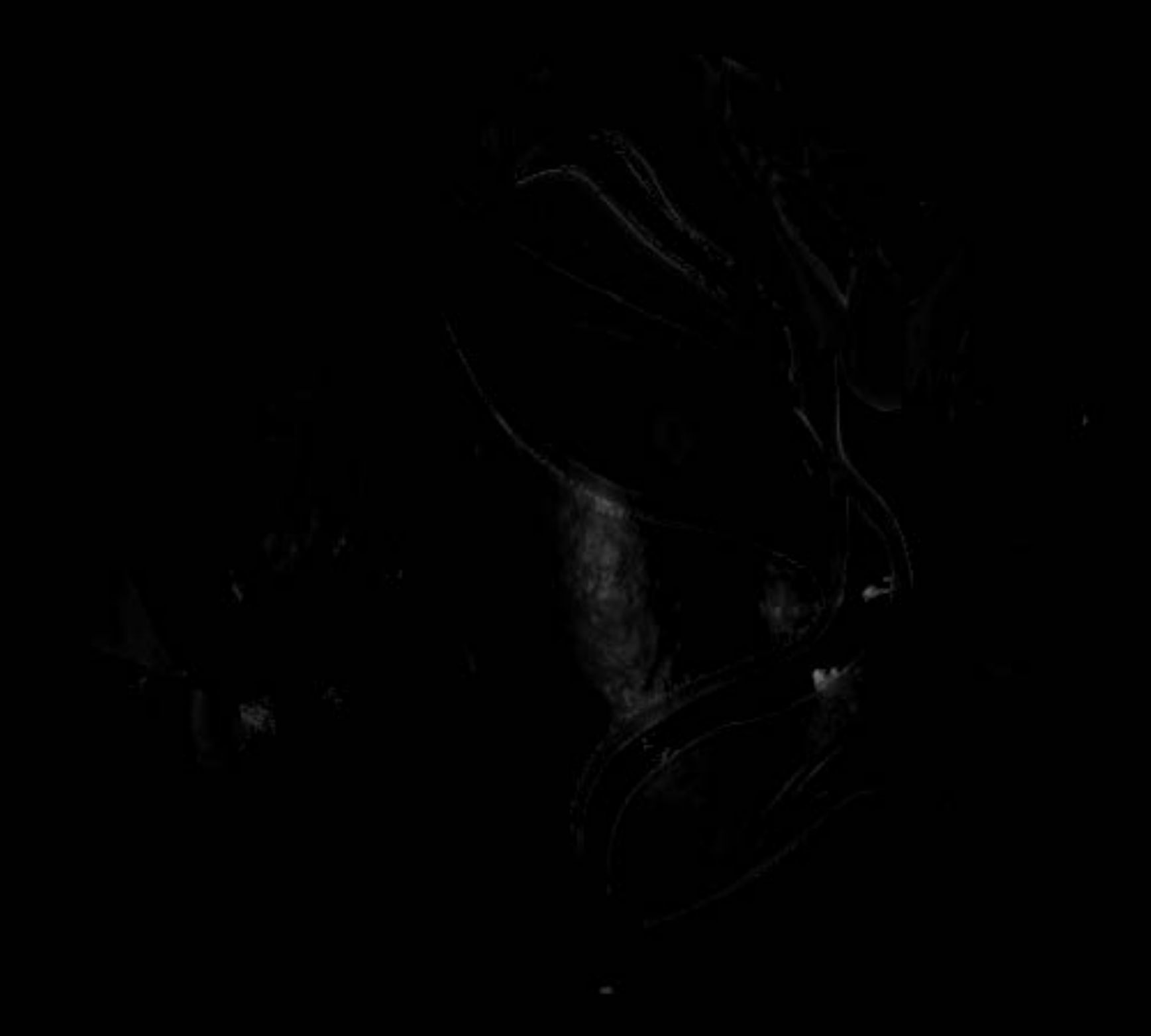}\\\vspace{-3mm}
      \includegraphics[width=0.98\linewidth]{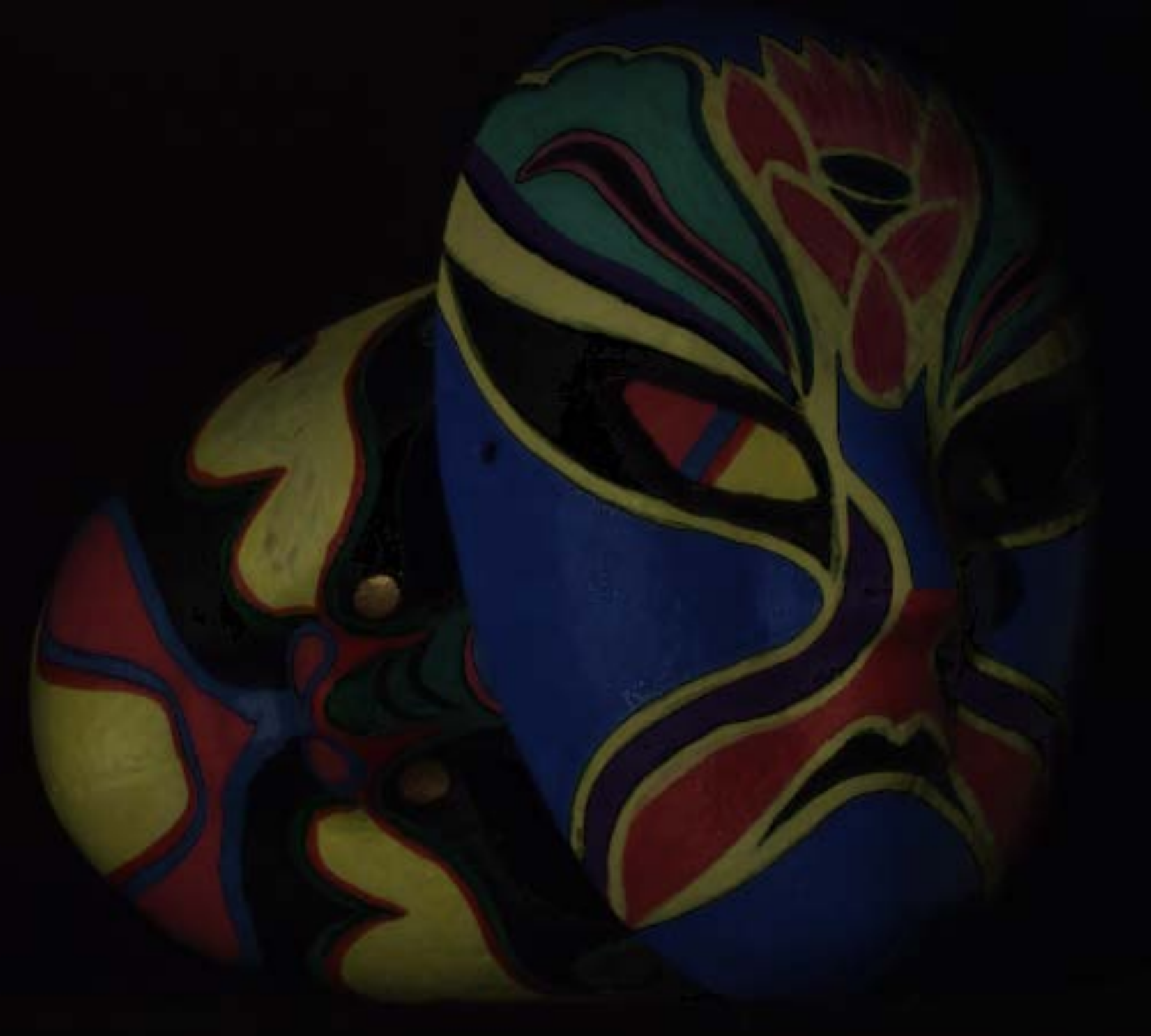}\\
      \includegraphics[width=0.98\linewidth]{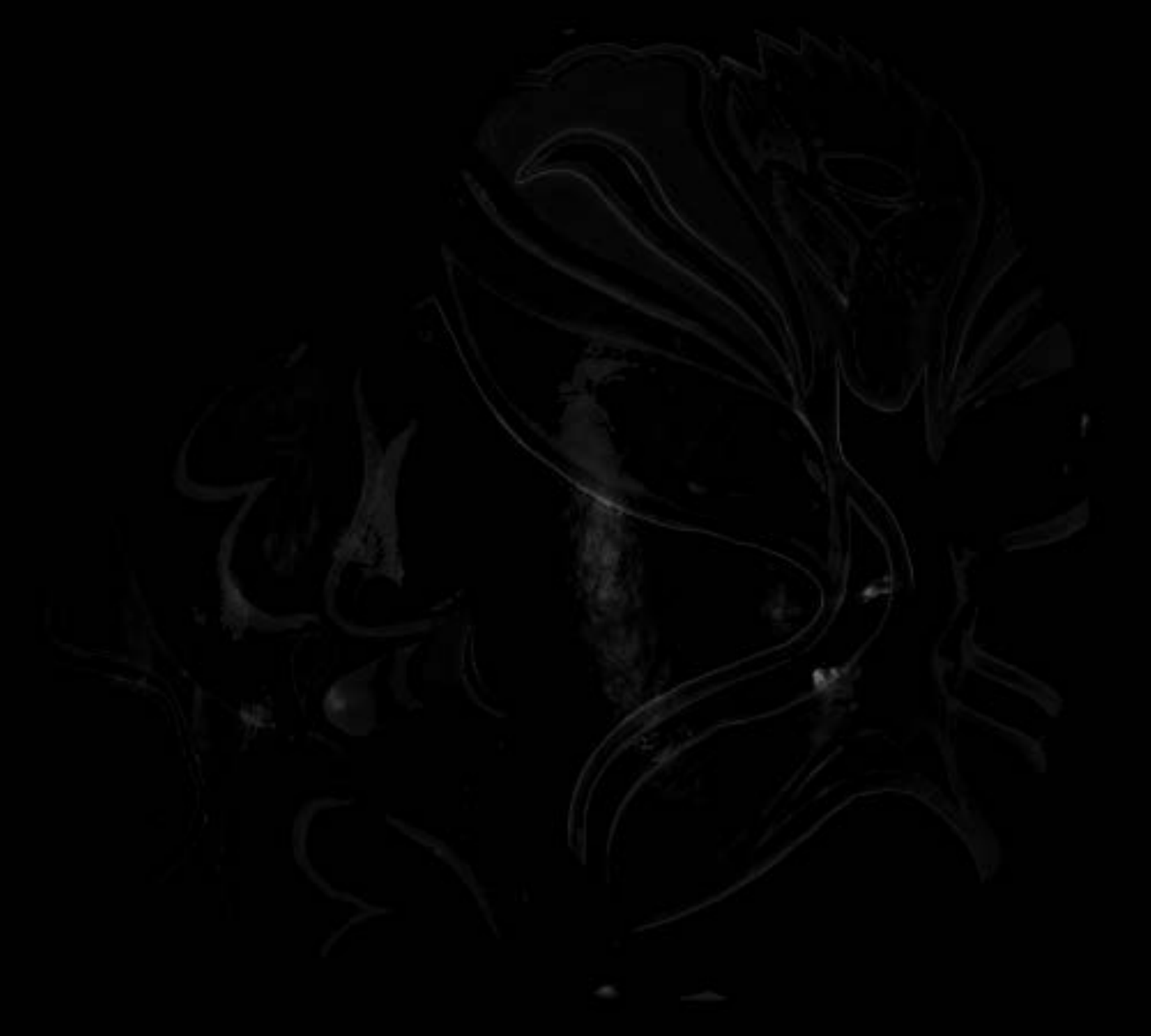}\\\vspace{-3mm}
      \includegraphics[width=0.98\linewidth]{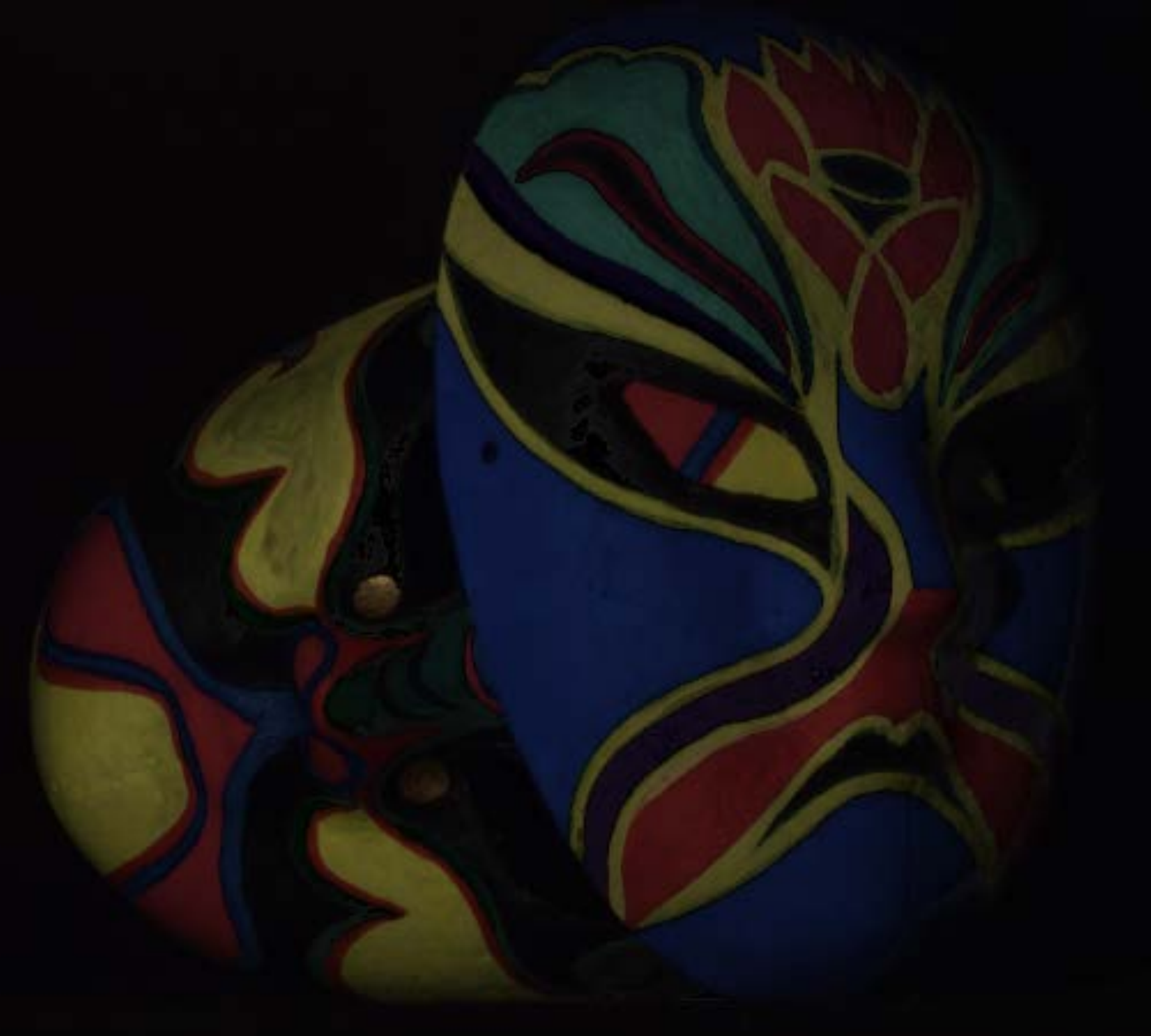}\\
      \includegraphics[width=0.98\linewidth]{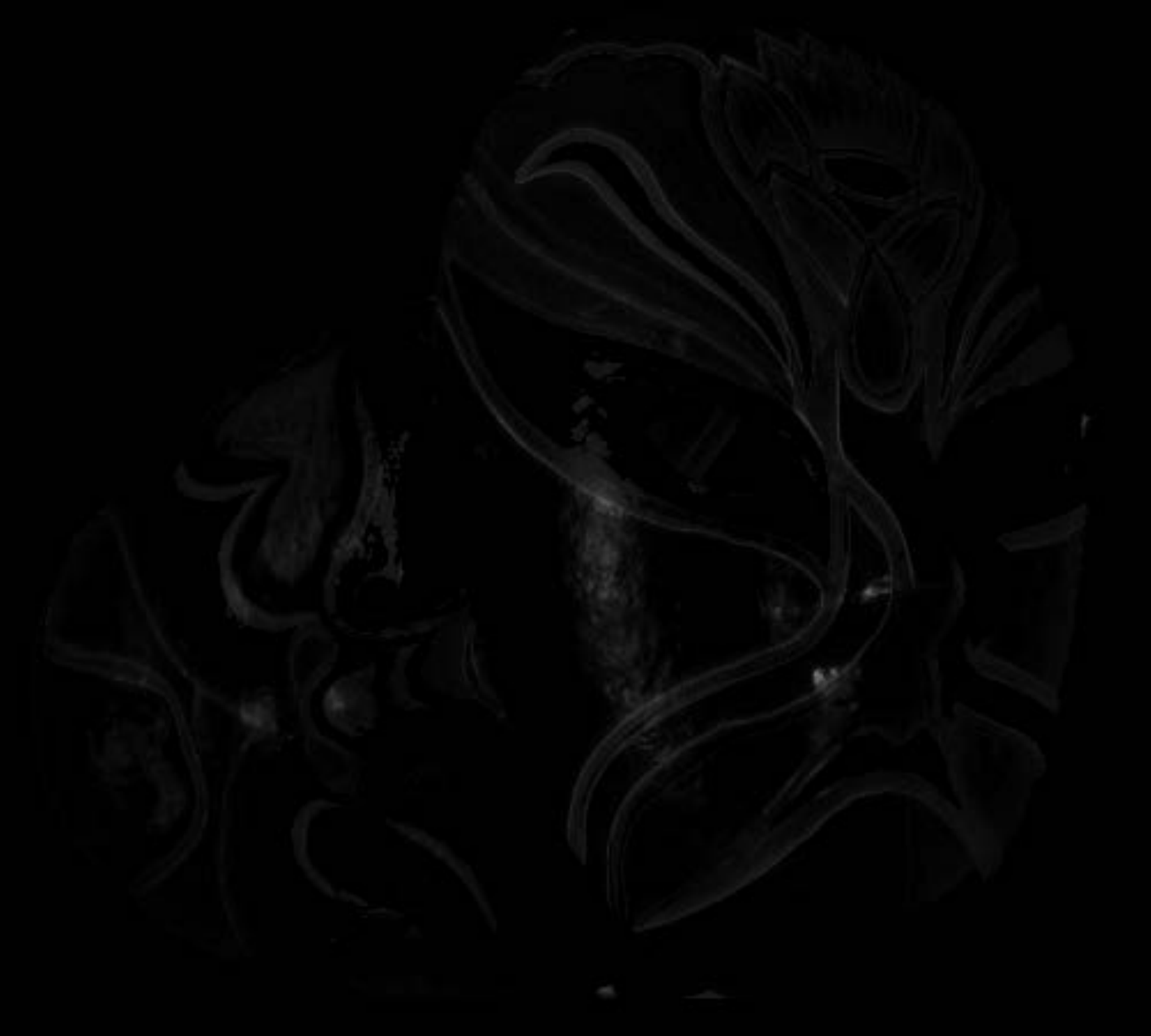}
 \end{minipage}
  \caption{Quantitative comparison between our method and \cite{Shen13} \cite{Yang01} on images with single diffuse color (1st-3rd columns) and color textures (4th-6th columns). 
  (a) Input images. (b) The ground truths. (c)(d)(e) The results of our method, Shen et al.\cite{Shen13}'s and Yang et al.\cite{Yang01}'s.}
  \label{fig:synth_singlecolor}
\end{figure*}

\subsection{PDDR based Diffuse Component Recovering}
Theoretically, the pure diffuse pixels in each cluster will be those with the minimum value of $\beta$ ($\beta=0$), which is corresponding to the smallest parallel component $\gamma({\bf{x}})=\alpha({\bf{x}})b({\bf{x}})+\beta({\bf{x}})$. Thus, the leftmost pixels in each circle arc in Fig.~\ref{fig:hr_remove}(e) are the pure diffuse pixels, { as illustrated in Fig.~\ref{fig:notation}}. However, there is always noise in real cases and we treat the pixels falling around the first peak of $\gamma({\bf{x}})'s$ histogram as purely diffuse, i.e., ${\widetilde{\bf{I}}}({\bf x})={\Lambda}({\bf{x}})$. The histograms of $\gamma({\bf{x}})$s for the four regions in Fig.~\ref{fig:hr_remove}(a) are plotted in Fig.~\ref{fig:hr_remove}(f), with the solid dots denoting the coefficients corresponding to the pure diffuse reflections. 

After finding the pure diffuse pixels in each cluster, it is easy to remove the specular highlight. From Eq.~(\ref{eqs:d_decom}), the problem of recovering the diffuse component at ${\bf x}$ is equivalent to find the coefficients $\alpha({\bf x})a({\bf x})$ along $\Gamma^\bot({\bf{x}})$ and $\alpha({\bf x})b({\bf x})$ along $\Gamma$. Since $\gamma^\bot({\bf x})=\alpha({\bf x})a({\bf x})$ can be calculated by directly projecting $\bf I({\bf x})$ onto $\Gamma^\bot({\bf{x}})$, the problem can be further simplified into finding the ratio of $a({\bf{x}})$ to $b({\bf{x}})$. On one hand, for the pixels with the same diffuse component, this ratio is invariant to the specular strength. On the other hand, the ration of $a({\bf{x}})$ to $b({\bf{x}})$ is equivalent to the ratio of $\gamma^{\bot}({\bf{x}})$ to $\gamma({\bf{x}})$ for the pure diffuse pixels with $\beta({\bf x})=0$. Thus, the ratio of $a({\bf{x}})$ to $b({\bf{x}})$ in each cluster can be directly calculated with the known pure diffuse pixels.



\section{Experiments and Analysis}
To validate the proposed method, a series of experiments are conducted. Firstly, we quantitatively test our method on both synthetic and real data with ground truths, and demonstrate its higher performance than previous methods. Then, we run our algorithm on several images including non-Lambertian reflection and provide comparison with several state-of-the-art (STAR) methods. Next, considering that above experiments both assume known illumination color, and the influence from improper clustering is not considered either, we perform two experiments to test the robustness of our approach to inaccurate illumination color and clustering. Lastly, we discuss the efficiency of our approach, in comparison with other two existing methods achieving near realtime processing on VGA images.

For performance comparison with STAR, we compare with the most cited method \cite{Yang01}, the newest one \cite{Akashi14}, and one also based on material clustering \cite{Shen13}. Since the authors of \cite{Akashi14} only provide the images on real data, we do not include their results in synthetic experiments. Neither source code nor running results of \cite{Hyeongwoo01} is available, so the comparison is omitted.

In terms of parameter setting, we only need to set the threshold for stopping cluster subdivision. In implementation, we subdivide a cluster when more than 10\% pixels within this cluster deviate more than 0.1 from the unit circle.
In order to be robust to sensor noise, we force the pixel number within each cluster larger than 300.

\begin{table*}[t]
\centering
\caption{\label{tabl:multicolor_noise} The PSNRs (dB) of recovered diffuse components on the synthetic data displayed in 4th-6th columns of Fig.~\ref{fig:synth_singlecolor}.}
\begin{tabular}{p{3cm}<{\centering}|p{2cm}<{\centering}|p{3.4cm}<{\centering}|p{3.4cm}<{\centering}|p{3.4cm}<{\centering}p{3.4cm}<{\centering}}
\hline
Scenes & $\sigma$ &The proposed & Shen et al.\cite{Shen13} & Yang et al.\cite{Yang01}\\
\hline
\multirow{3}{*}{Synth} & 0 & {\bf51.4} & 29.8 & 39.0\\
& 3 & {\bf34.4} & 29.3 & 34.3\\
& 6 & {\bf32.4} & 27.5 & 29.4\\
\hline
\multirow{3}{*}{Fruits} & 0 & {\bf40.4} & 38.9 & 37.6\\
& 3 & {\bf37.4} & 35.5 & 34.0\\
& 6 & {\bf35.1} & 31.8 & 30.1\\
\hline
\multirow{3}{*}{Masks} & 0 & {\bf34.2} & 34.1 & 32.2\\
& 3 & {\bf33.0} & 32.5 & 29.8\\
& 6 & {\bf32.0} & 29.9 & 27.5\\
\hline
\end{tabular}
\end{table*}

\subsection{Quantitative Evaluation}


Separating diffuse and specular components of a close-to-white surface is challenging. To illustrate the advantage of our method in this situation, we synthesize images with diffuse chromaticity being [0.7053 0.7053 0.0705], [0.6667 0.6667 0.3333] and [0.5965 0.5965 0.5369] respectively, and the illumination chromaticity being [0.5774, 0.5774, 0.5774], as shown in 1-3 columns of Fig.~\ref{fig:synth_singlecolor}. Although behaving well in first two columns, \cite{Shen13} cannot cope with the scene with diffuse color approaching white, i.e., the 3rd column, due to the approximation in the criterion for recovering the diffuse color. Yang et al.\cite{Yang01} give plausible results, but the separated specular component tends to be weaker than the ground truth. In comparison, our approach demonstrates best performance. This is mainly due to that we strictly follow the dichromatic reflection model, while the assumptions made by \cite{Shen13} and \cite{Yang01}  are violated in such cases.

\begin{figure}[t]
\centering
\subfigure[]{
\begin{minipage}[ht]{0.46\textwidth}
    \includegraphics[width=0.48\linewidth]{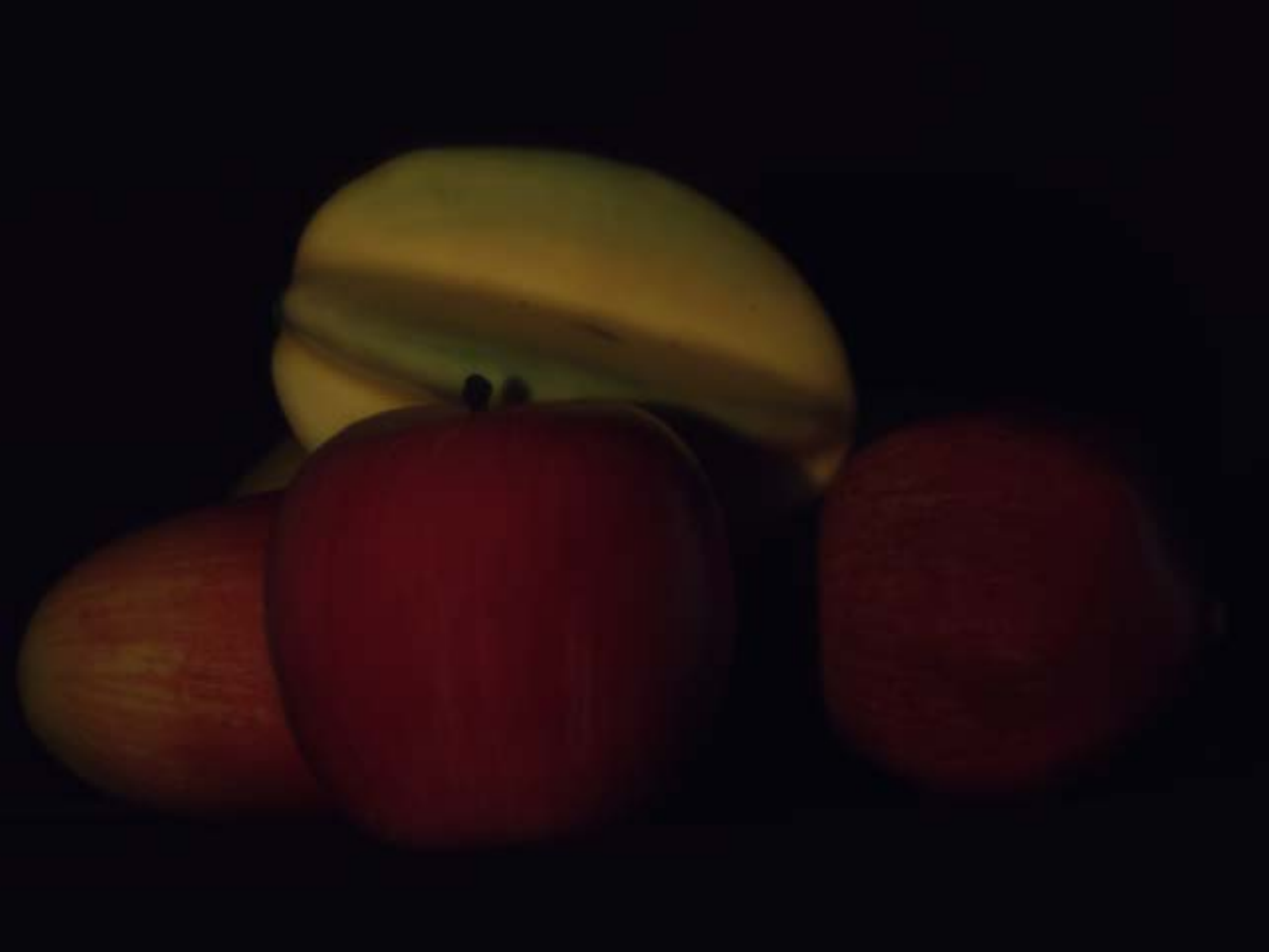}
    \includegraphics[width=0.48\linewidth]{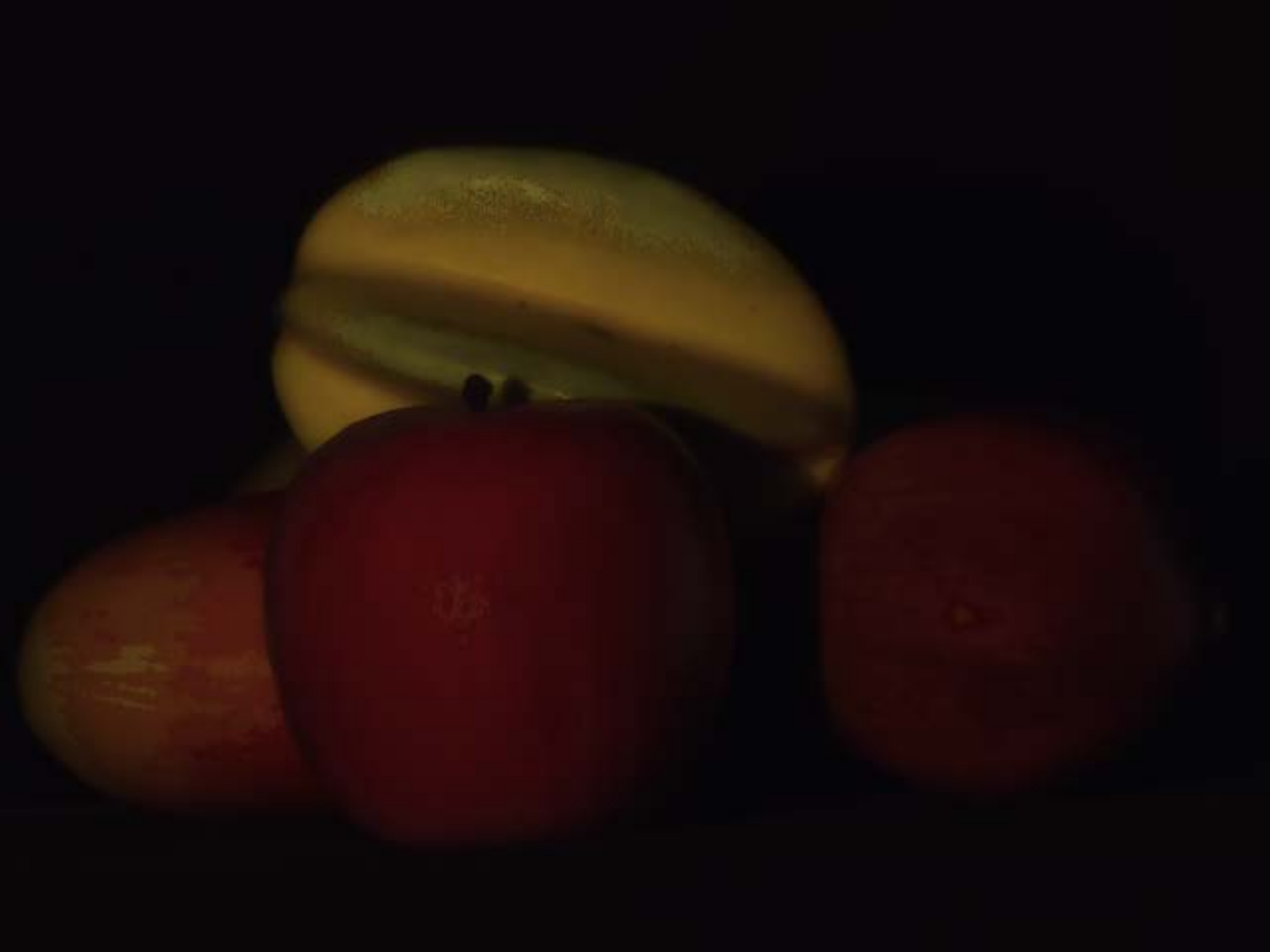}\\
    \includegraphics[width=0.48\linewidth]{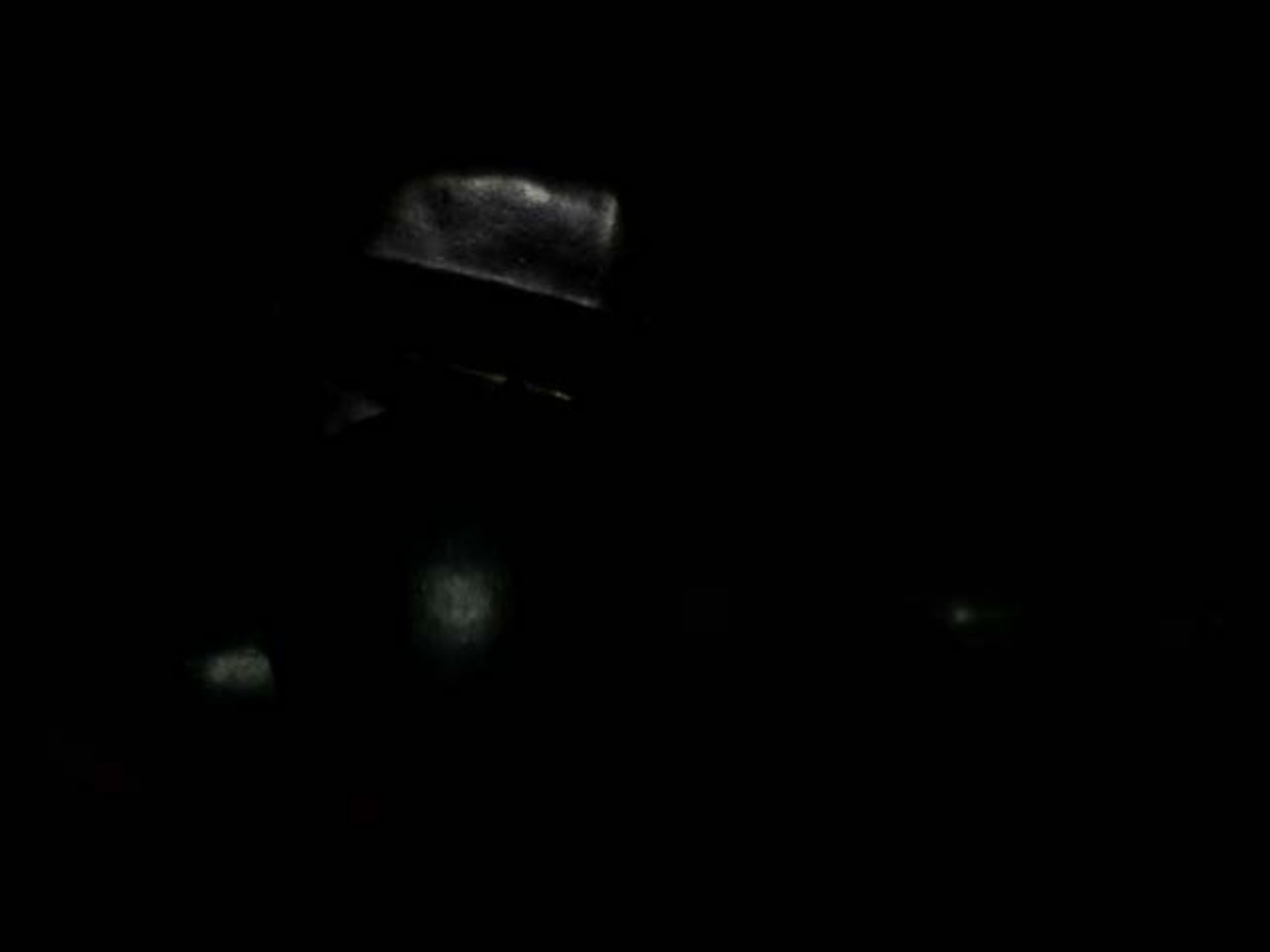}
    \includegraphics[width=0.48\linewidth]{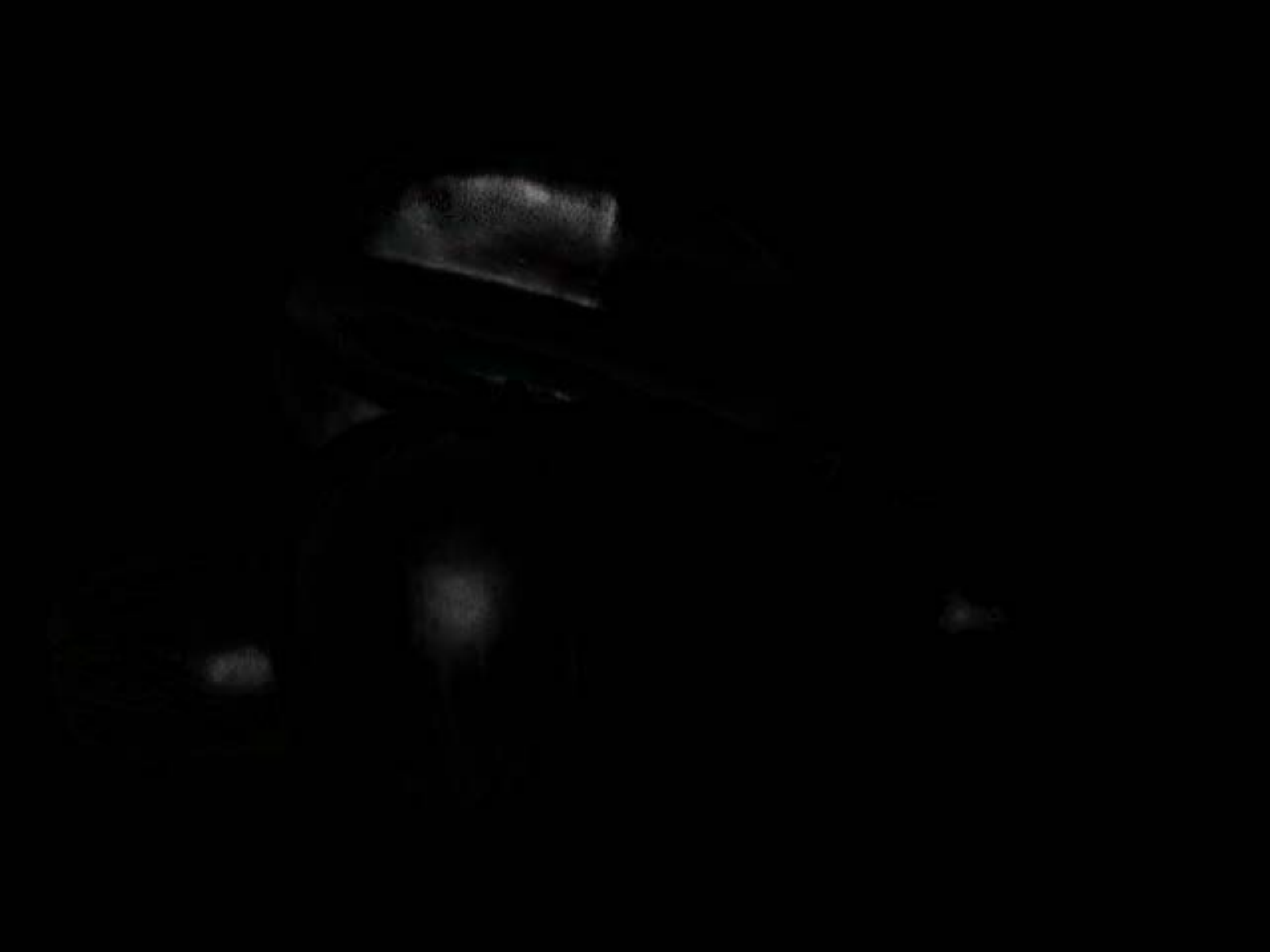}\\\vspace{-1mm}
 \end{minipage}}
\subfigure[]{
\begin{minipage}[ht]{0.46\textwidth}
 \includegraphics[width=0.48\linewidth]{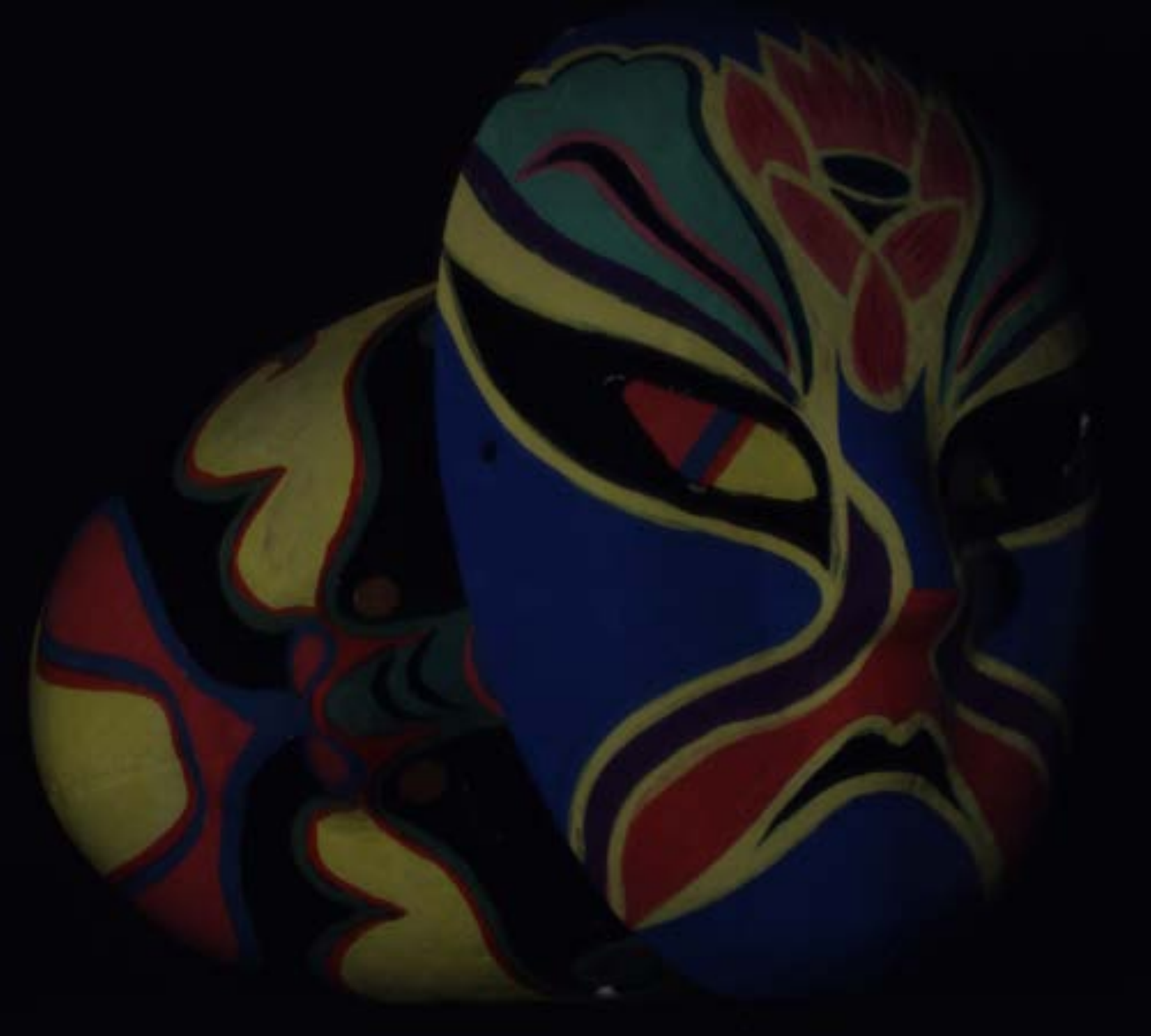}
 \includegraphics[width=0.48\linewidth]{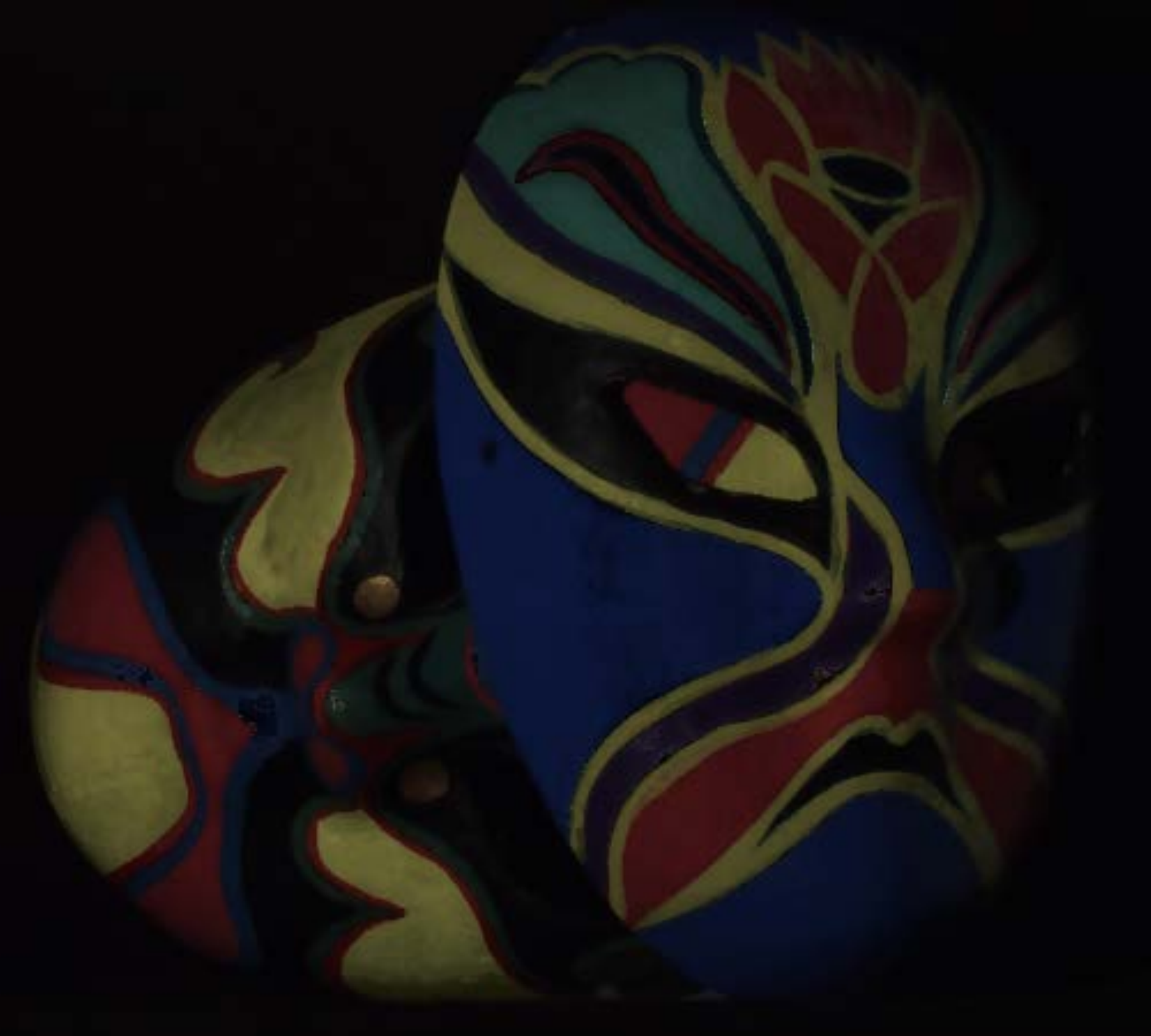}\\
    \includegraphics[width=0.48\linewidth]{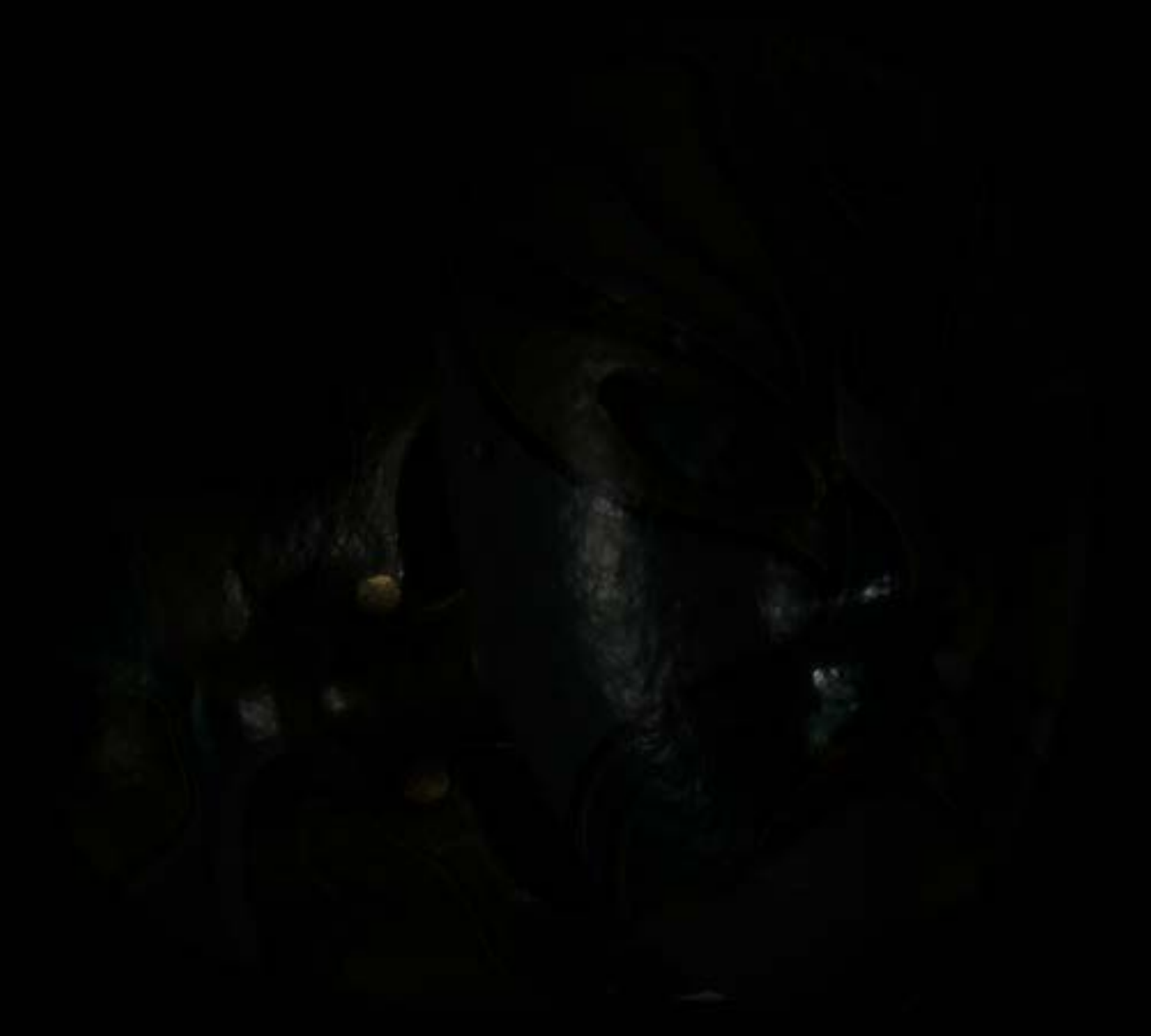}
    \includegraphics[width=0.48\linewidth]{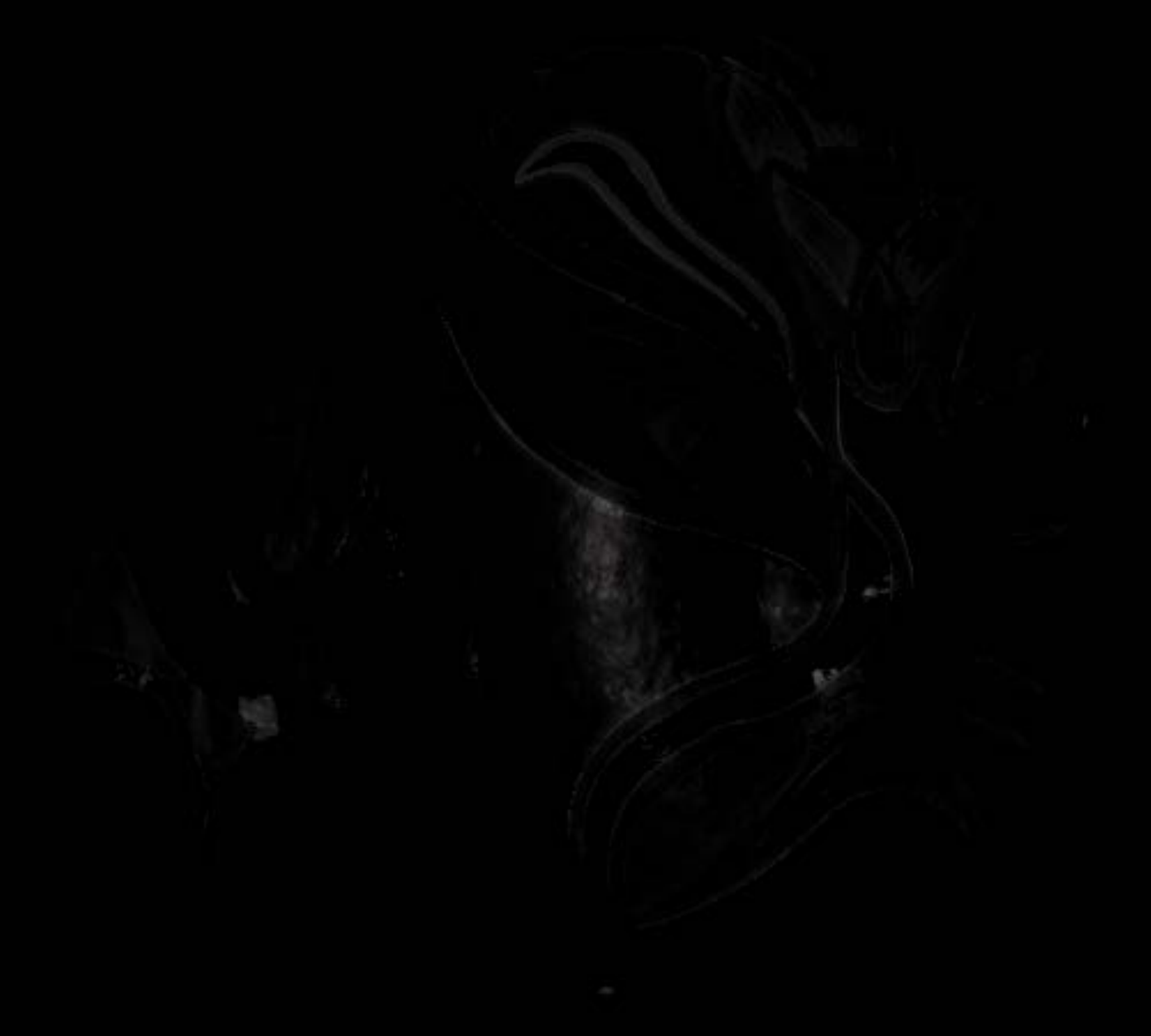}\\\vspace{-1mm}
 \end{minipage}}
  \caption{The separated diffuse and specular components of the last two scenes shown in Fig.~\ref{fig:synth_singlecolor}, with true and estimated illumination color. (a) The 'Fruits' scene. (b) The 'Masks' scene. In each subfigure, the left column shows the results with ground truth illumination, and the right column gives those based on the estimated illumination by the estimator proposed in \cite{Tan01}.}
  \label{fig:synth_illu}
\end{figure}

To test our performance in textured images, we compare it with \cite{Shen13} and \cite{Yang01} using images with ground truth diffuse component, as shown in 4-6 columns of Fig.~\ref{fig:synth_singlecolor}. The 4th column shows a synthetic image, and the other two are from \cite{Yang01}. Since noise always exists in data capturing, we also add Gaussian white noise with standard derivation $\sigma$=3 and 6 to the input images. PSNR is used as evaluation metric, as shown in Table \ref{tabl:multicolor_noise}. The results on the three textured images consistently show our superior performance, and the superiority is more prominent at higher noise levels.

\subsection{Influence of Inaccurate Illumination or Material Clustering}
When accurate illumination color is unavailable, one can also use the estimated illumination by some existing methods, such as \cite{Tan01} and \cite{Drew14}. To test the robustness to estimation deviation of the illumination color, we compare the performance of our highlight removal algorithm on the last two scenes in Fig.~\ref{fig:synth_singlecolor} with the ground truth illumination color and that estimated by Tan et al.\cite{Tan01}. The ground truth illumination color in Fig.~\ref{fig:synth_singlecolor} is pre-calibrated into white (i.e., [r g b] = [0.577 0.577 0.577]), while the estimated illumination for these two scenes are [r g b] = [0.600 0.588 0.542] and [r g b] = [0.633 0.575 0.518], respectively.
The results are shown in Fig.~\ref{fig:synth_illu}. Comparing the results from estimated (left column) and accurate illumination (right column), we can see that the visual performance does not drop a lot.
Quantitatively, the PSNR of the recovered diffuse component of the 'Fruit' scene drops by 2.3 dB (from 40.4 dB to 38.1dB), and that of the 'Mask' scene drops by 1.2 dB (from 34.2 dB to 33.0dB), respectively. The scores arrive at the same conclusion: the deviation of the illumination estimators would not affect our final highlight removal performance apparently.

\begin{figure}[t]
\centering
\subfigure[]{
  \includegraphics[width=0.48\linewidth]{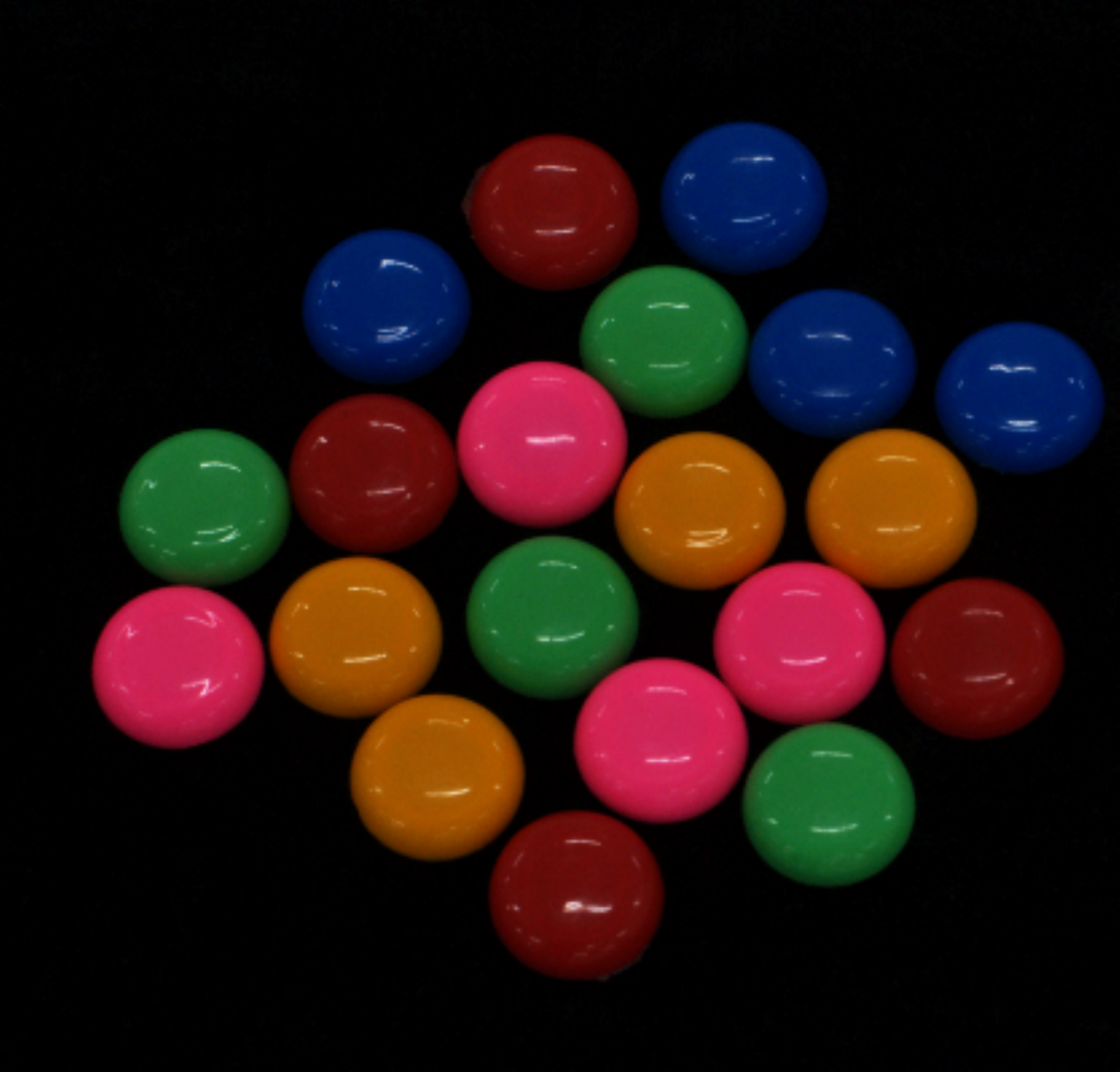}}\hspace{0.0mm}
\subfigure[]{
  \includegraphics[width=0.48\linewidth]{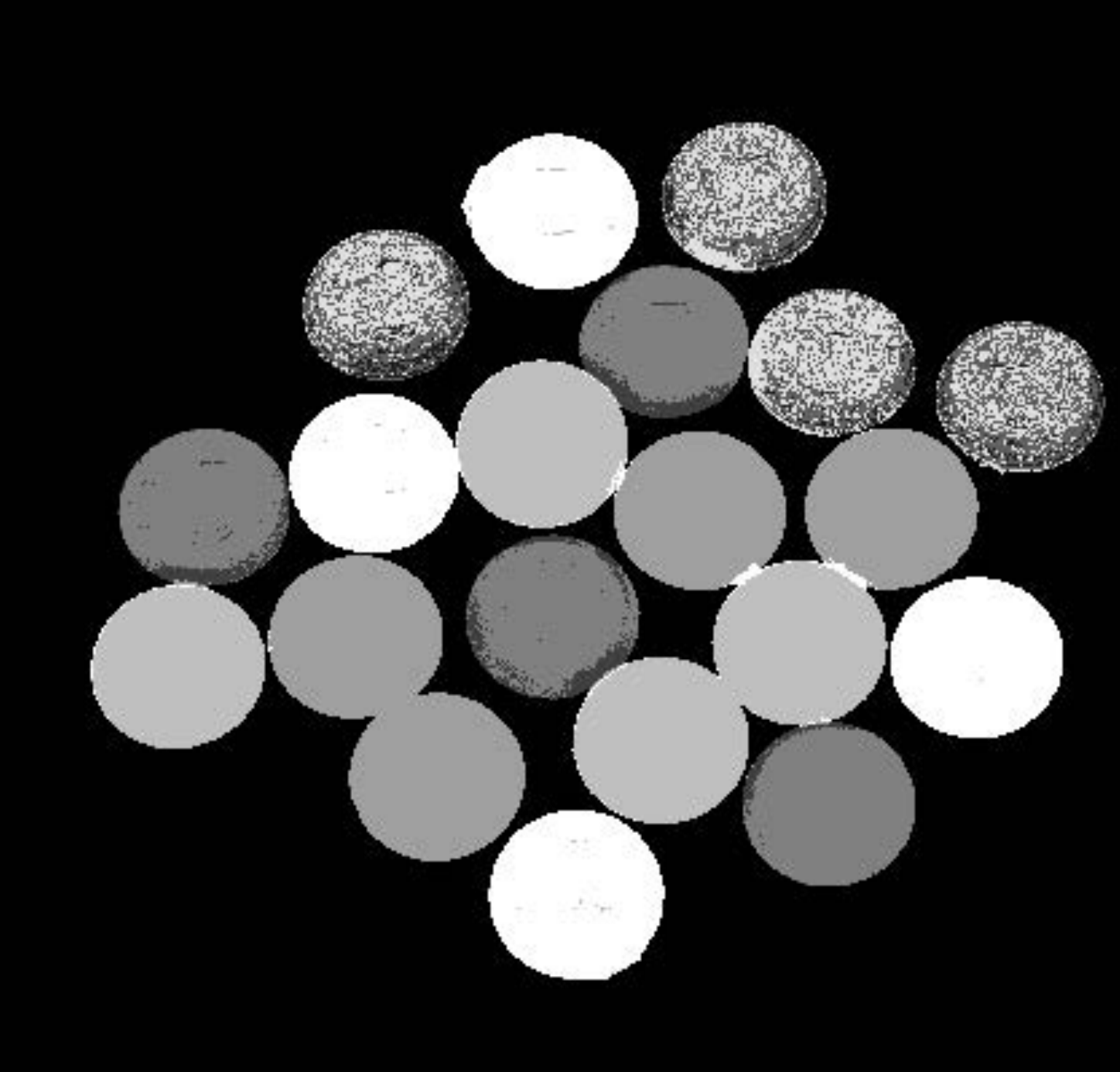}}\vspace{2mm}
\subfigure[]{
  \includegraphics[width=0.48\linewidth]{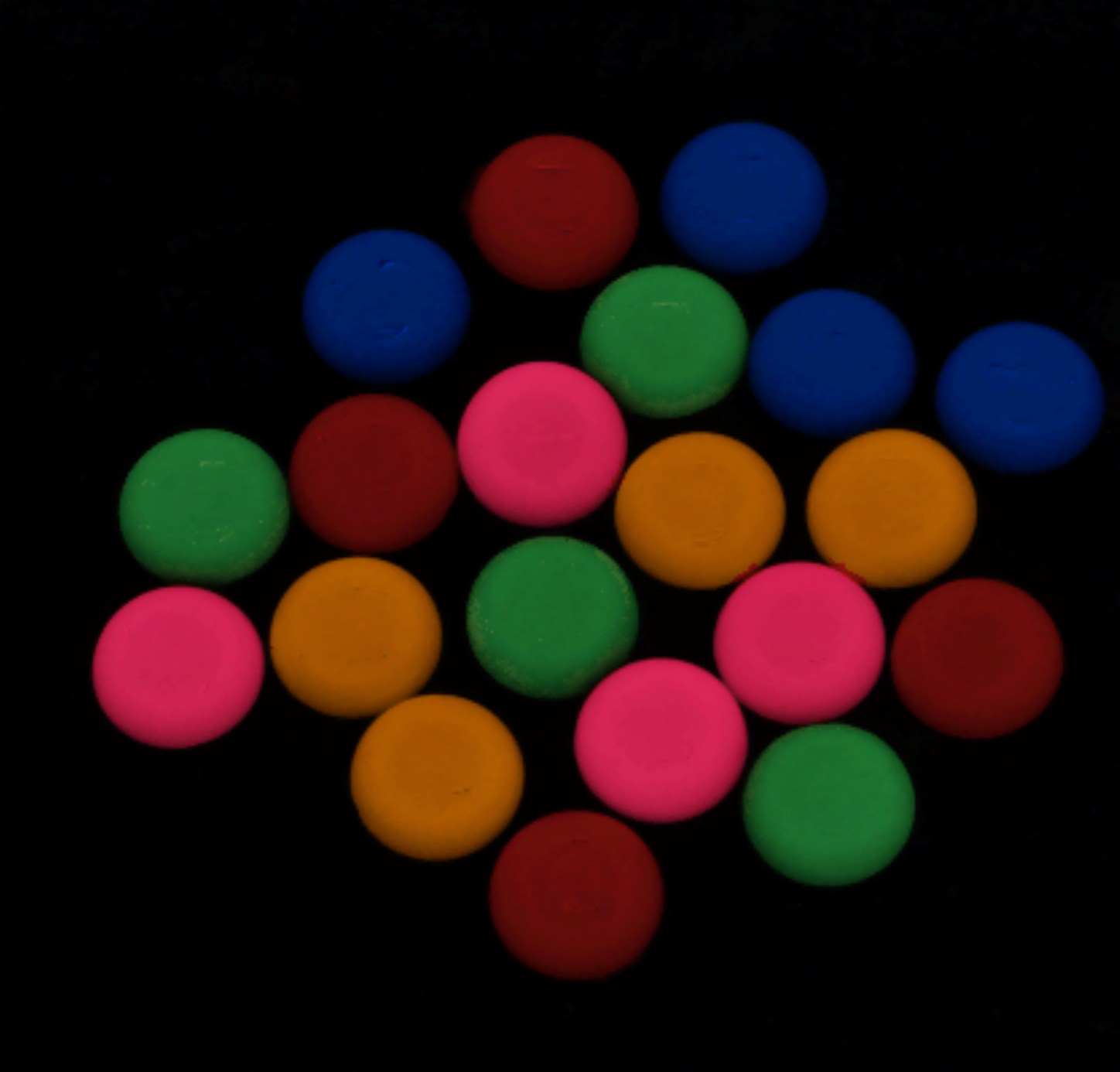}}\hspace{0.0mm}
\subfigure[]{
  \includegraphics[width=0.48\linewidth]{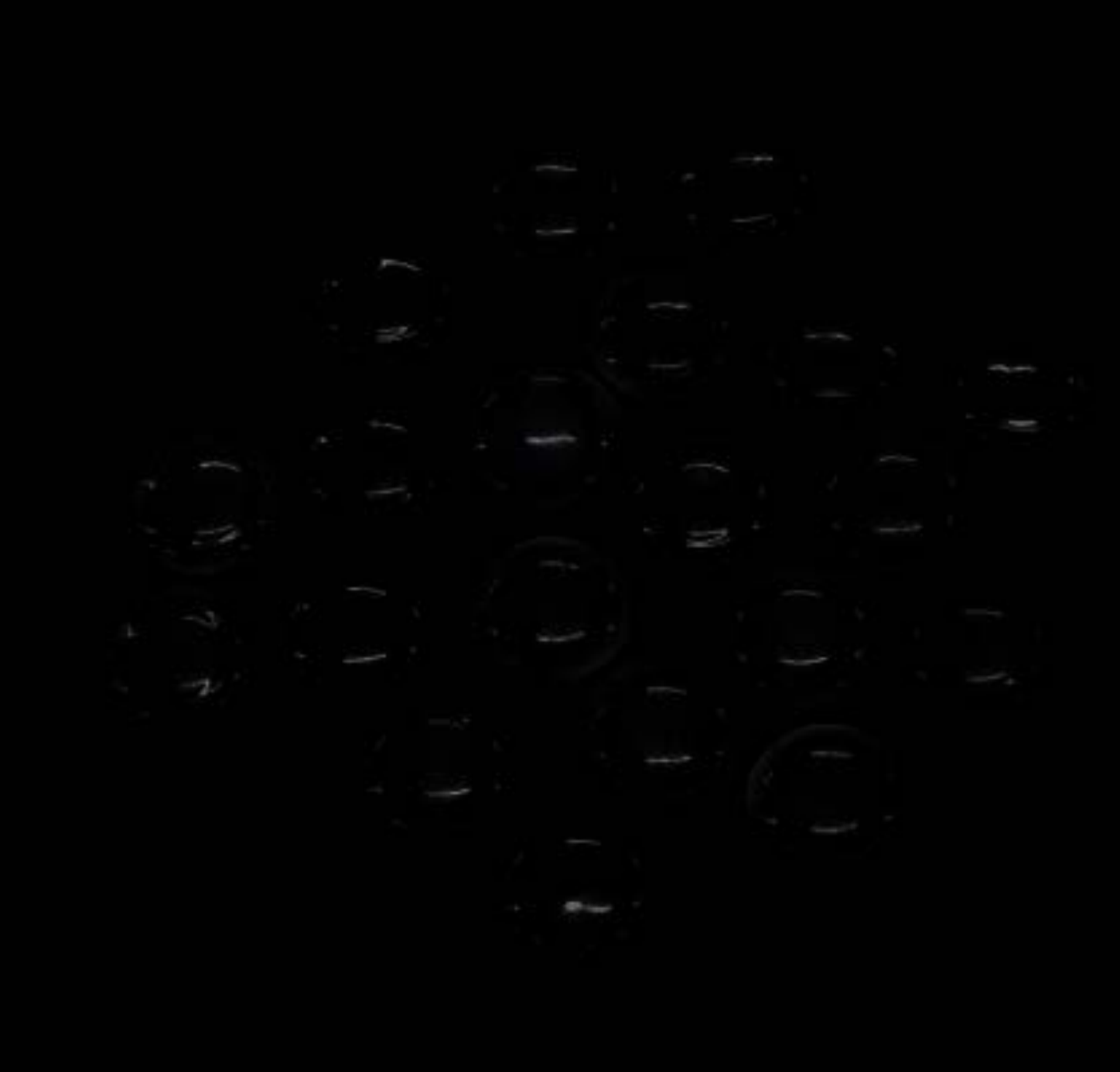}}\hspace{0.0mm}
  \caption{{The results with inaccurate clustering (more clusters than the true material types). (a) Input image. (b) Over-segmented clustering result. (c) Recovered diffuse component. (c) Recovered specular component.}}
  \label{fig:cluster_result}
\end{figure}

Another factor might degenerate the final performance is the accuracy of material clustering. Because of the noise,  a precise clustering cannot be guaranteed even with a theoretically strict reflection model and clustering criterion. It is worth noting that, the proposed approach is clustering based but does not require accurate clustering, which is prone to noise. Actually, we tend to use a strict clustering criterion and over segment the materials when noise exists. Even with a cluster number more than the real material types, our approach still works well if only there exist pure diffuse pixels in each cluster. For example, the scene in Fig.~\ref{fig:cluster_result}(a) includes 5 kinds of materials but is clustered into 8 groups (the blue magnets are over segmented), as visualized in Fig.~\ref{fig:cluster_result}(b), where we label different clusters with different grey levels.
From the separated diffuse and specular components displayed in Fig.~\ref{fig:cluster_result}(c) and Fig.~\ref{fig:cluster_result}(d), one can see that we can get promising separation even with an incorrect clustering. 

\subsection{Results on Non-Lambertian Nature Scenes}
\begin{figure*}[p]
\centering
\subfigure[]{
\begin{minipage}[ht]{0.188\textwidth}
      \vspace{0.4\linewidth}
      \includegraphics[width=1.0\linewidth]{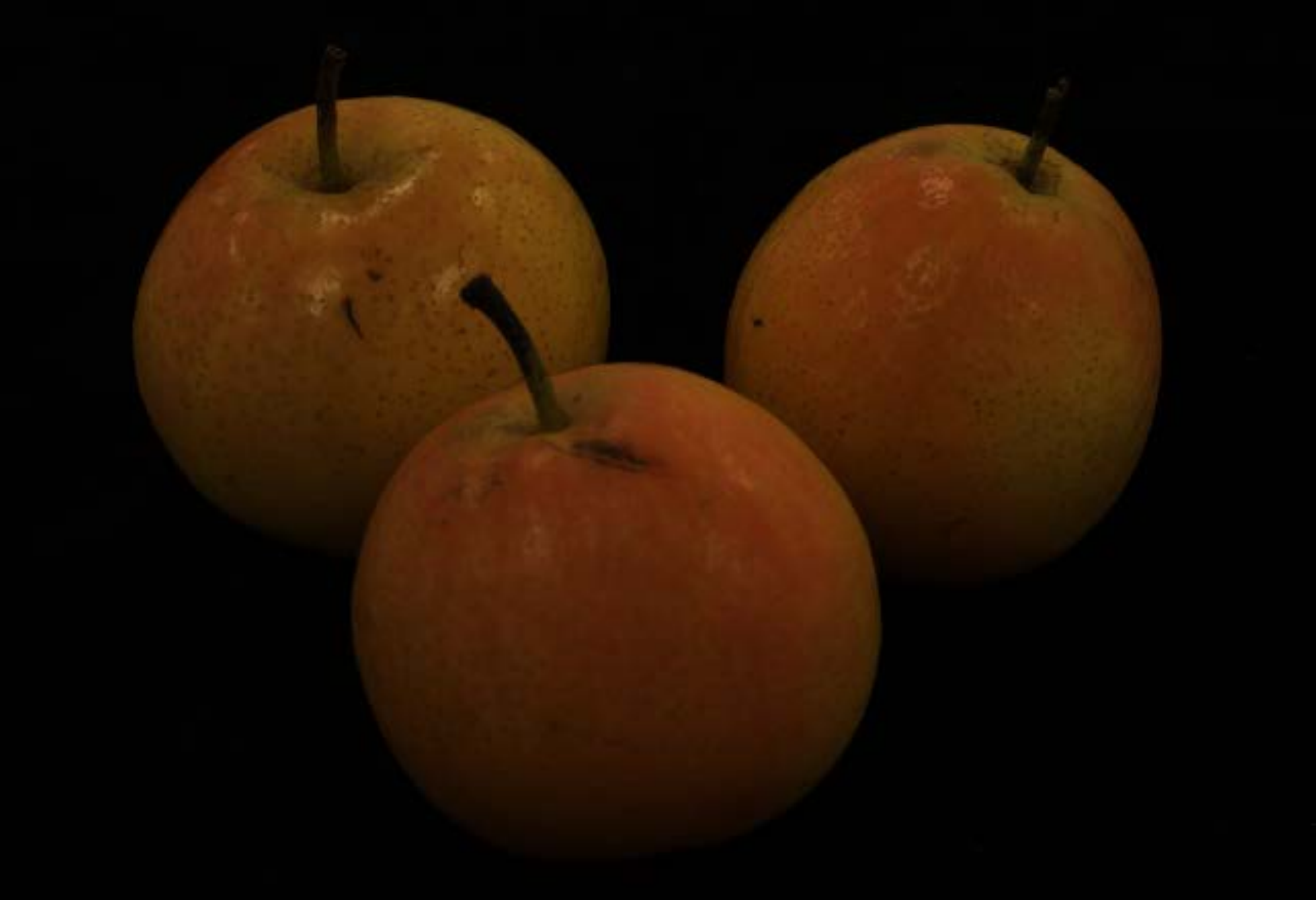}\\\vspace{0.55\linewidth}
      \includegraphics[width=1.0\linewidth]{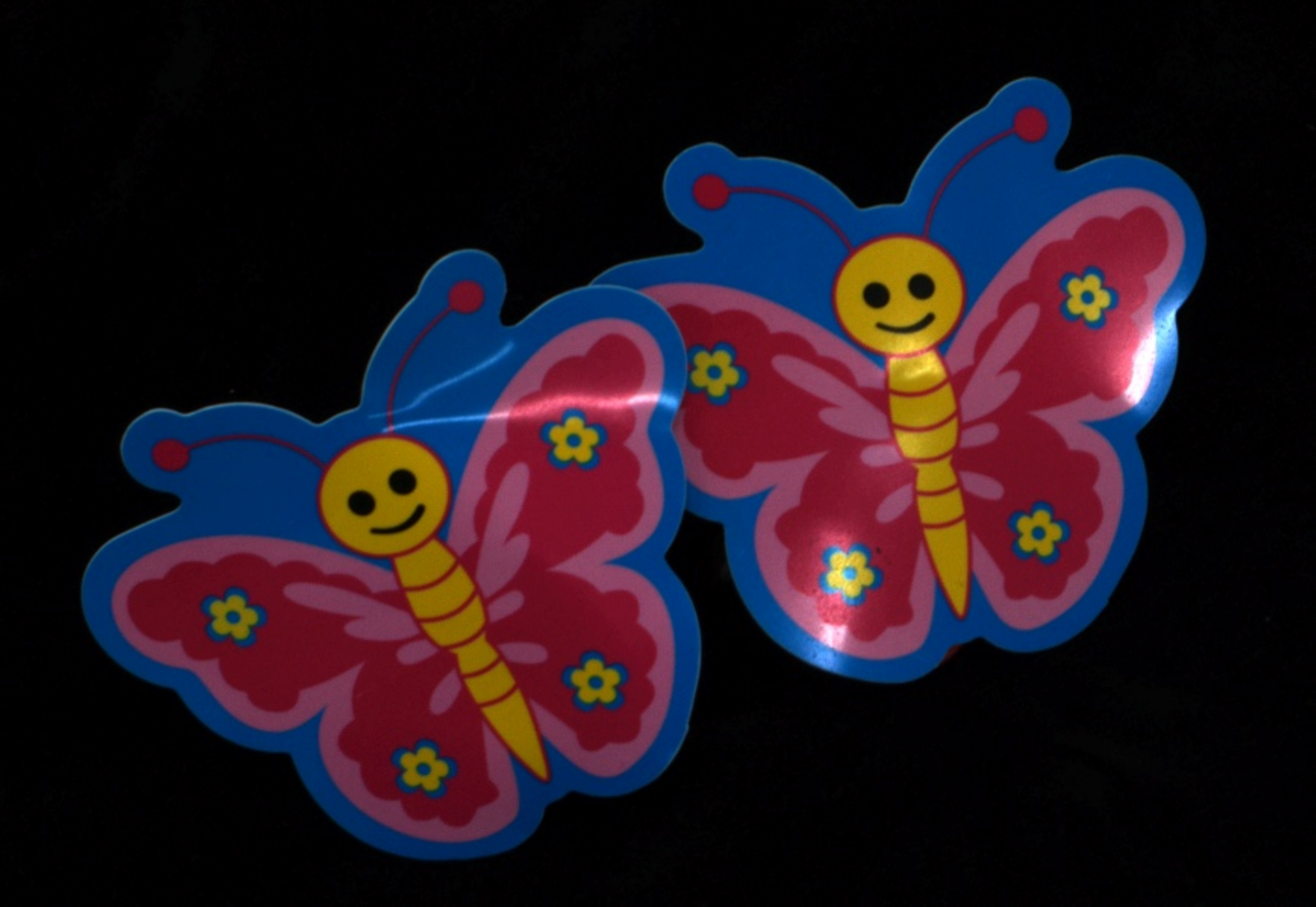}\\\vspace{0.7\linewidth}
      \includegraphics[width=1.0\linewidth]{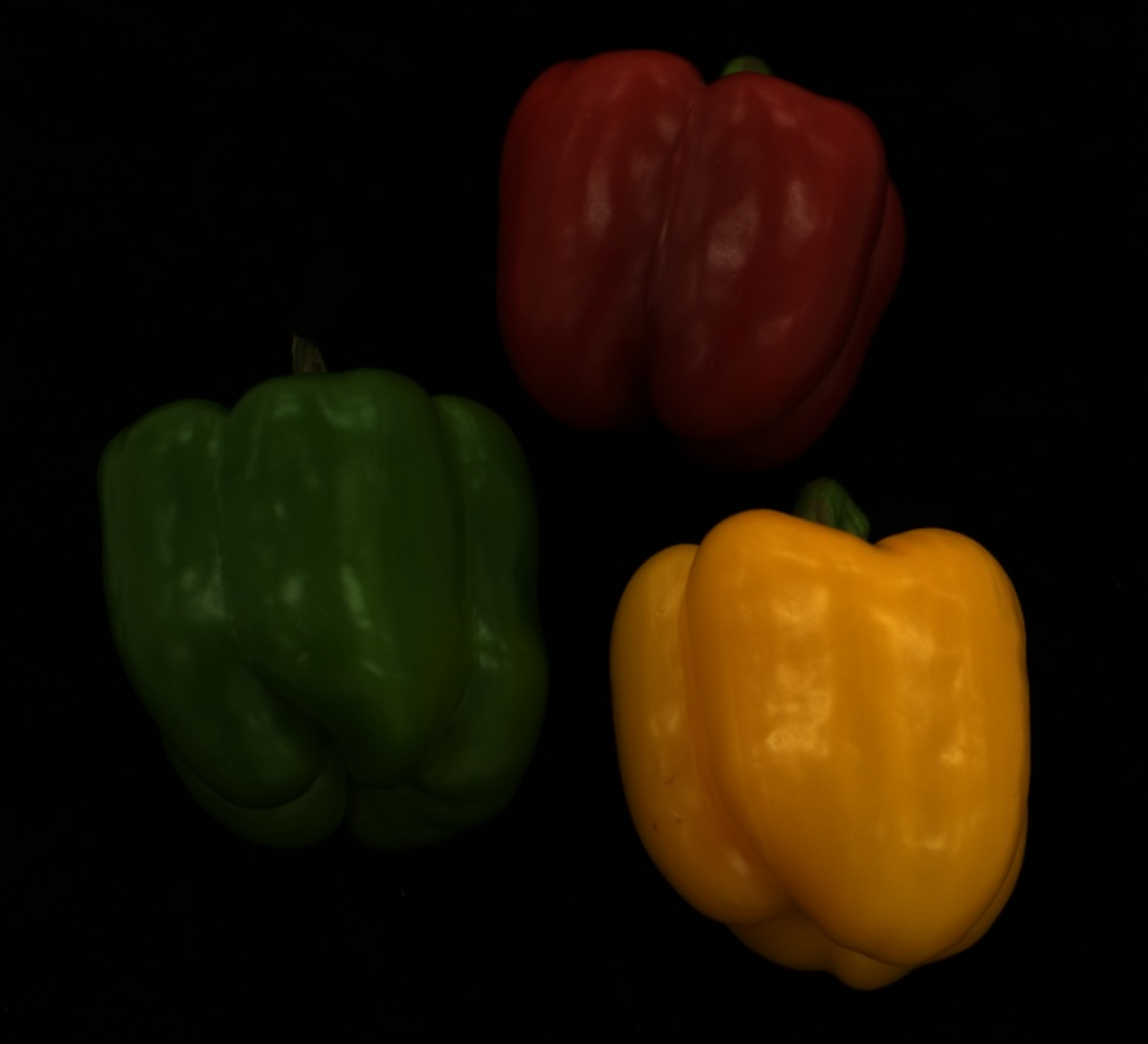}\\\vspace{0.75\linewidth}
      \includegraphics[width=1.0\linewidth]{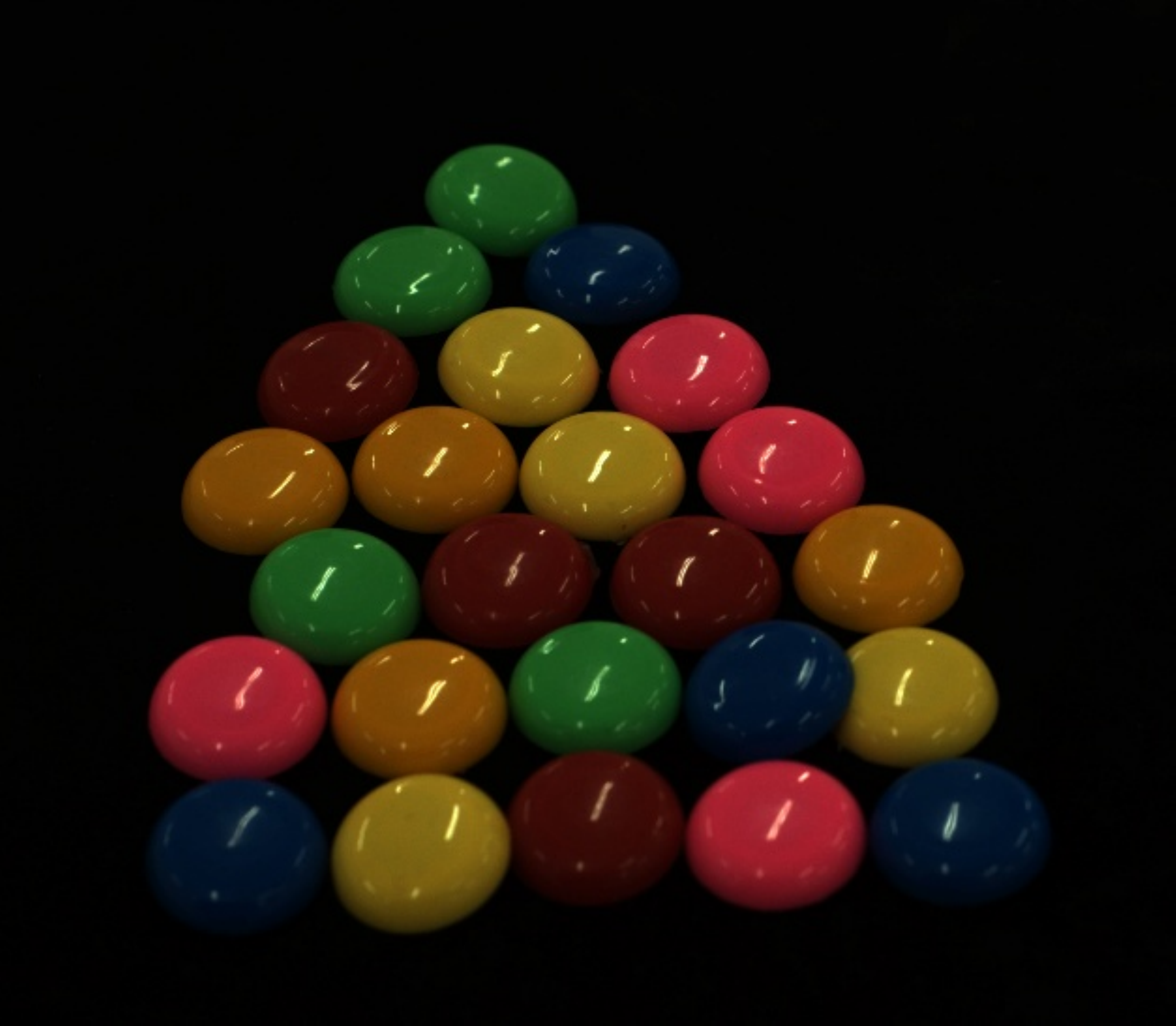}\\\vspace{0.35\linewidth}
 \end{minipage}}\hspace{-0.1mm}
\subfigure[]{
\begin{minipage}[ht]{0.188\textwidth}
      \includegraphics[width=1.0\linewidth]{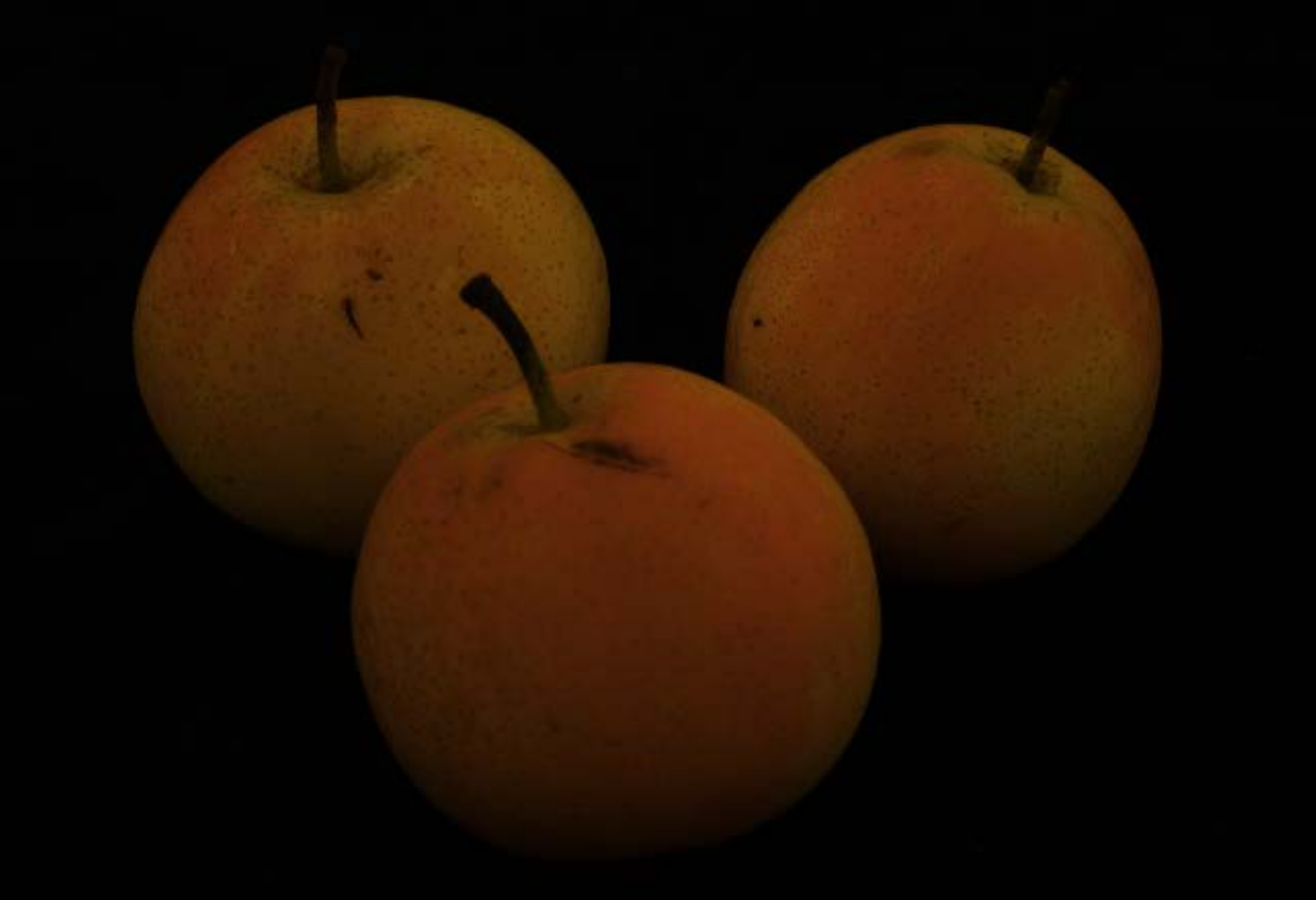}
      \includegraphics[width=1.0\linewidth]{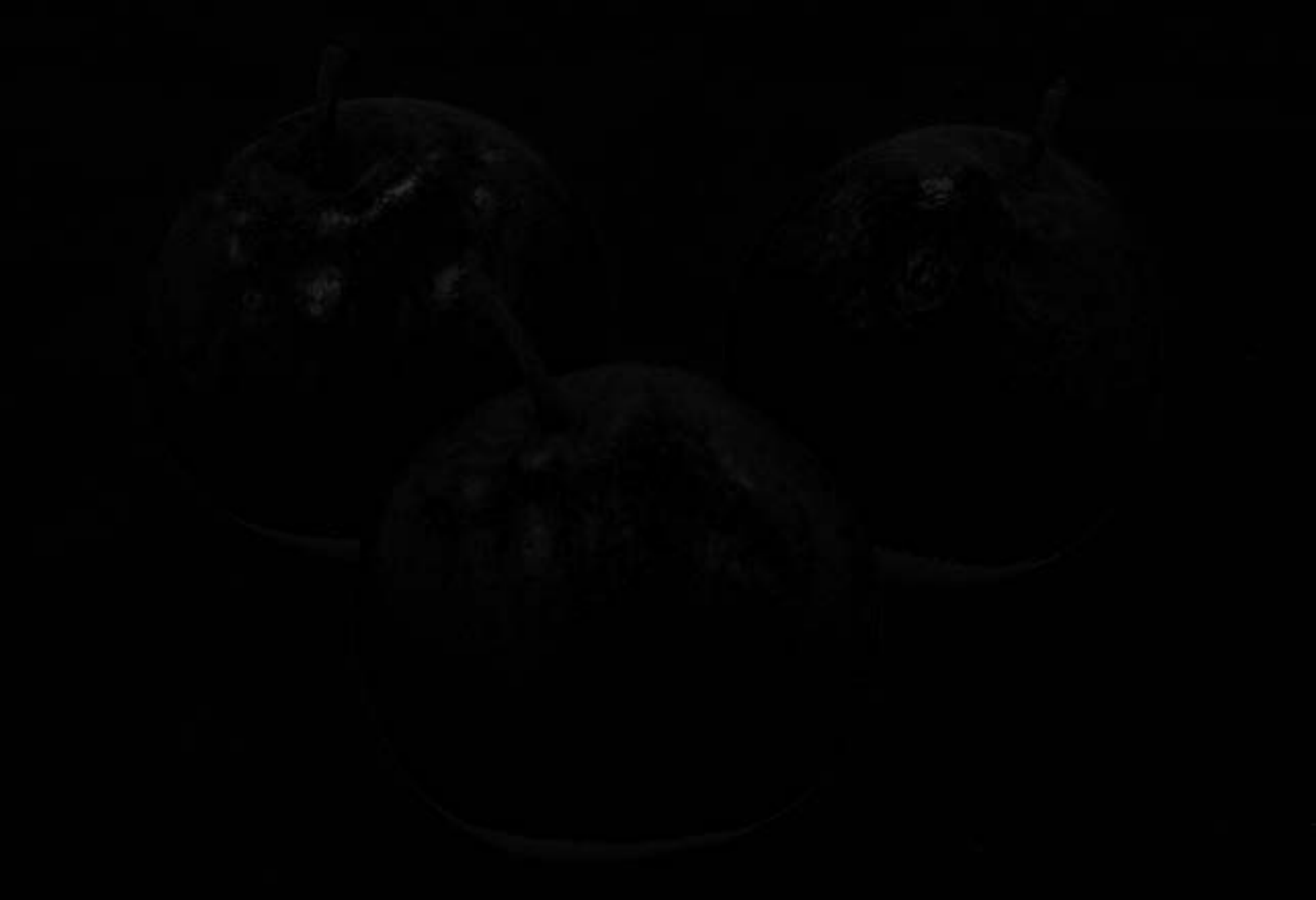}\vspace{1mm}
      \includegraphics[width=1.0\linewidth]{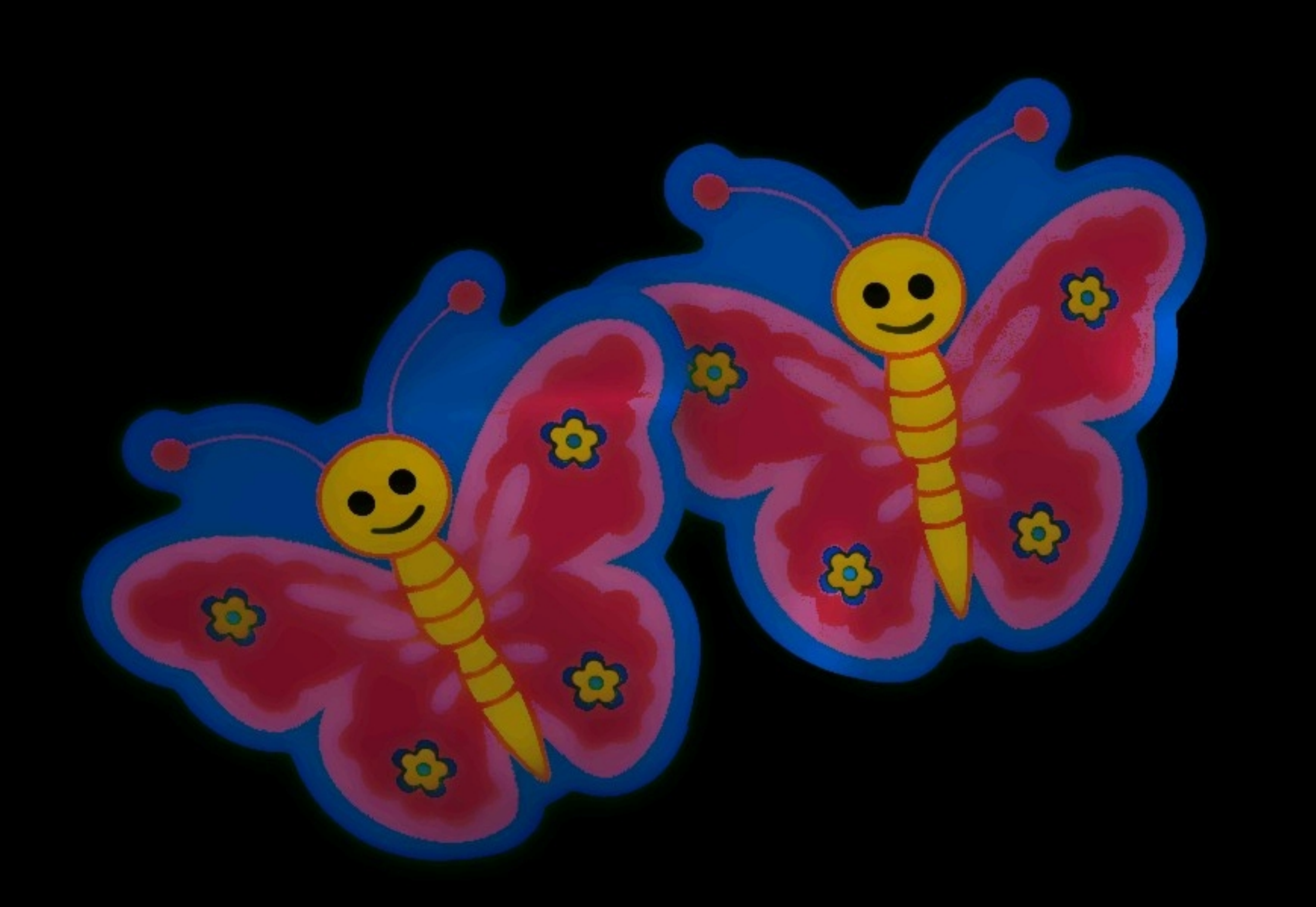}
      \includegraphics[width=1.0\linewidth]{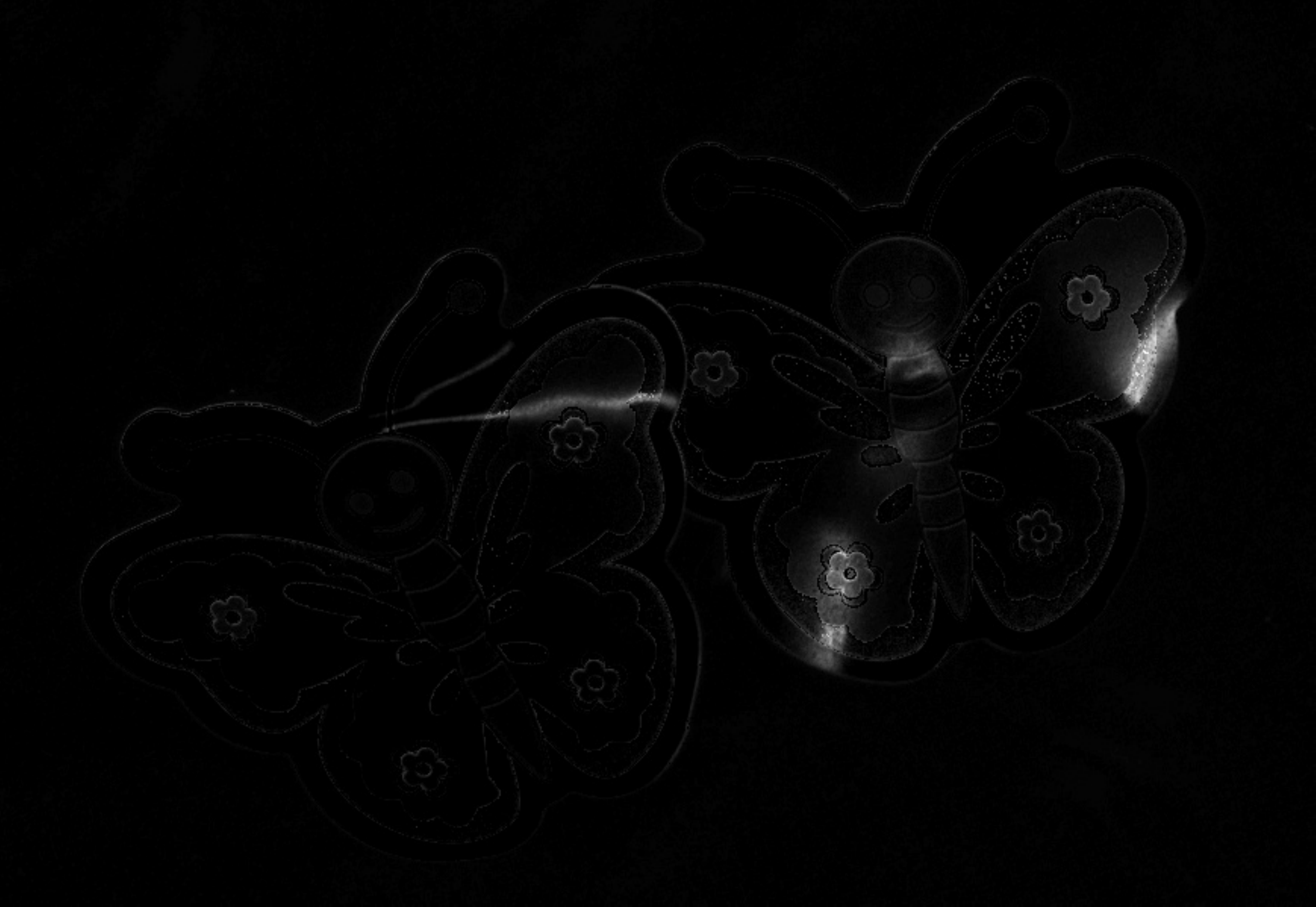}\vspace{1mm}
      \includegraphics[width=1.0\linewidth]{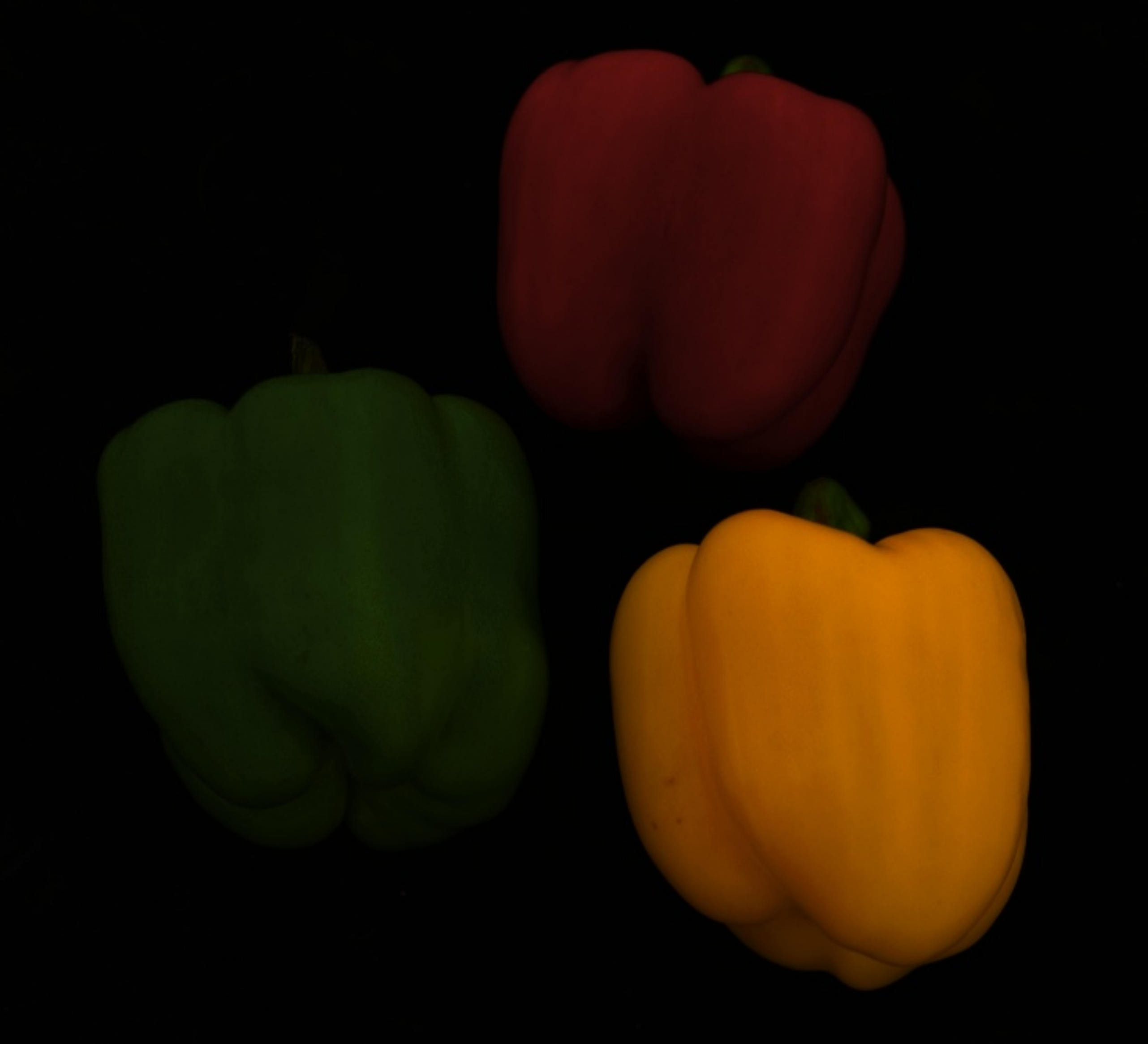}
      \includegraphics[width=1.0\linewidth]{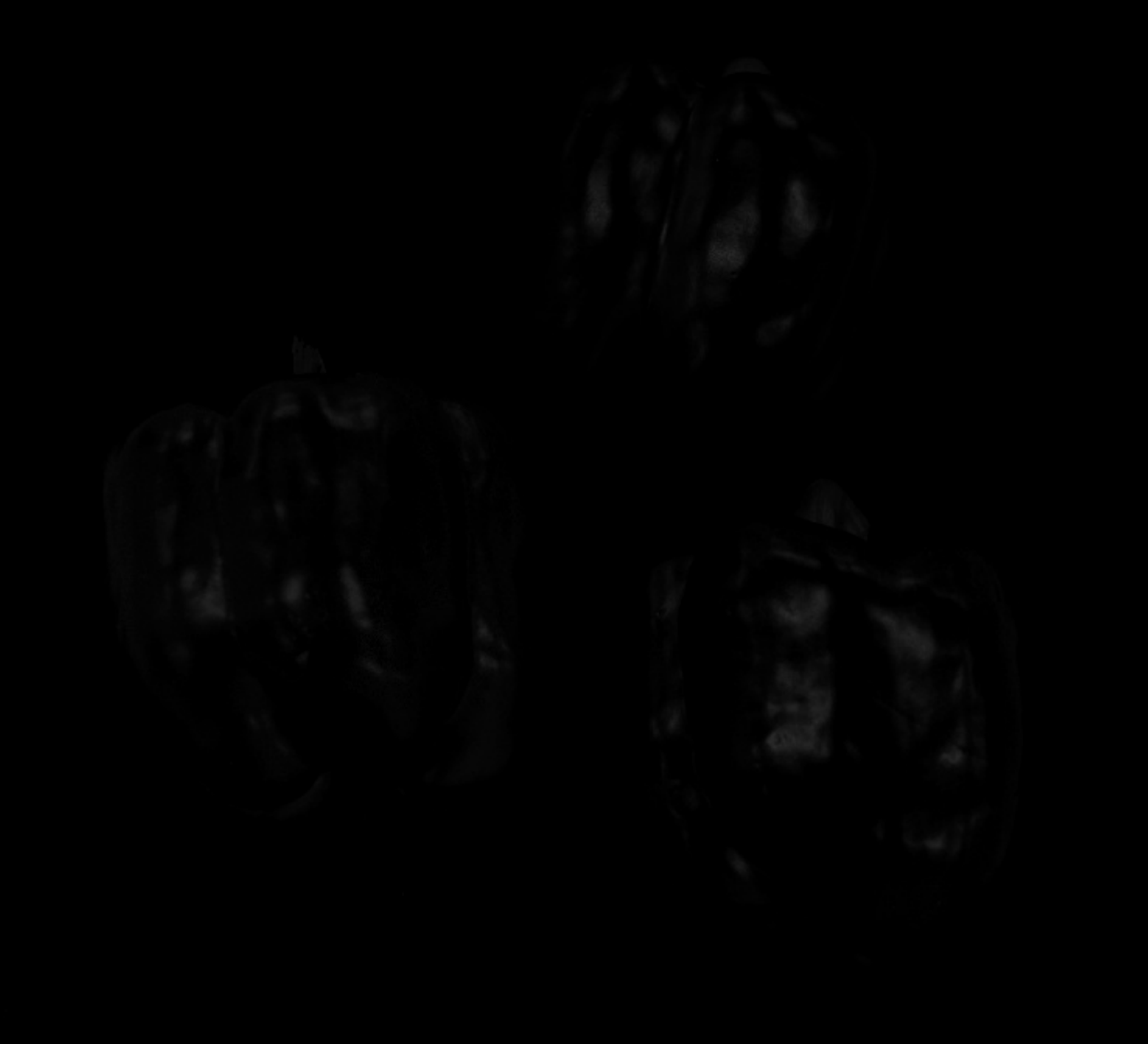}\vspace{1mm}
      \includegraphics[width=1.0\linewidth]{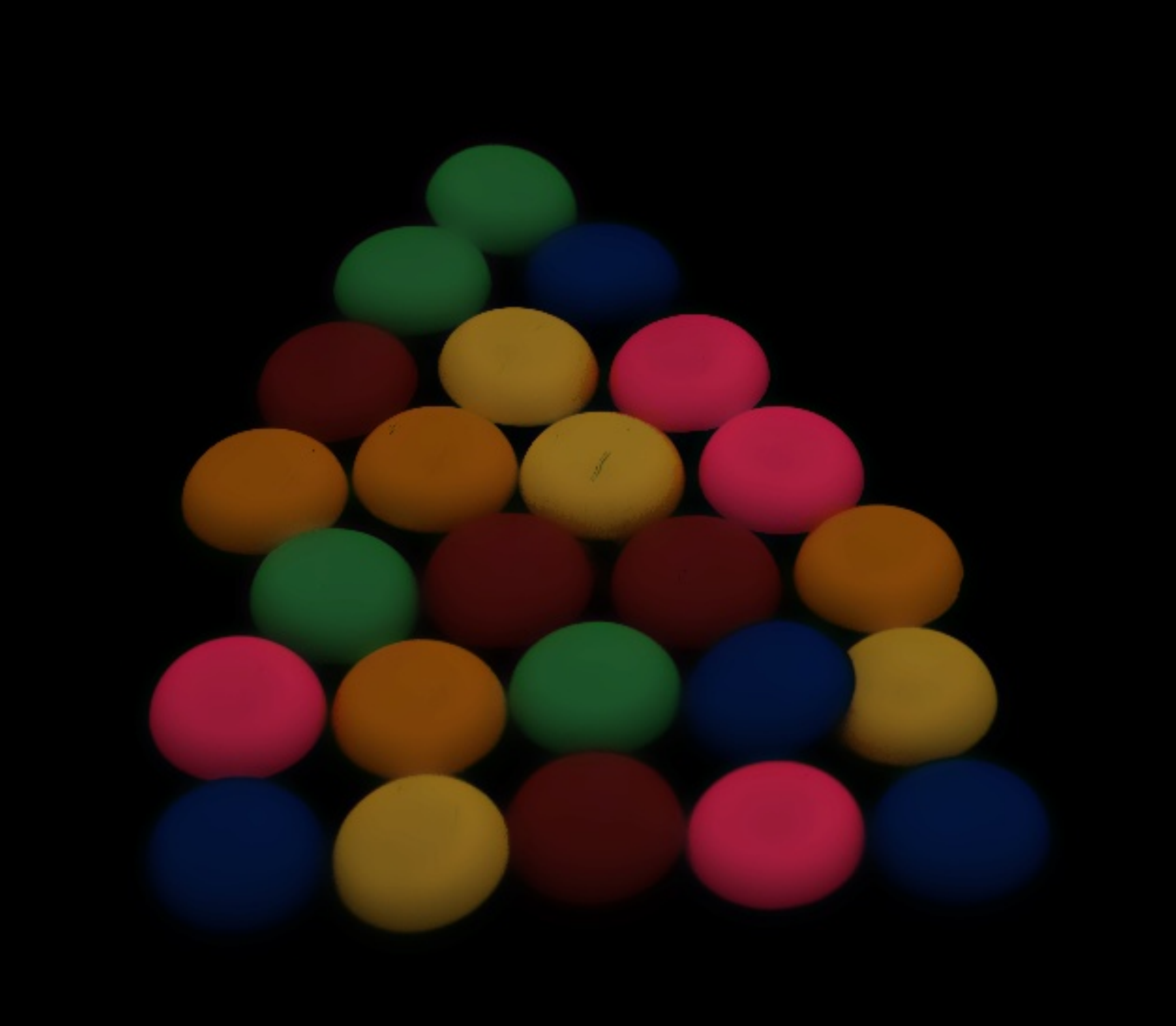}\\
      \includegraphics[width=1.0\linewidth]{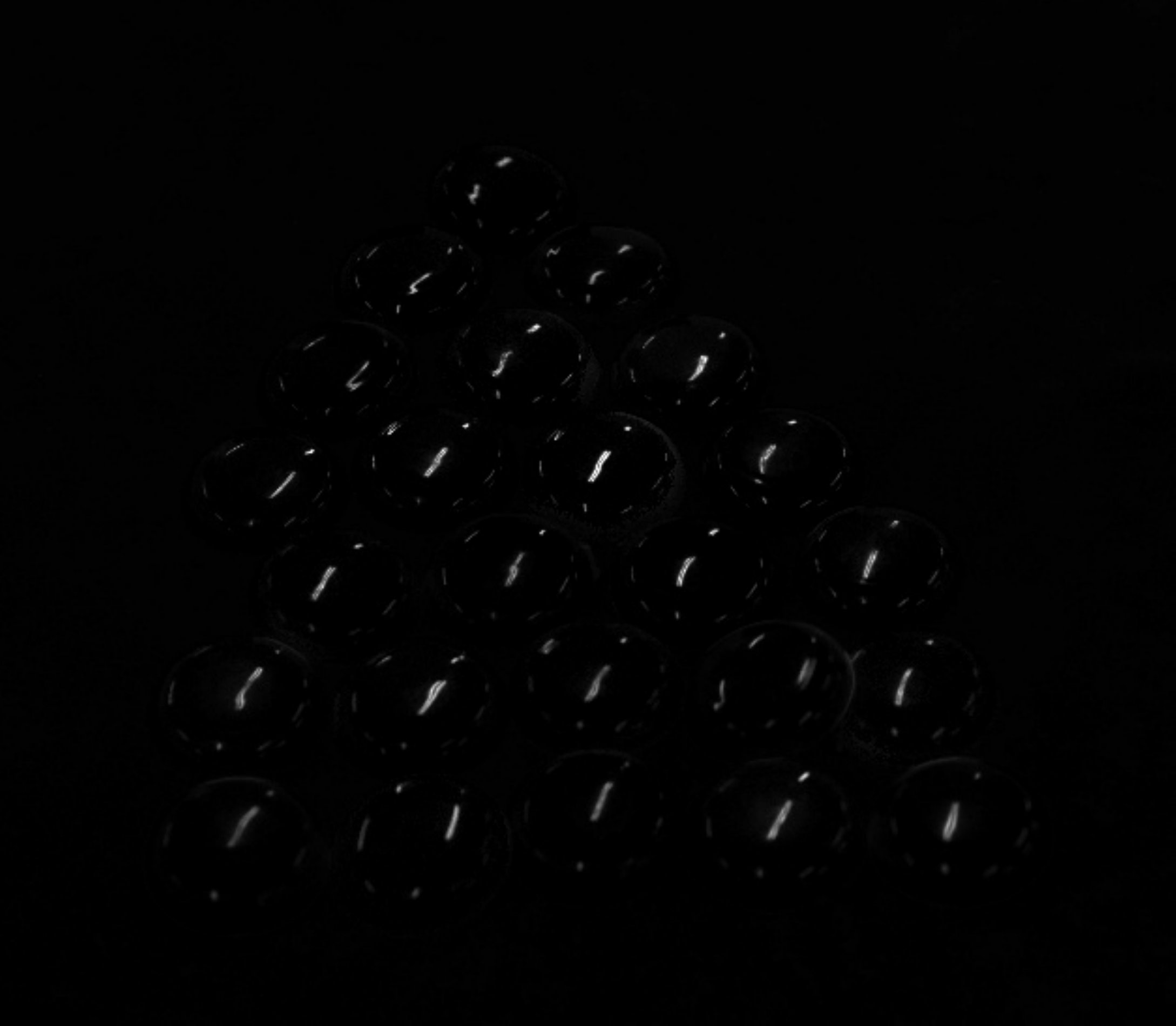}\vspace{2mm}
 \end{minipage}}\hspace{-0.1mm}
 \subfigure[]{
 \begin{minipage}[ht]{0.188\textwidth}
      \includegraphics[width=1.0\linewidth]{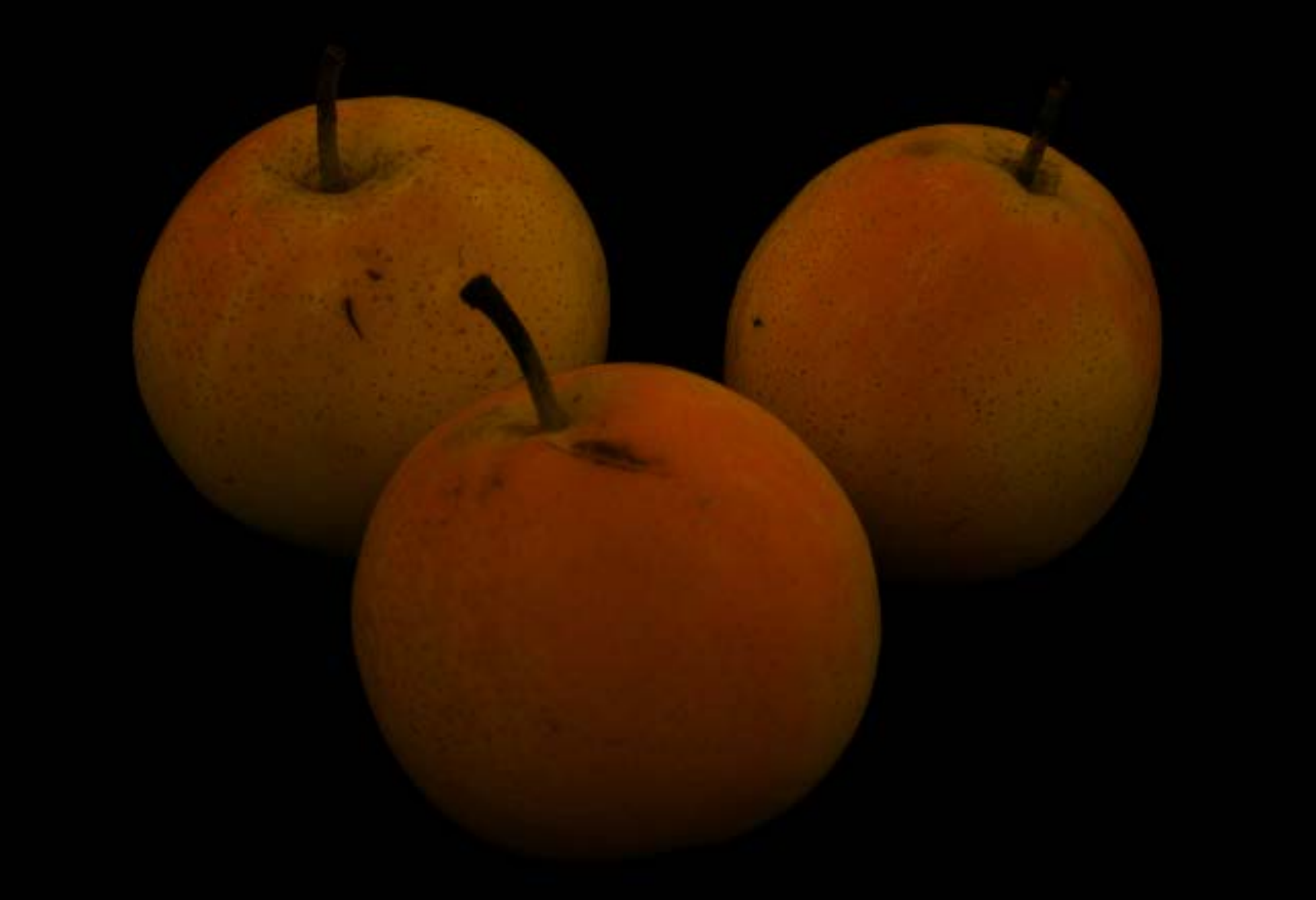}
      \includegraphics[width=1.0\linewidth]{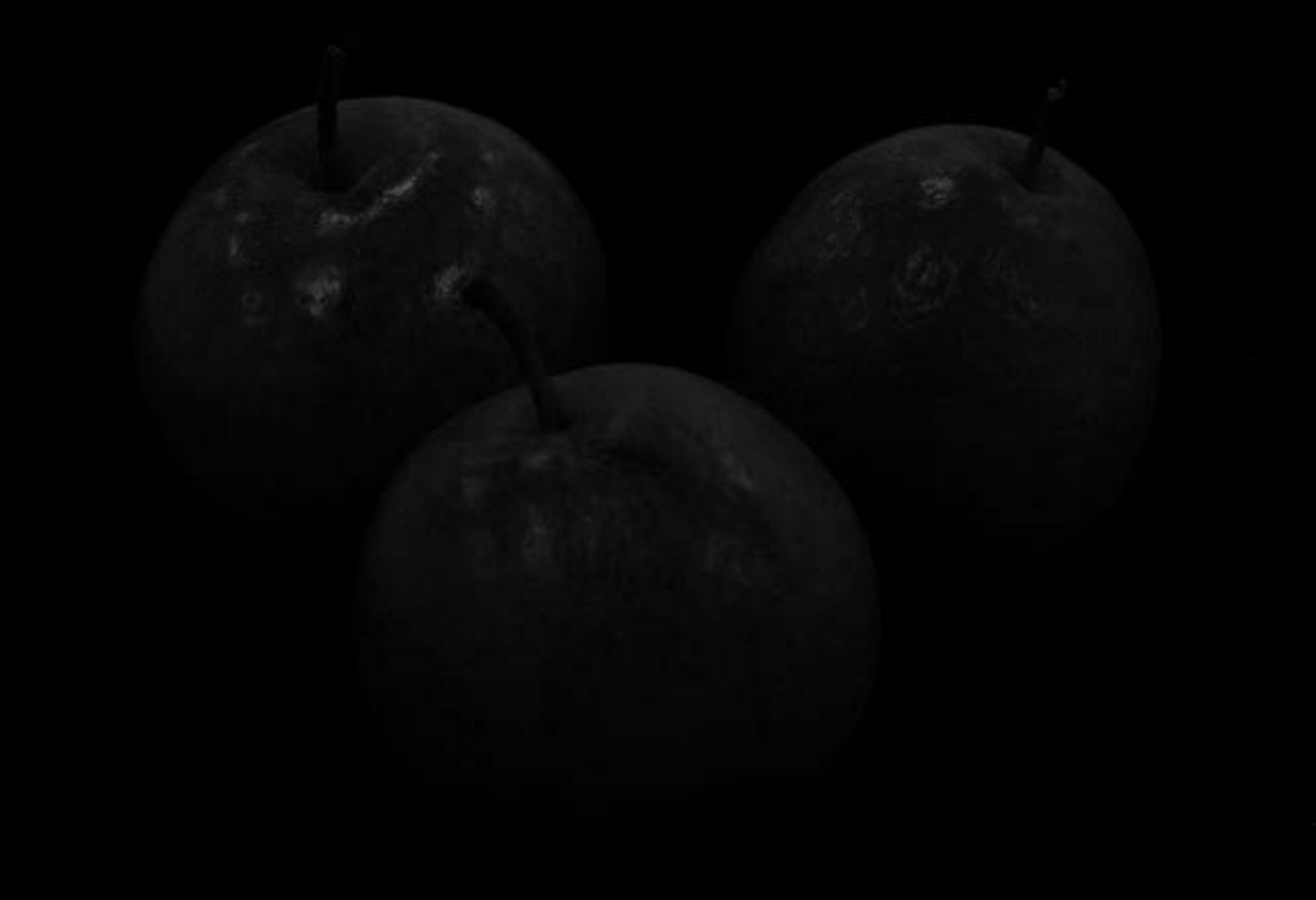}\vspace{1mm}
      \includegraphics[width=1.0\linewidth]{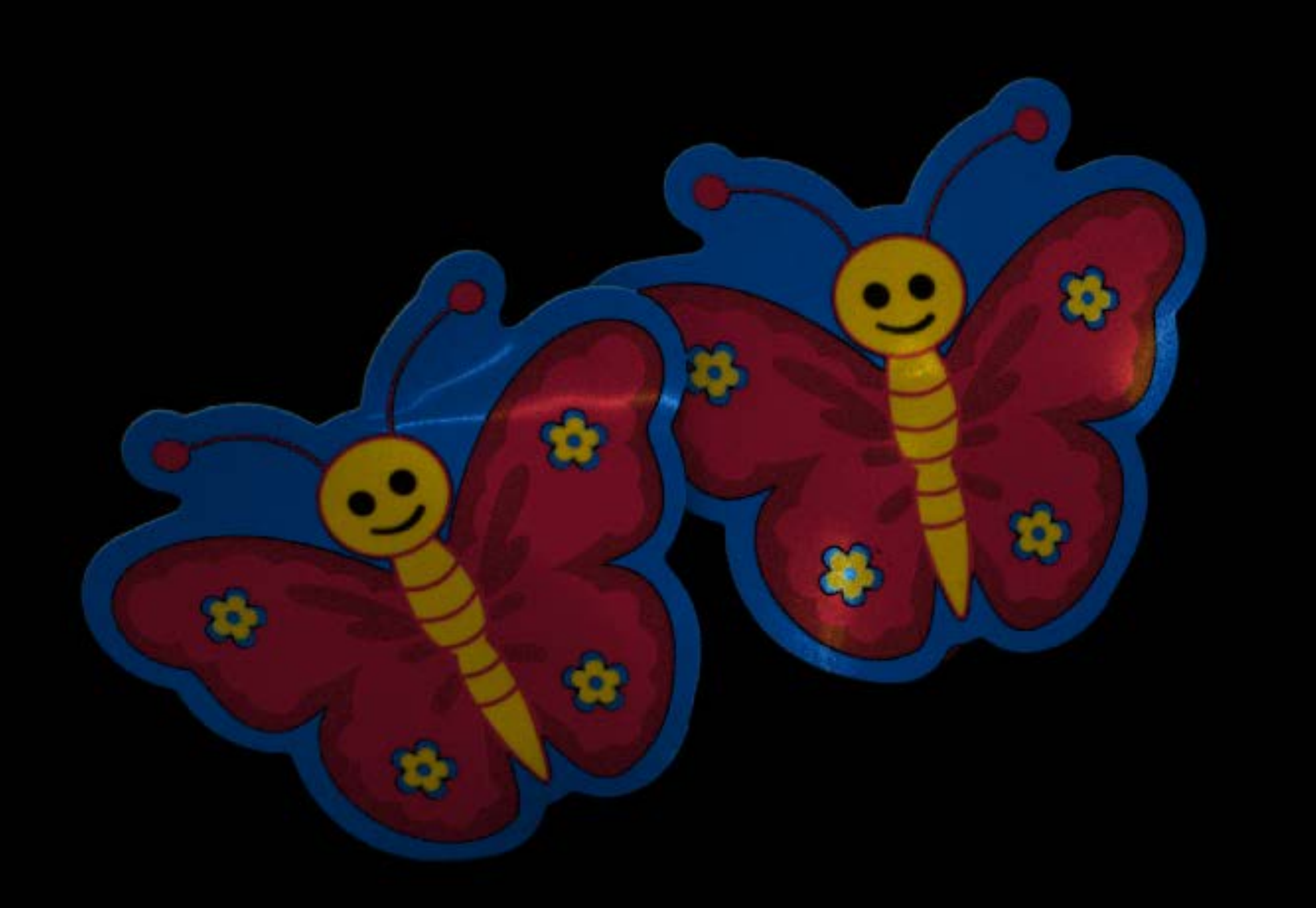}
      \includegraphics[width=1.0\linewidth]{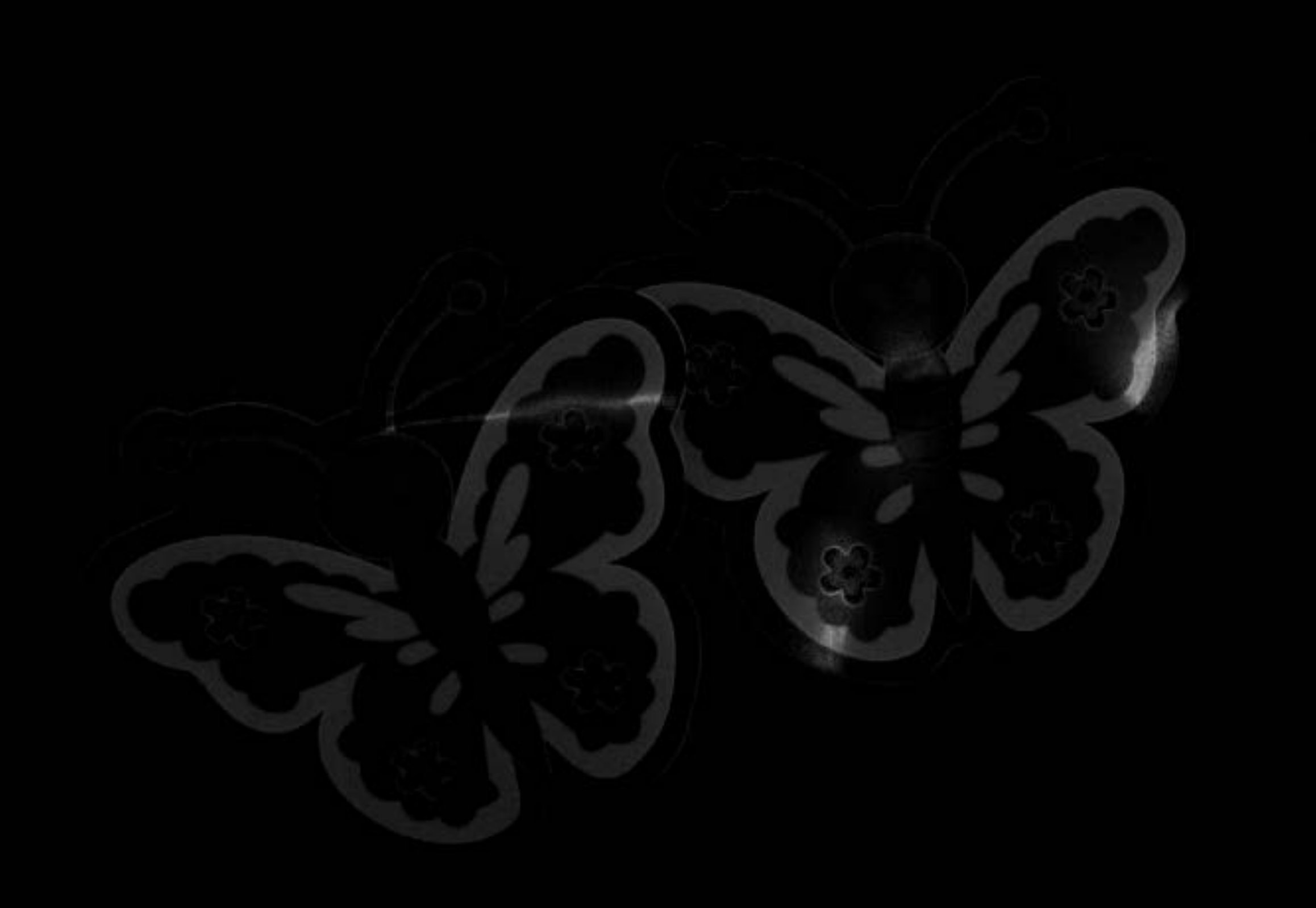}\vspace{1mm}
      \includegraphics[width=1.0\linewidth]{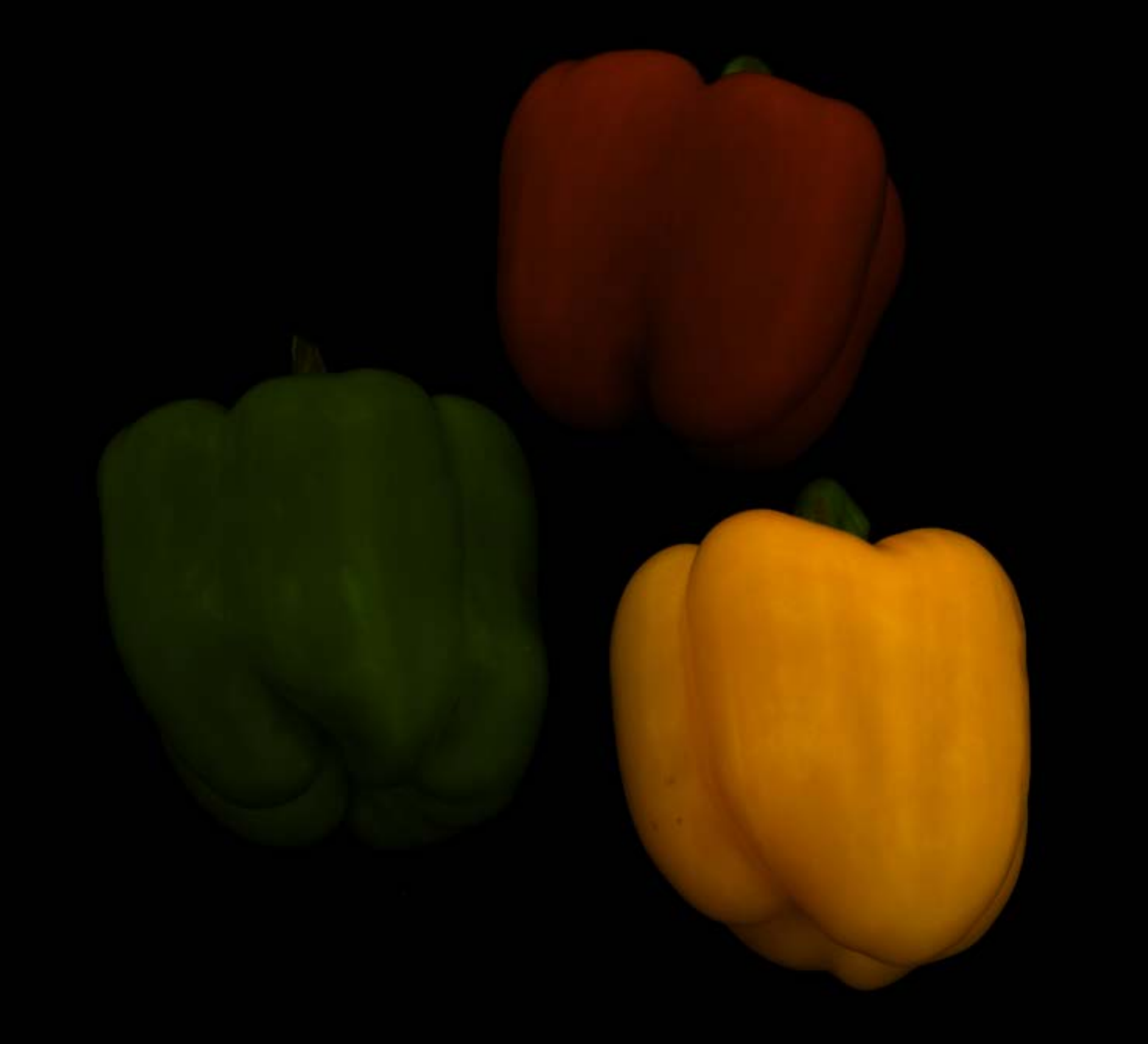}
      \includegraphics[width=1.0\linewidth]{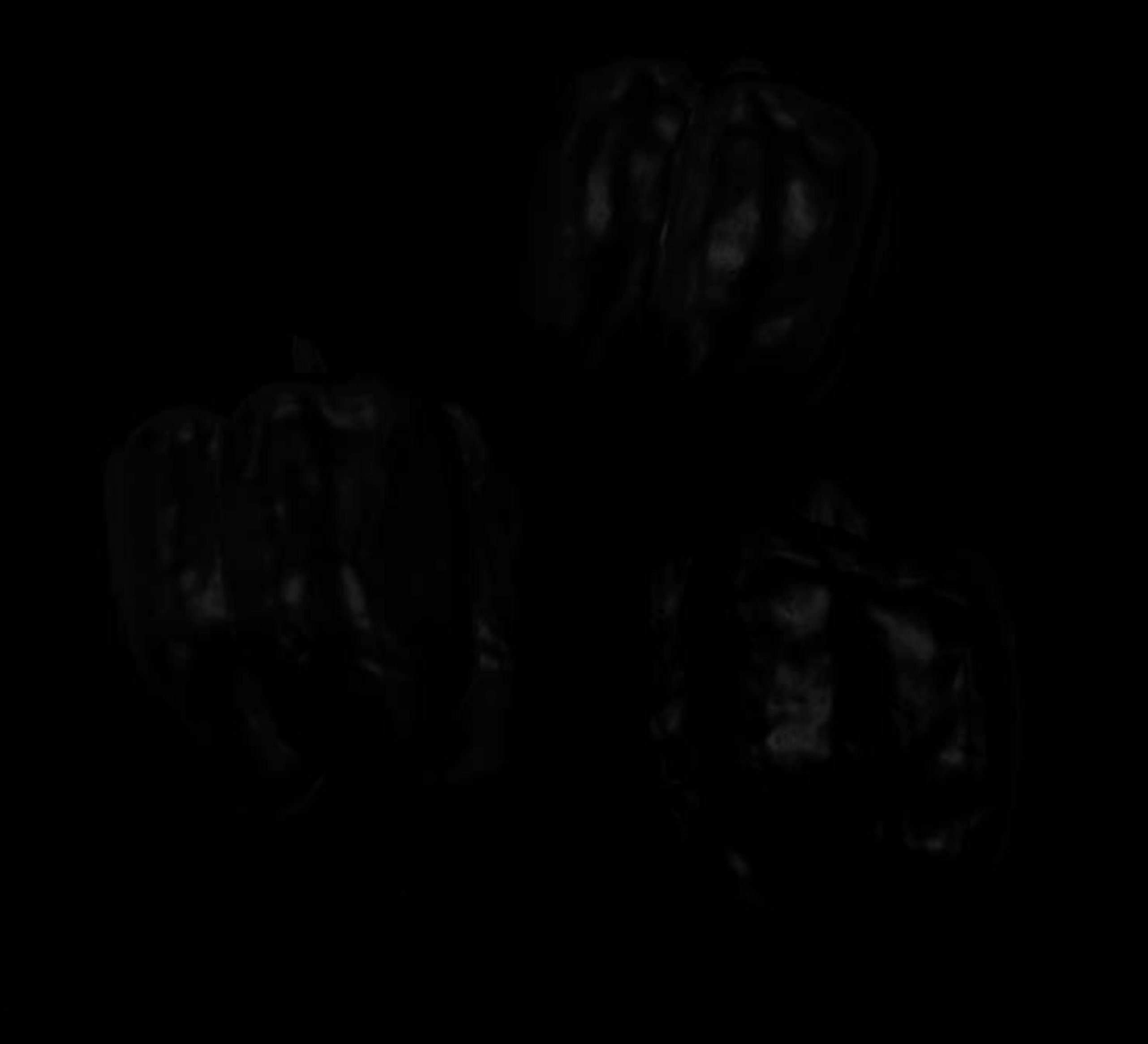}\vspace{1mm}
      \includegraphics[width=1.0\linewidth]{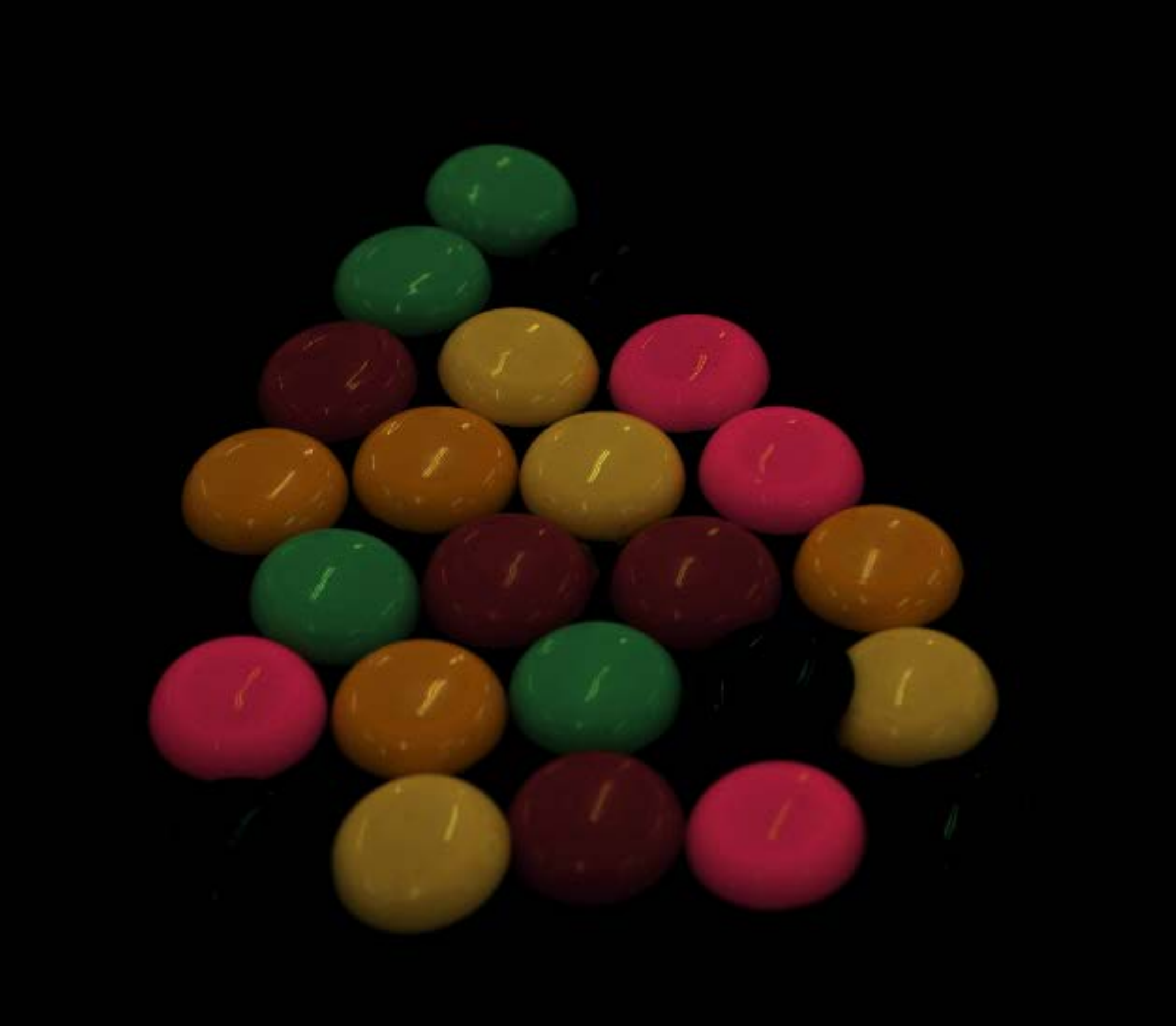}
      \includegraphics[width=1.0\linewidth]{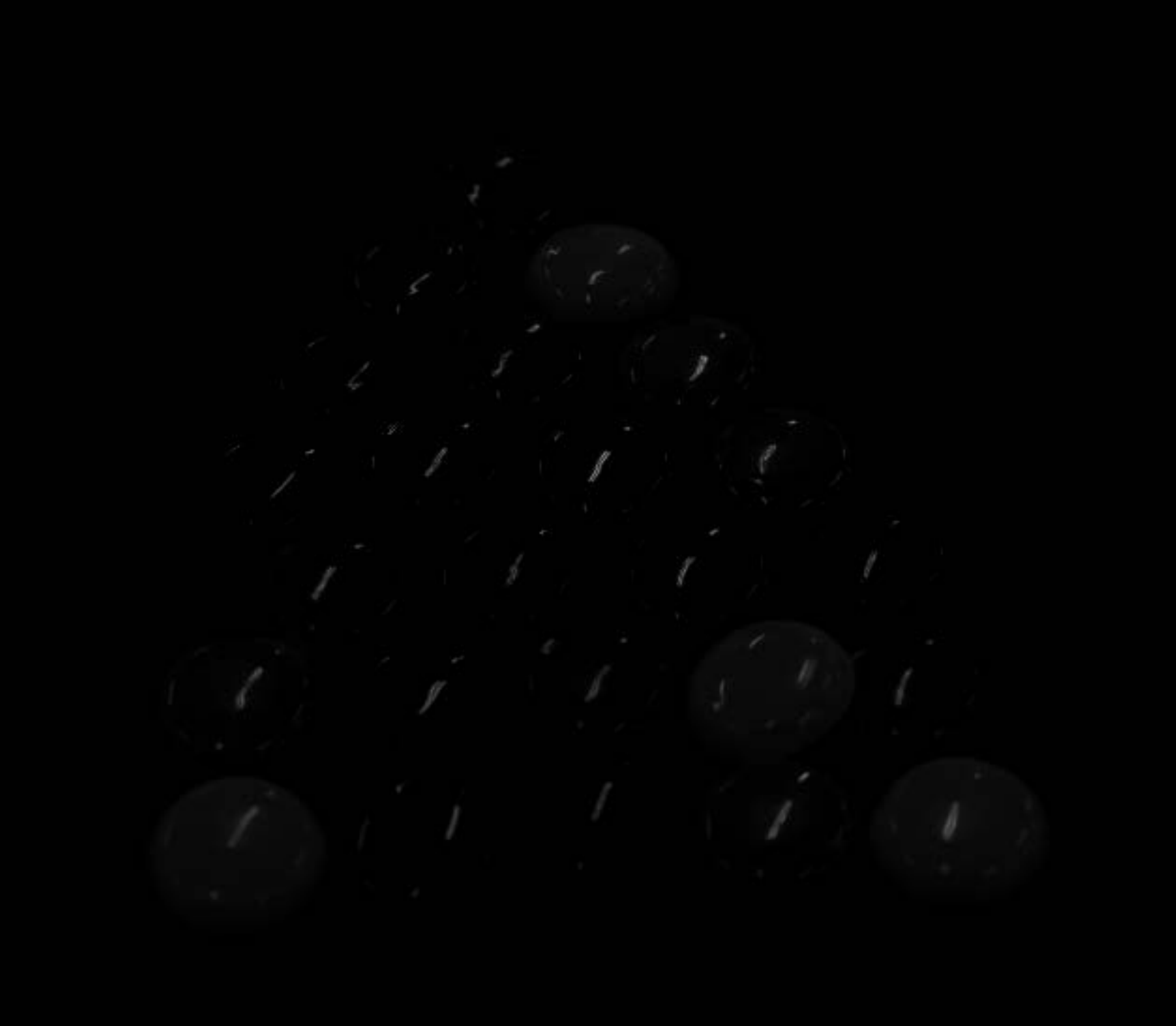}\vspace{2mm}
 \end{minipage}}\hspace{-0.1mm}
 \subfigure[]{
 \begin{minipage}[ht]{0.188\textwidth}
      \includegraphics[width=1.0\linewidth]{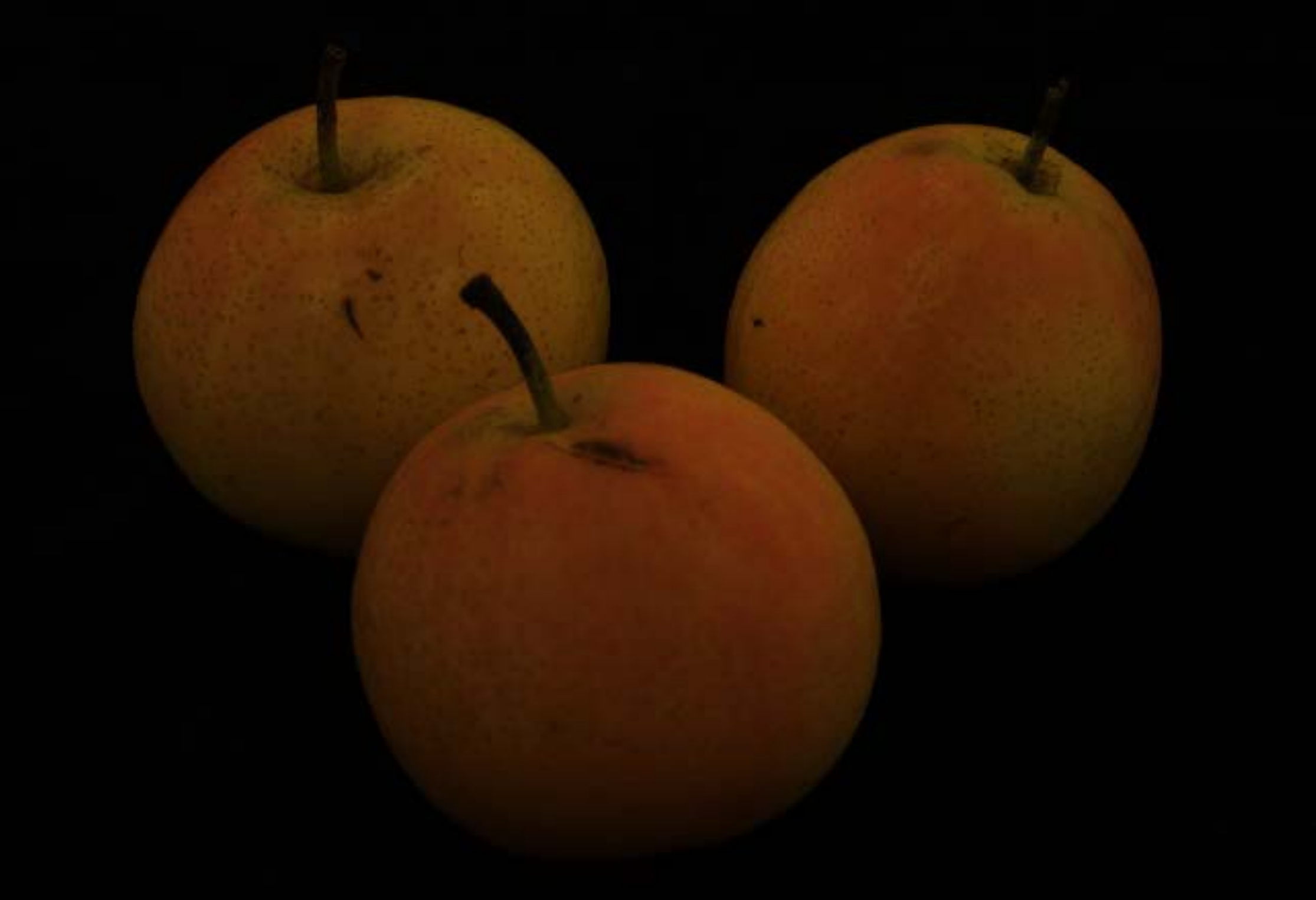}
      \includegraphics[width=1.0\linewidth]{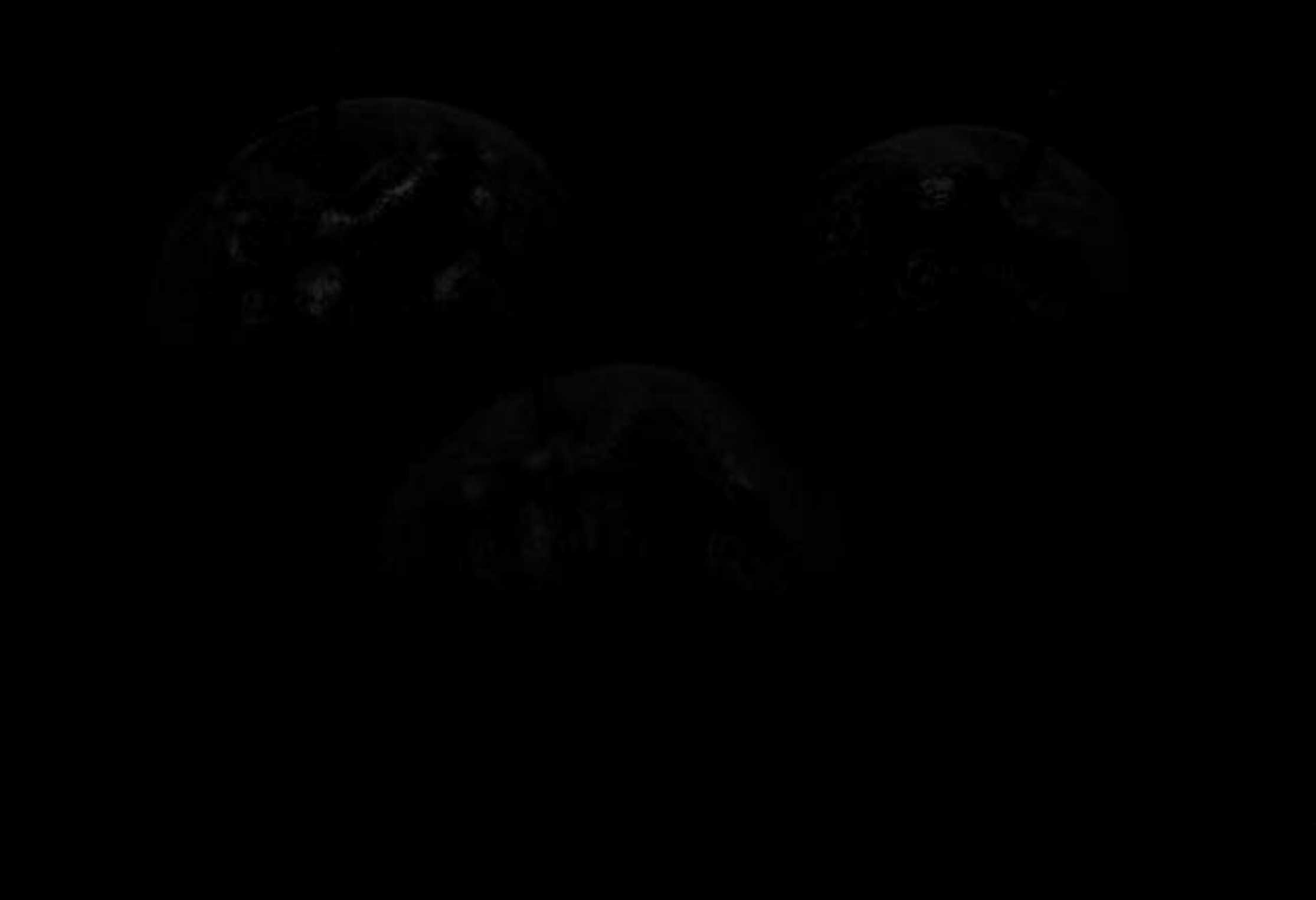}\vspace{1mm}
      \includegraphics[width=1.0\linewidth]{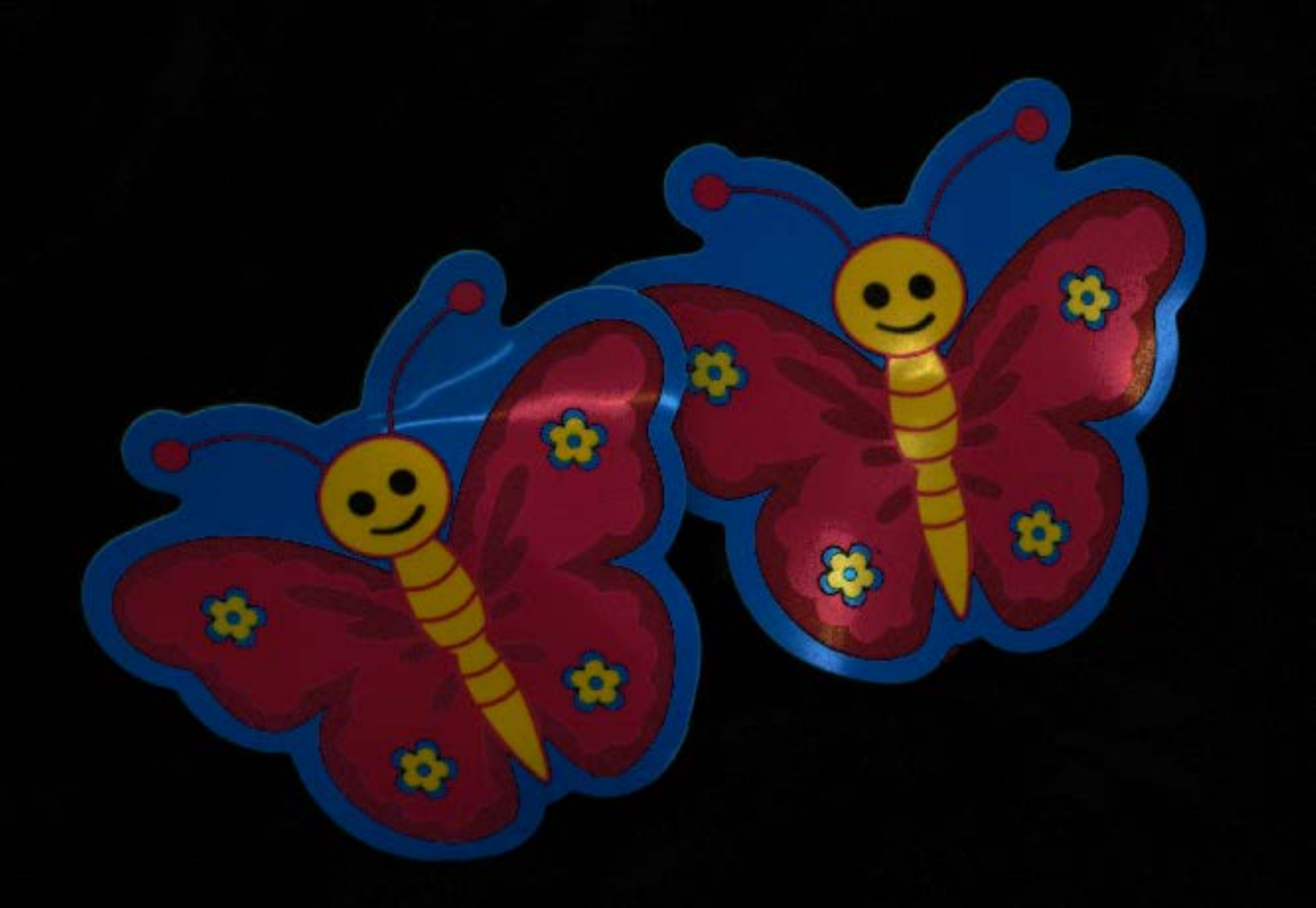}
      \includegraphics[width=1.0\linewidth]{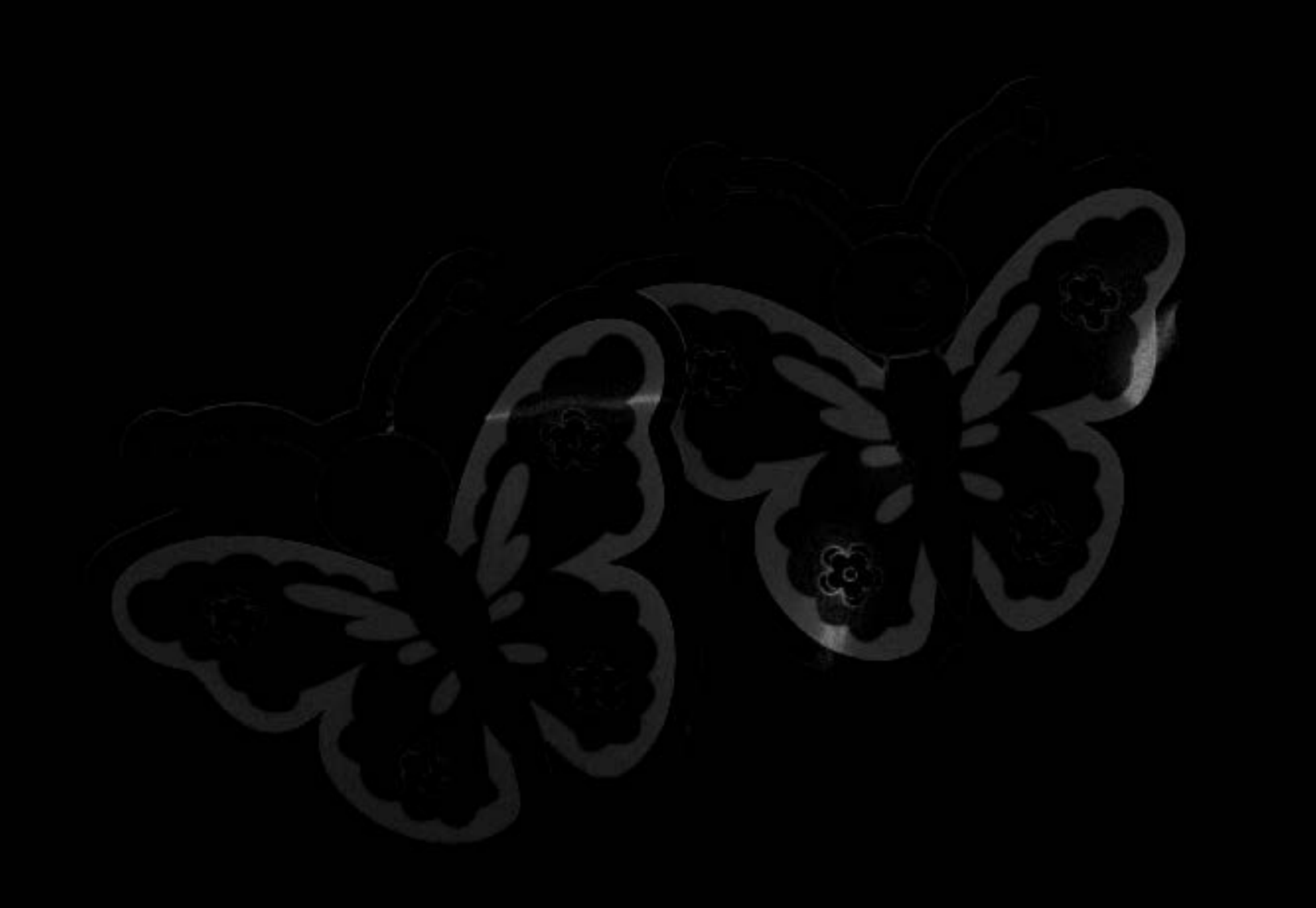}\vspace{1mm}
      \includegraphics[width=1.0\linewidth]{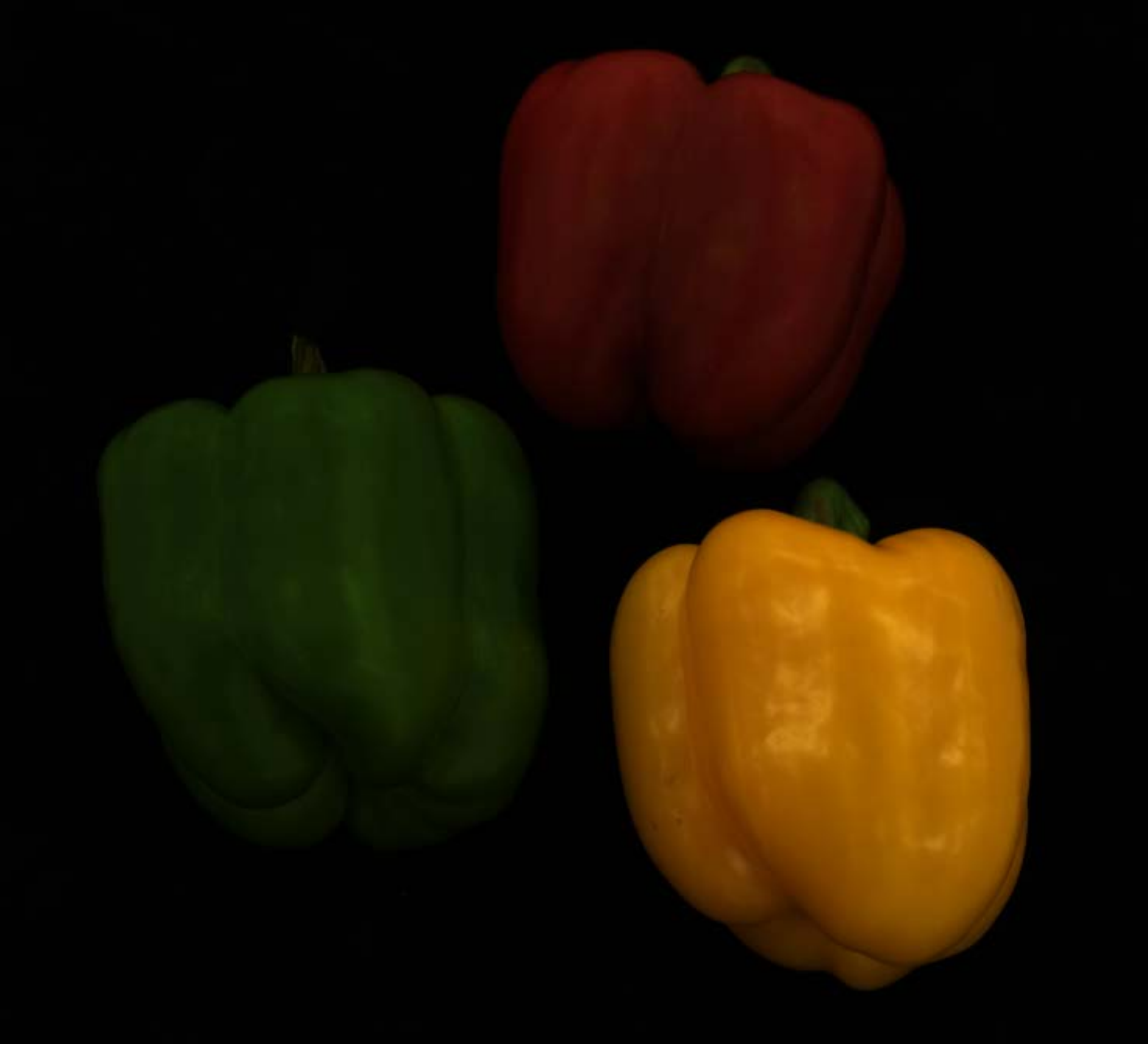}
      \includegraphics[width=1.0\linewidth]{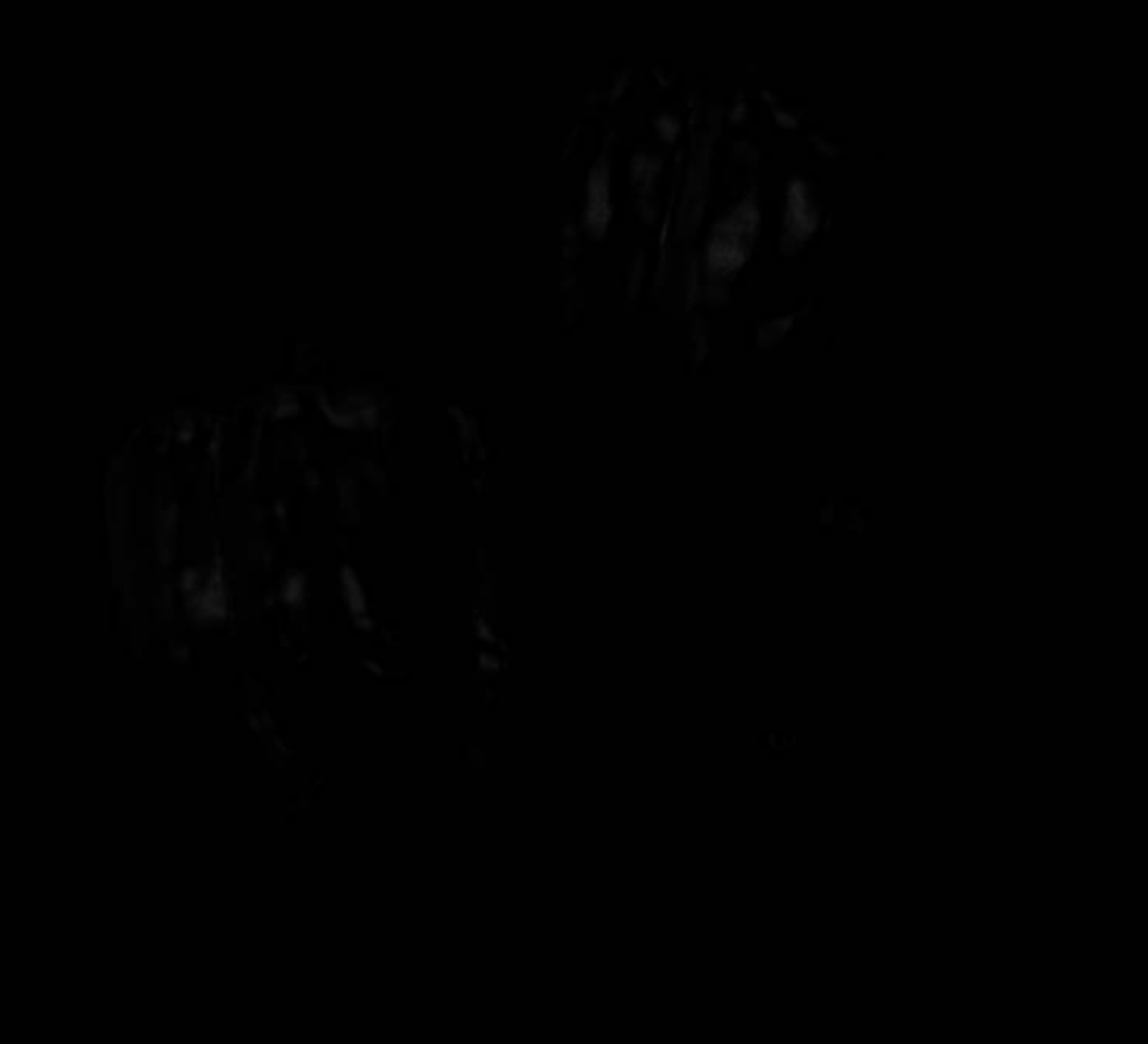}\vspace{1mm}
      \includegraphics[width=1.0\linewidth]{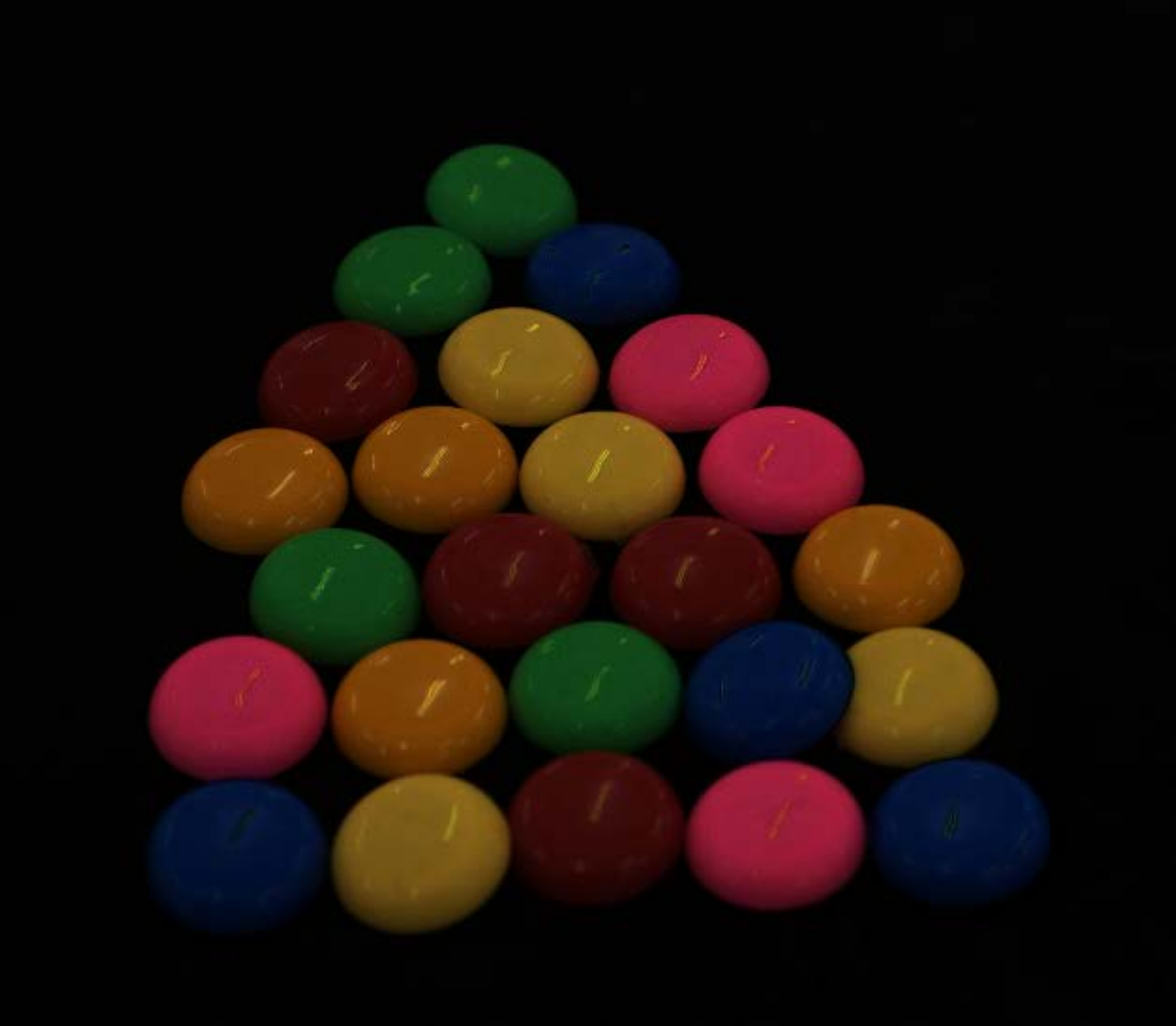}
      \includegraphics[width=1.0\linewidth]{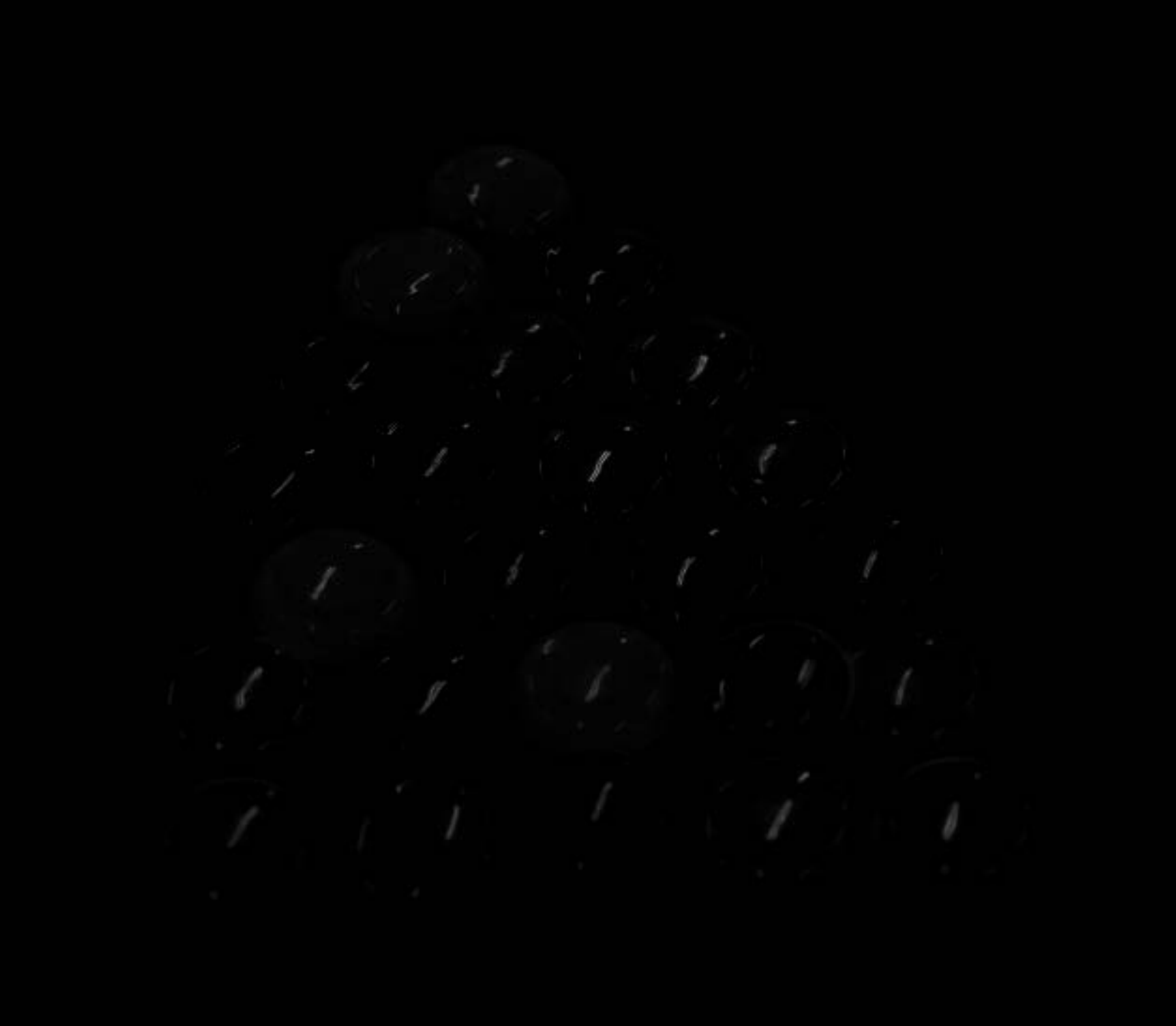}\vspace{2mm}
 \end{minipage}}\hspace{-0.1mm}
 \subfigure[]{
 \begin{minipage}[ht]{0.188\textwidth}
      \includegraphics[width=1.0\linewidth]{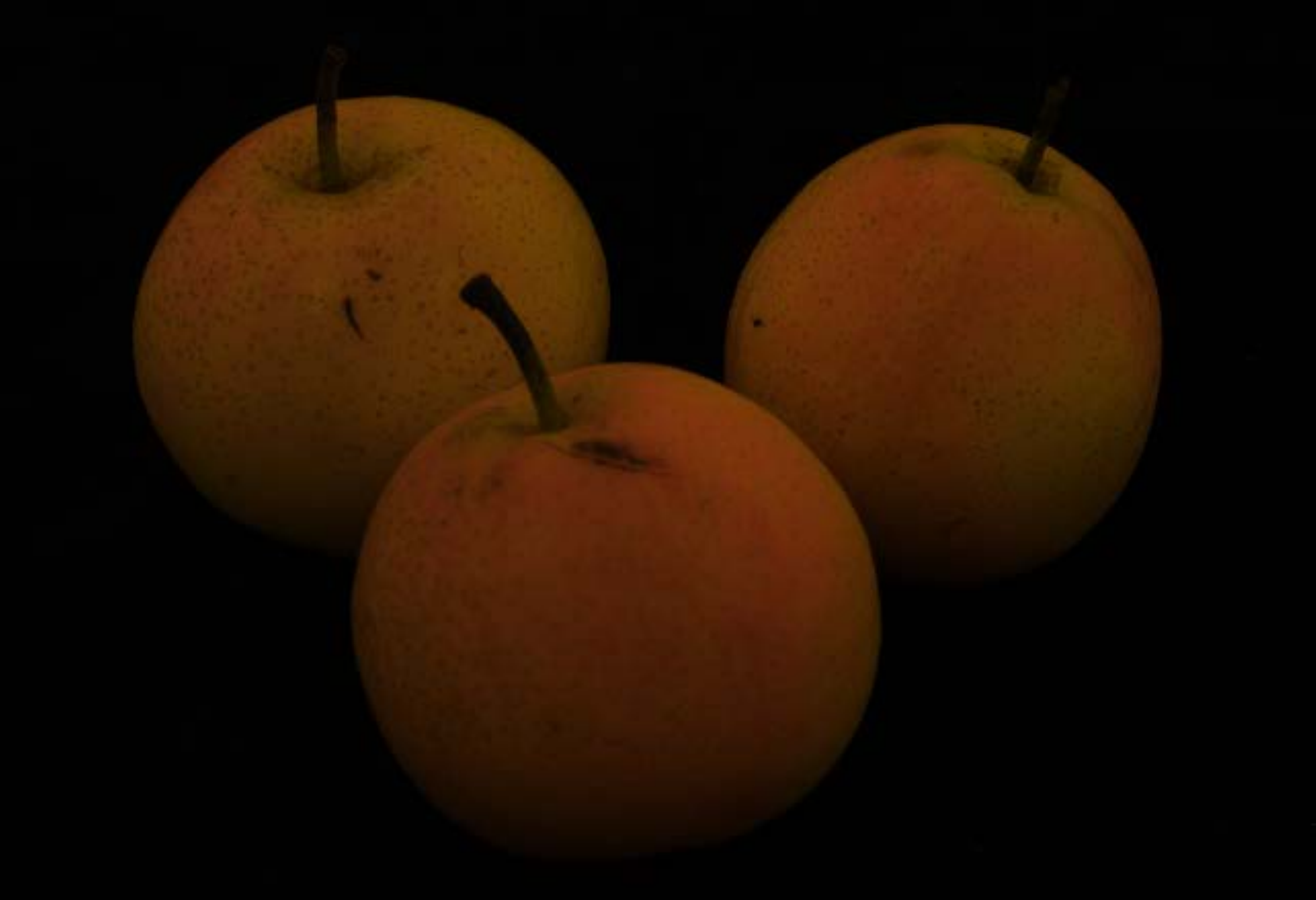}
      \includegraphics[width=1.0\linewidth]{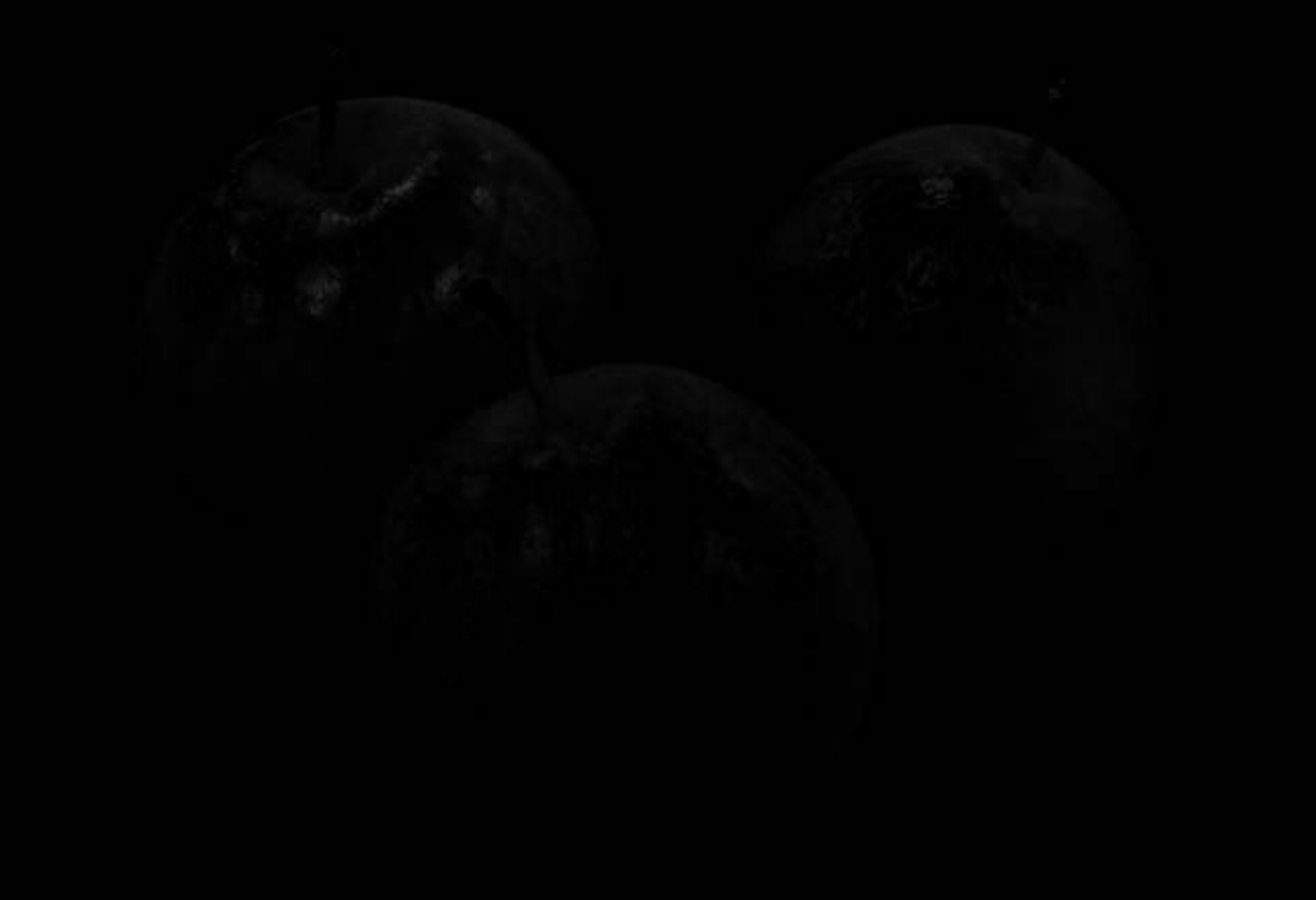}\vspace{1mm}
      \includegraphics[width=1.0\linewidth]{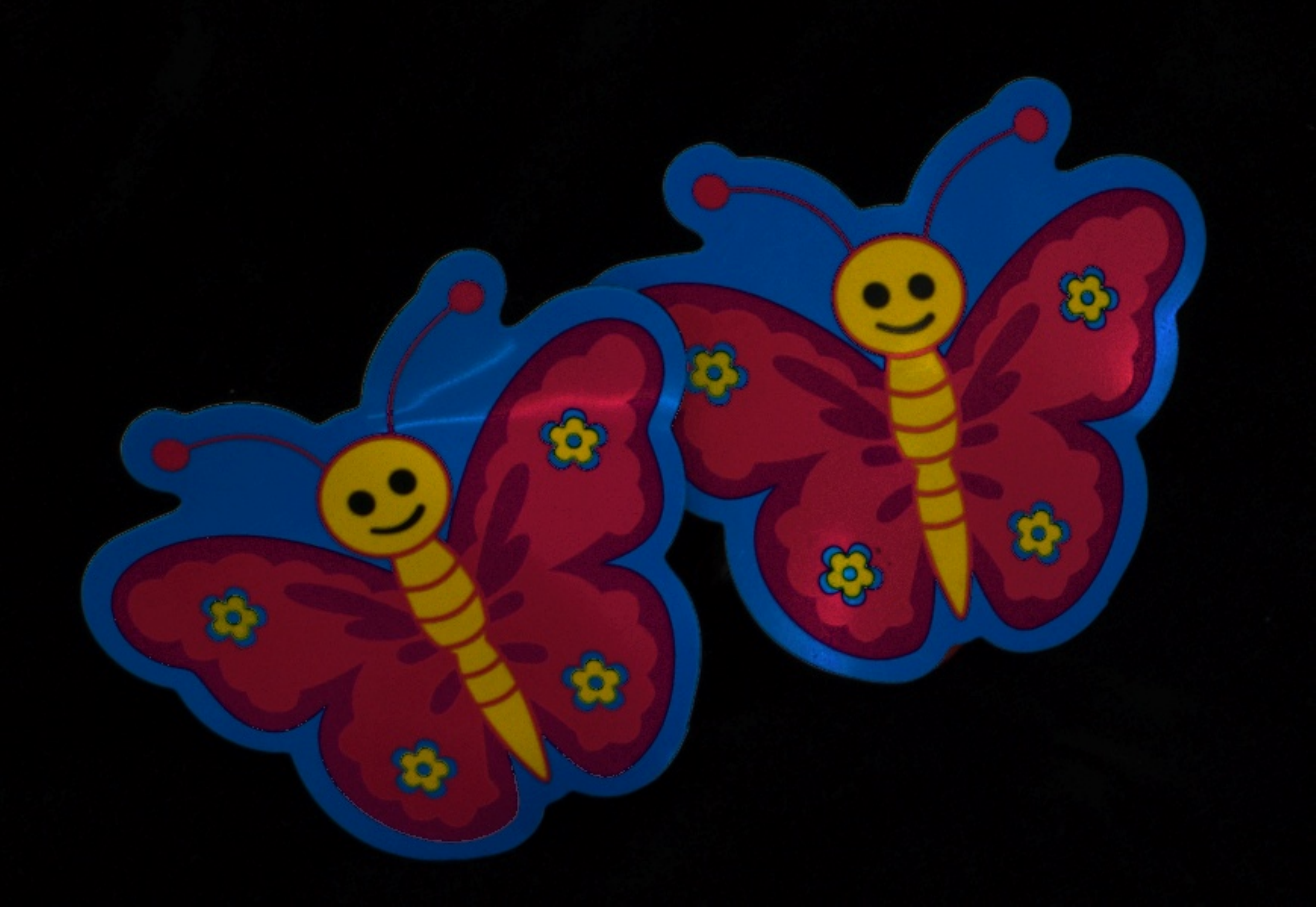}
      \includegraphics[width=1.0\linewidth]{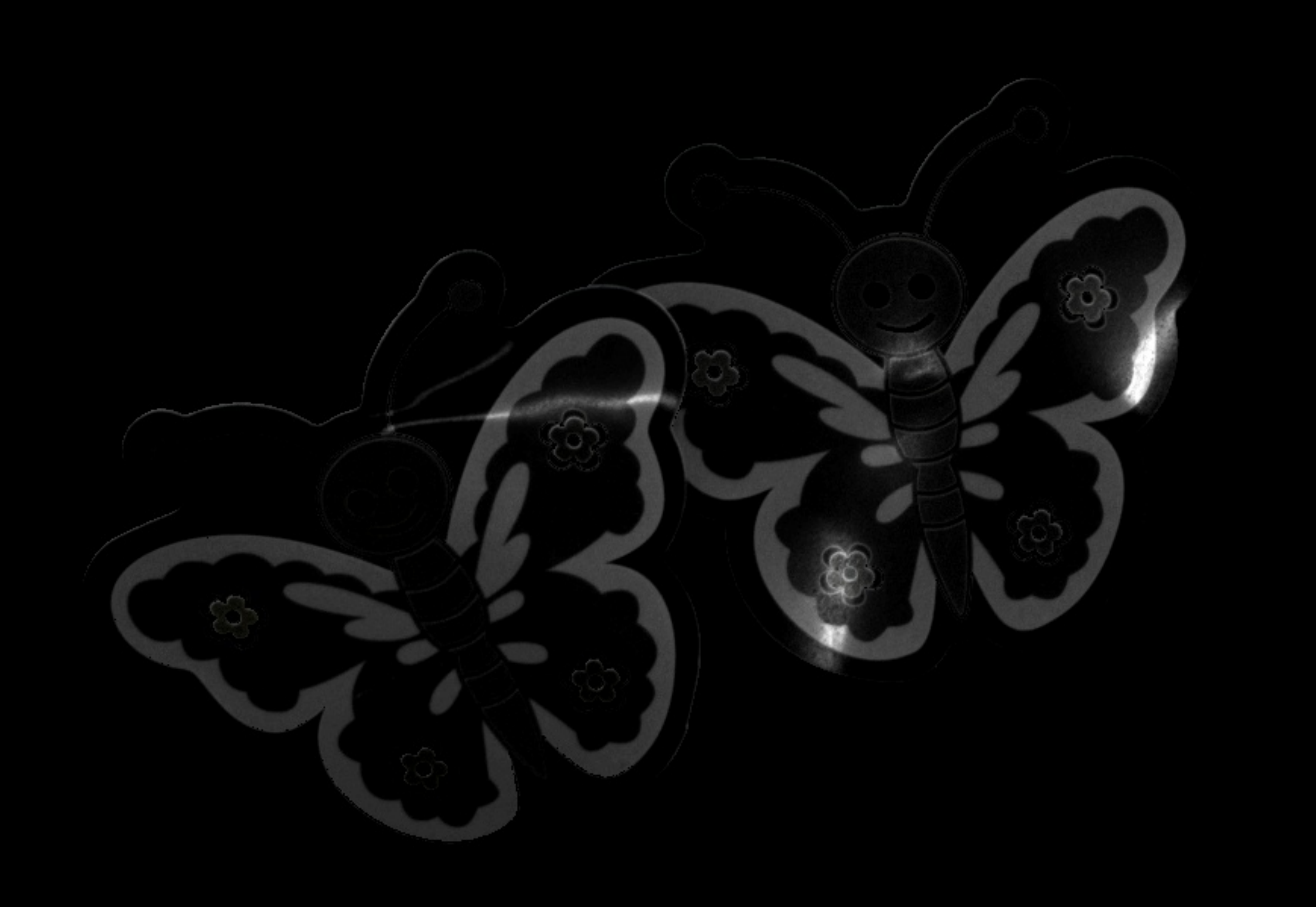}\vspace{1mm}
      \includegraphics[width=1.0\linewidth]{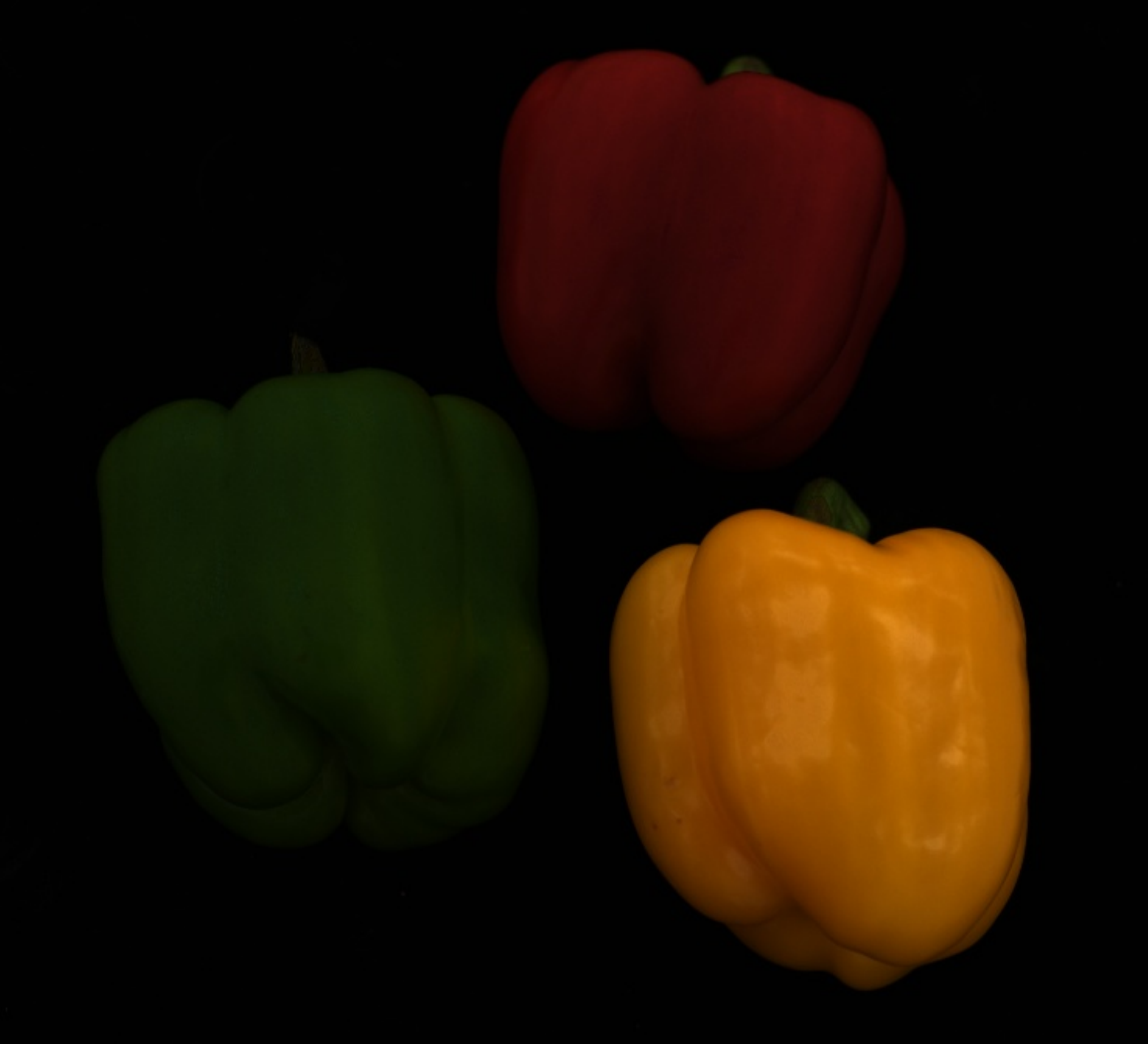}
      \includegraphics[width=1.0\linewidth]{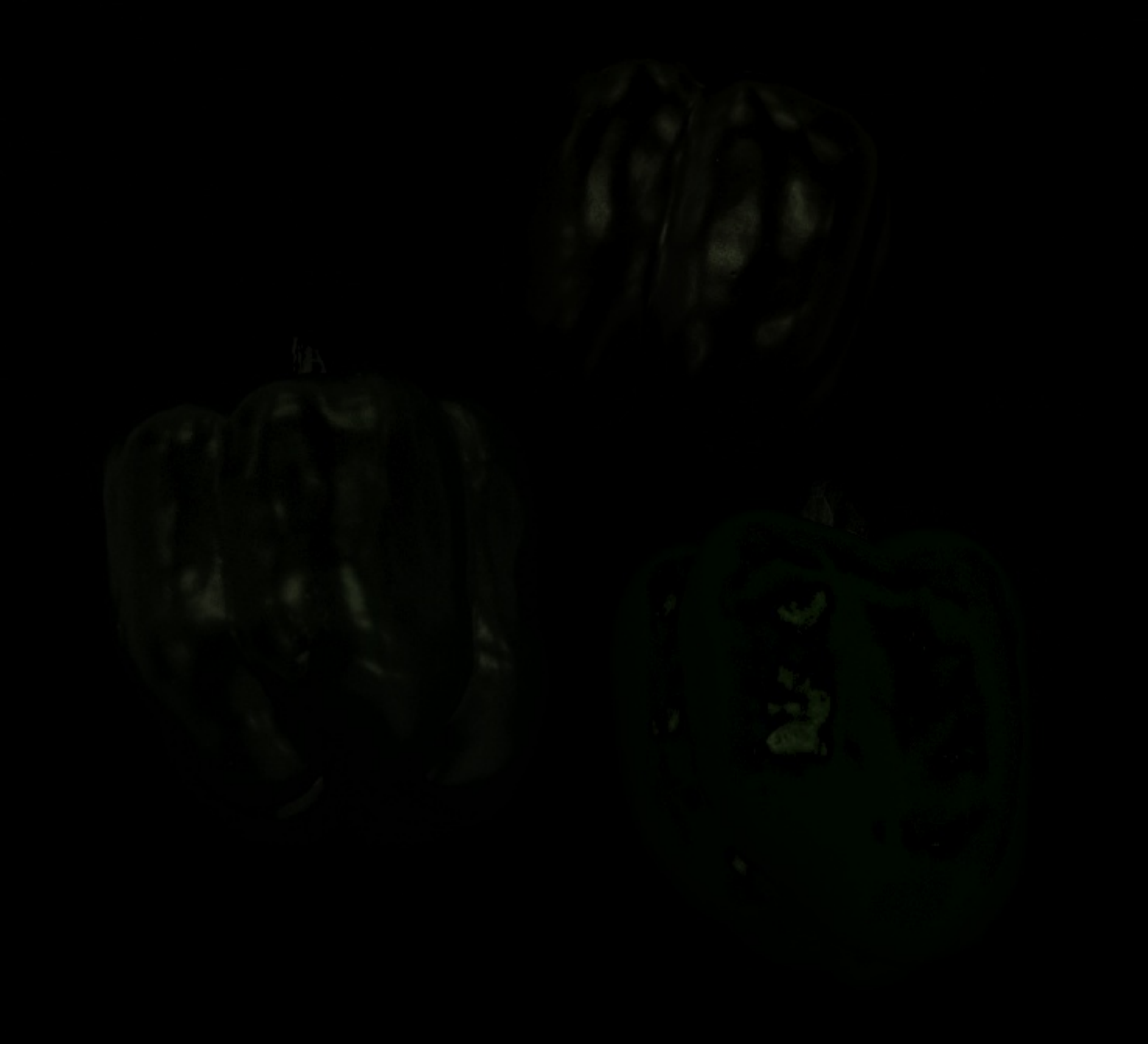}\vspace{1mm}
      \includegraphics[width=1.0\linewidth]{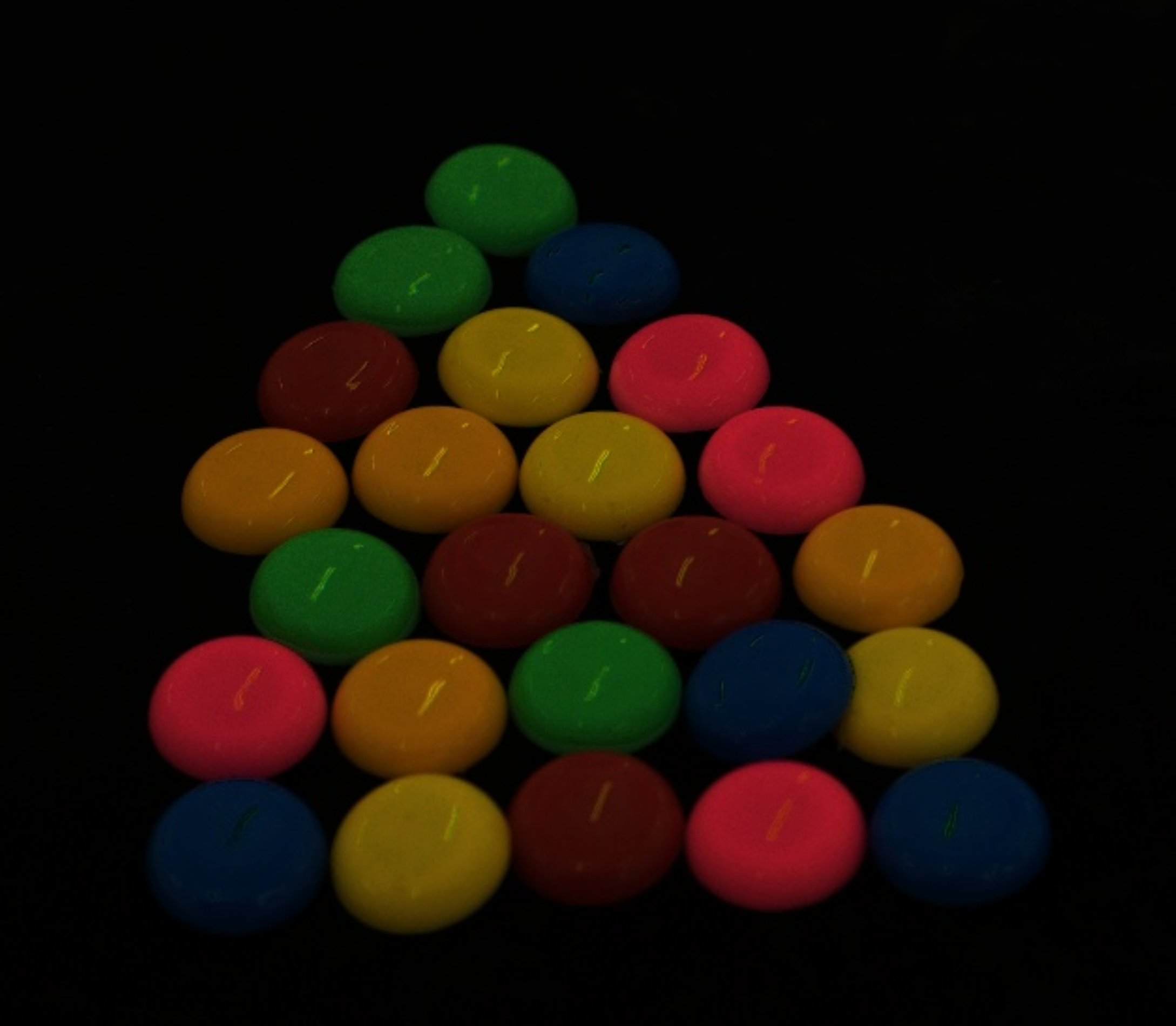}
      \includegraphics[width=1.0\linewidth]{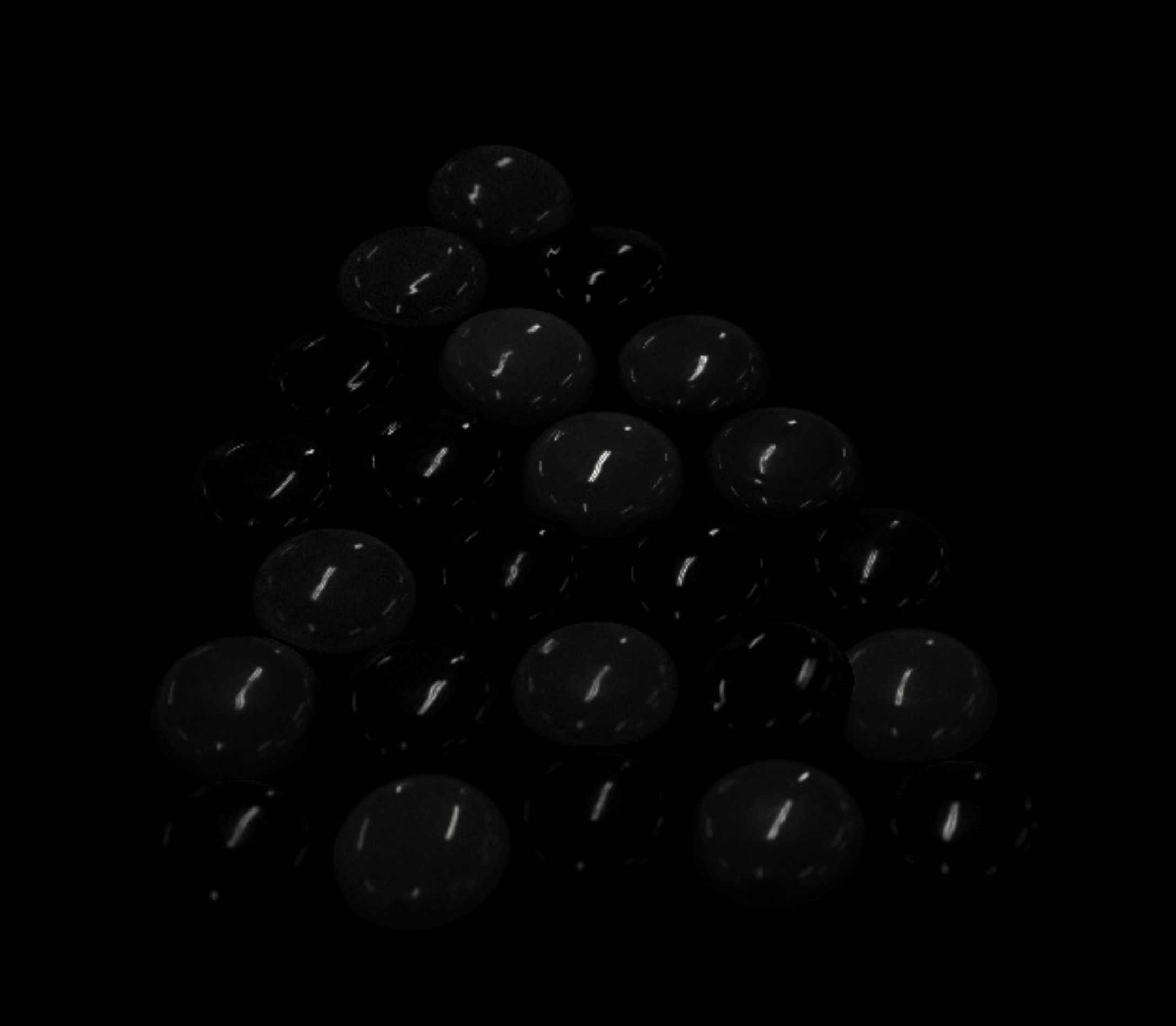}\vspace{2mm}
 \end{minipage}}
 \centering
  \caption{Comparison between our approach and three state-of-the-arts, both diffuse and  specular components are displayed. (a) Inputs. (b) Our results. (c) Akashi and Okatani's\cite{Akashi14}. (d) Shen et al.'s\cite{Shen13}. (e) Yang et al.'s\cite{Yang01}. }
  \label{fig:our_image}
\end{figure*}

In this experiment, we run our algorithm on a variety of nature images affected by specularity. Fig.~\ref{fig:our_image} displays our results in comparison with Akashi et al. \cite{Akashi14}'s, Shen et al. \cite{Shen13}'s and Yang et al. \cite{Yang01}'s algorithms. All the images are captured by ourselves using a Nikon D7000 camera with a 50mm $f$/1.8D lens. Demosaiced 16-bit raw data is used as input. The color of the illuminations are all normalized into white illumination by channel-wise division.

\begin{table*}[t]
\centering
\caption{\label{comparison}Comparison of running time among our algorithm, Yang et al.'s \cite{Yang01} and Shen et al.'s \cite{Shen13}}
\begin{tabular}{p{3cm}<{\centering}|p{3.6cm}<{\centering}|p{3cm}<{\centering}|p{3cm}<{\centering}|p{3cm}<{\centering}}
\hline
\multirow{2}{*}{Scenes} &Resolution & Ours (s)& Shen et al.\cite{Shen13} (s) & Yang et al.\cite{Yang01}(s)\\
 & (pixels) & Matlab & C++ & C++ \\
\hline
\multirow{2}{*}{Apples} & $650\times 450$ &  0.011 &  0.046 & 0.097\\
\cline{2-5}
&  $1950\times1350$ & 0.195 & 0.855 & 1.722\\
\hline
\multirow{2}{*}{Magnets} & $550\times 630$ &  0.022 &  0.023 & 0.089\\
\cline{2-5}
&  $1650\times1890$ & 0.119 & 0.199 & 0.625\\
\hline
\multirow{2}{*}{Butterfly} & $500\times 720$ &  0.020 &  0.026 & 0.100\\
\cline{2-5}
&  $1500\times2160$ & 0.110 & 0.189 & 0.809\\
\hline
\multirow{2}{*}{Peppers} & $630\times740$ &  0.020  &  0.031 & 0.157\\
\cline{2-5}
&  $1890\times2220$ & 0.084 & 0.273 & 0.883\\
\hline
\end{tabular}
\end{table*}

\begin{figure}[h]
\centering
\subfigure[]{
\begin{minipage}[ht]{0.232\textwidth}
 \includegraphics[width=0.91\linewidth]{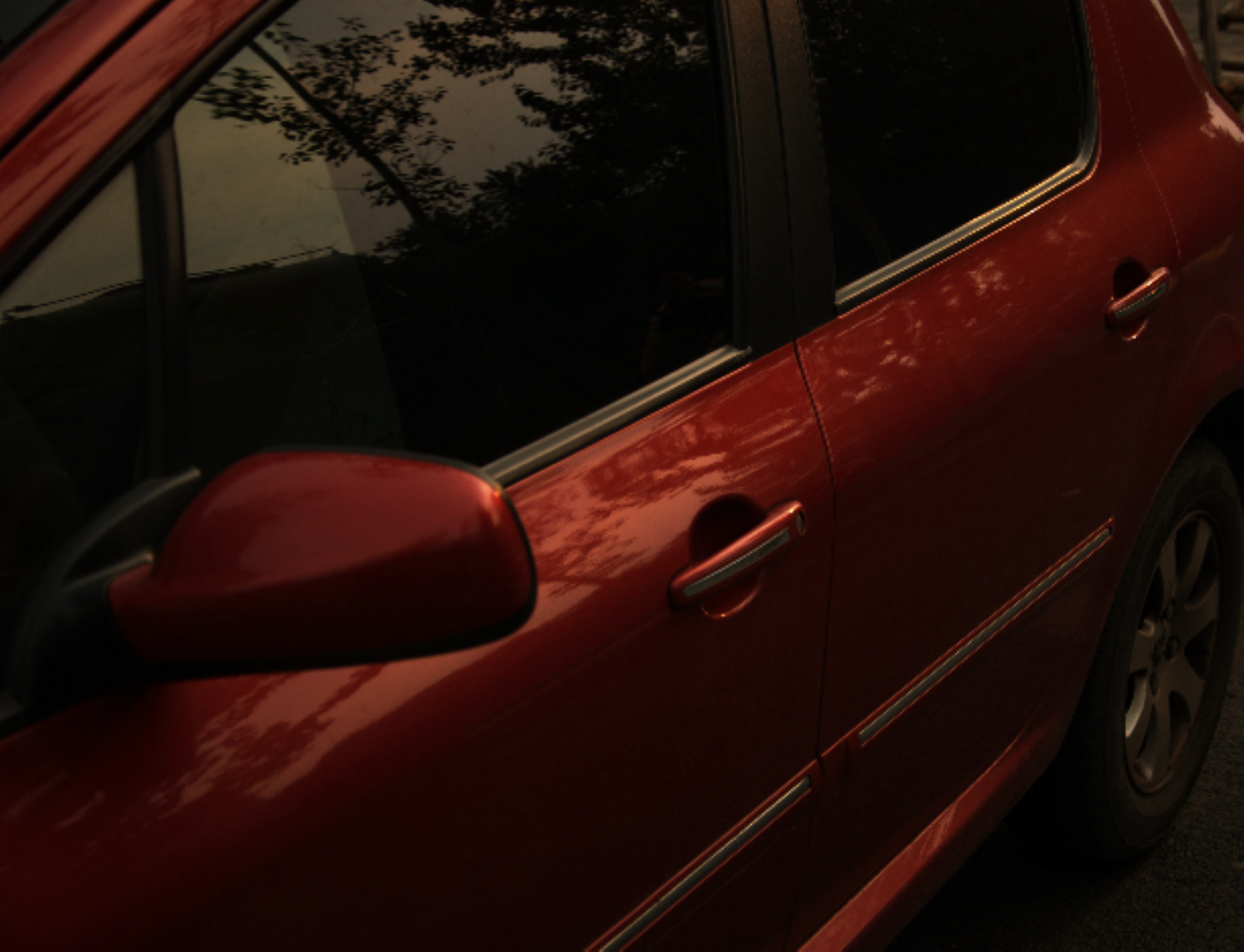}\vspace{1mm}\\
  \includegraphics[width=0.91\linewidth]{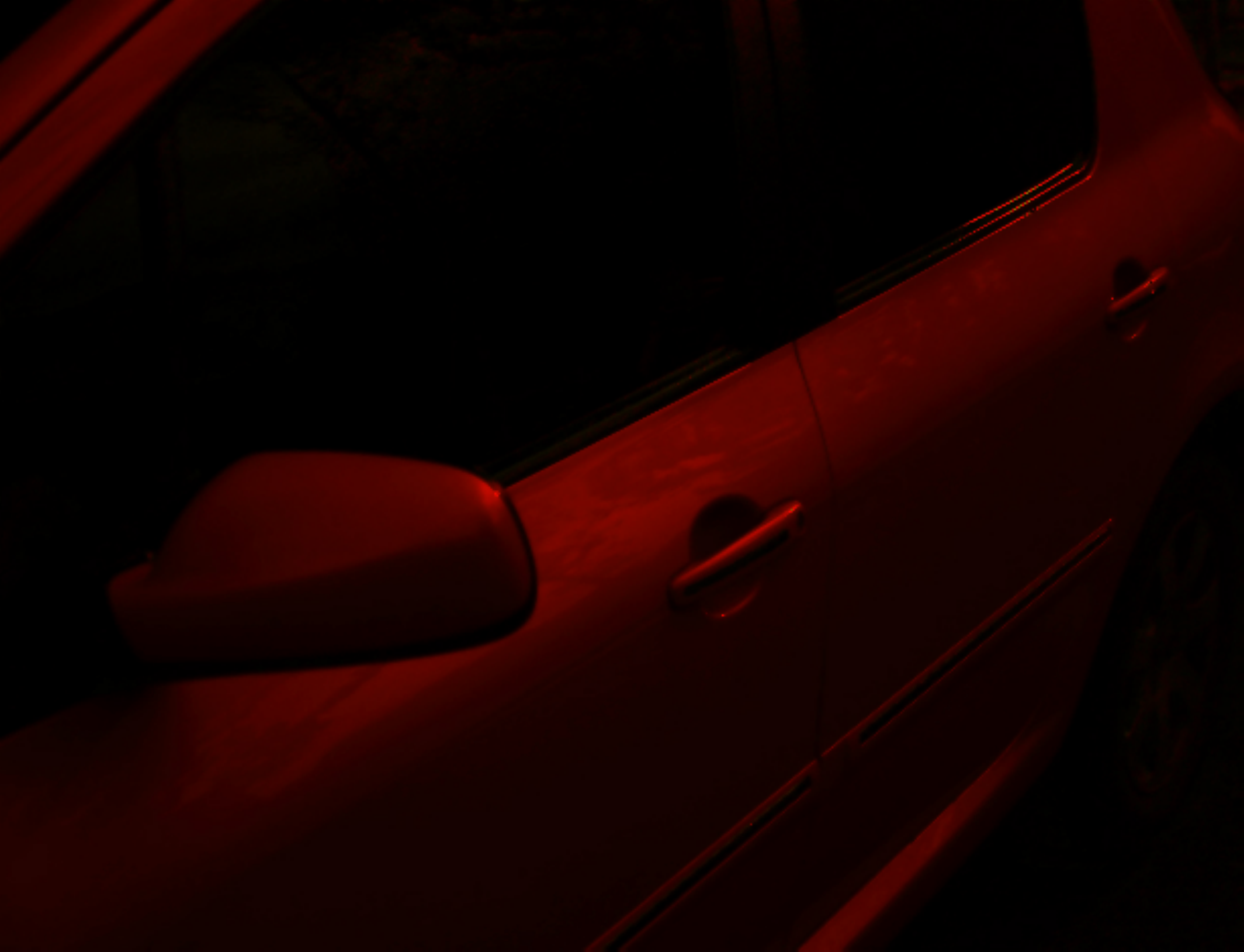}\vspace{1mm}\\
  \includegraphics[width=0.91\linewidth]{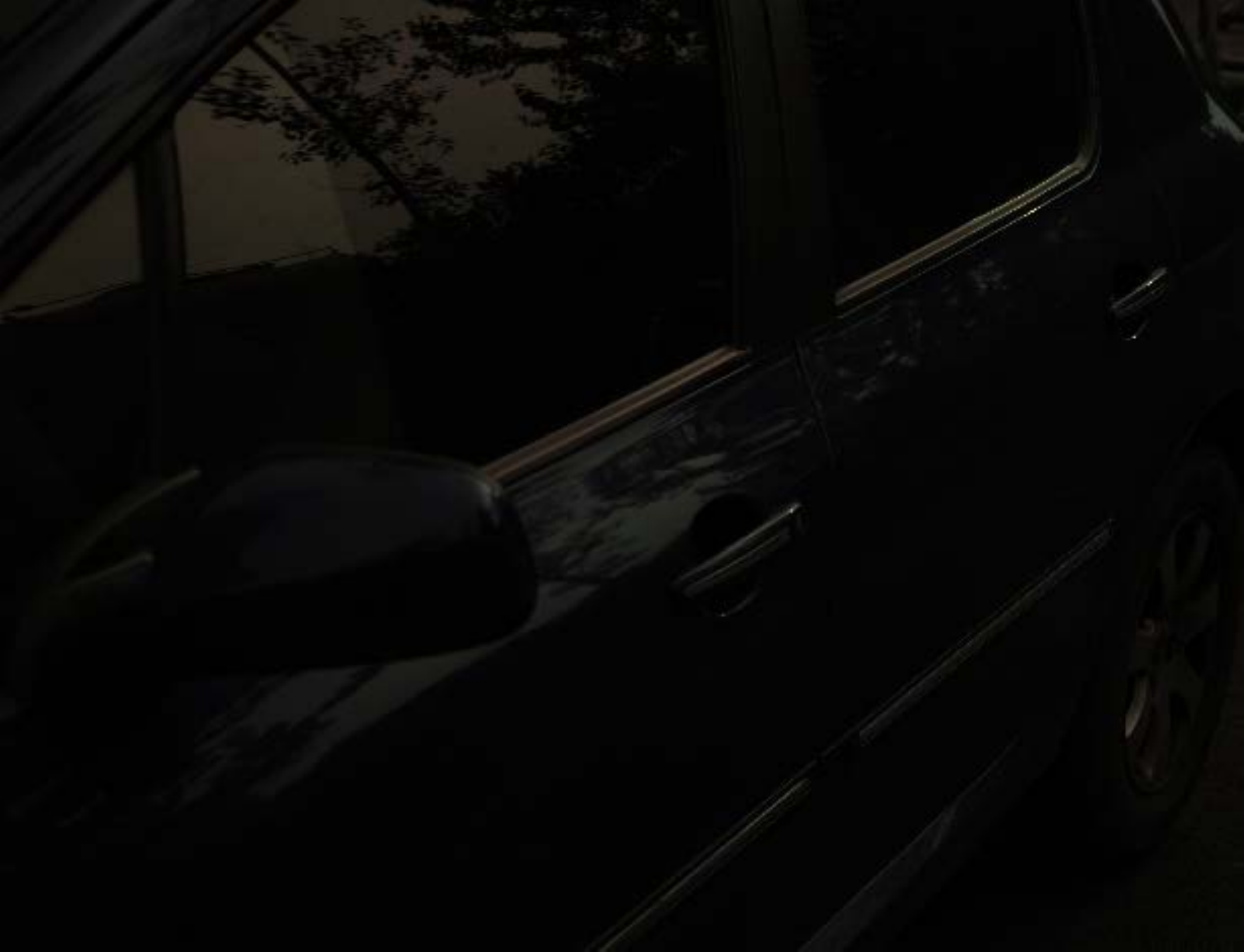}\\
 \end{minipage}}
\subfigure[]{
\begin{minipage}[ht]{0.232\textwidth}
 \includegraphics[width=1\linewidth]{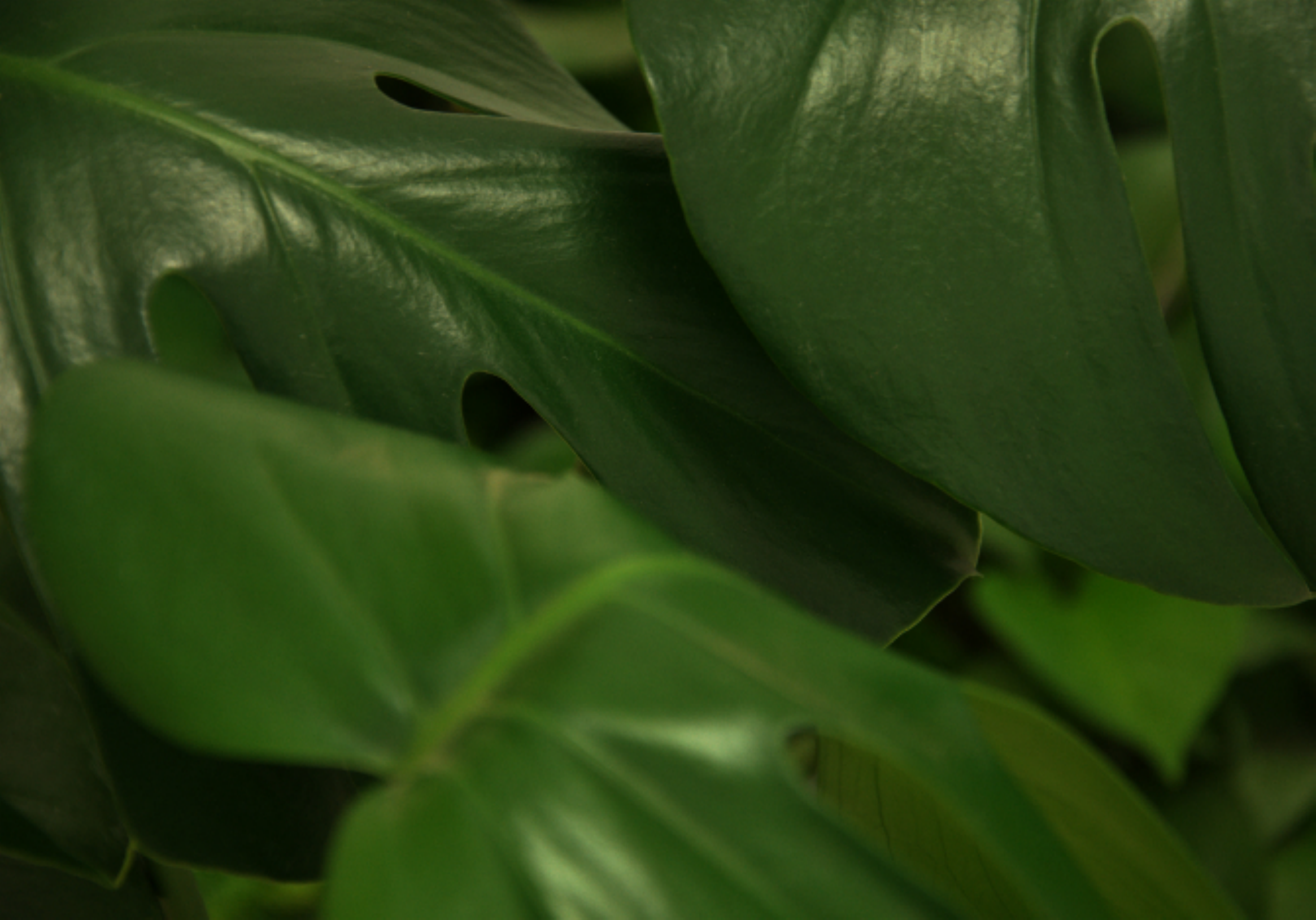}\vspace{1mm}\\
  \includegraphics[width=1\linewidth]{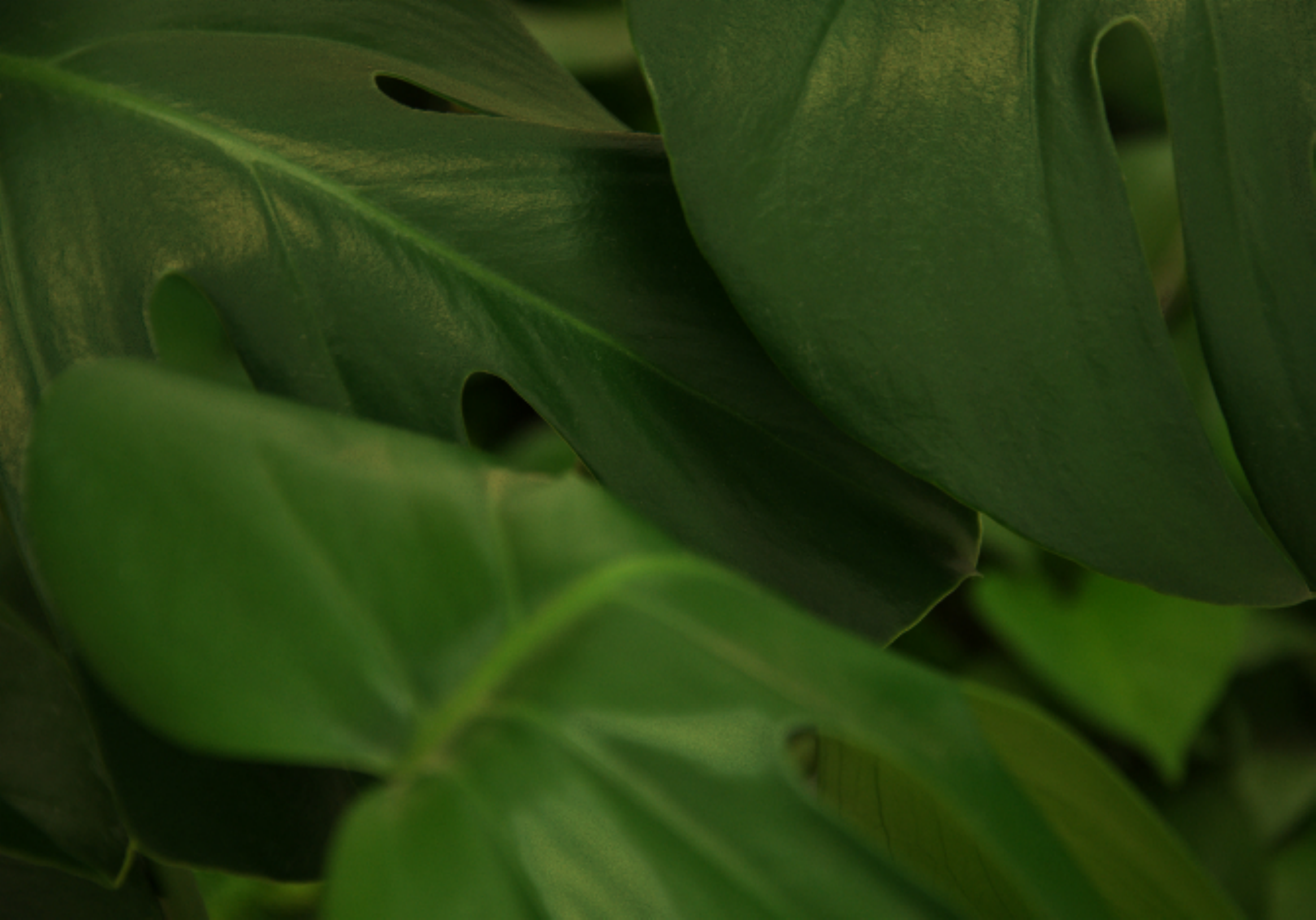}\vspace{1mm}\\
  \includegraphics[width=1\linewidth]{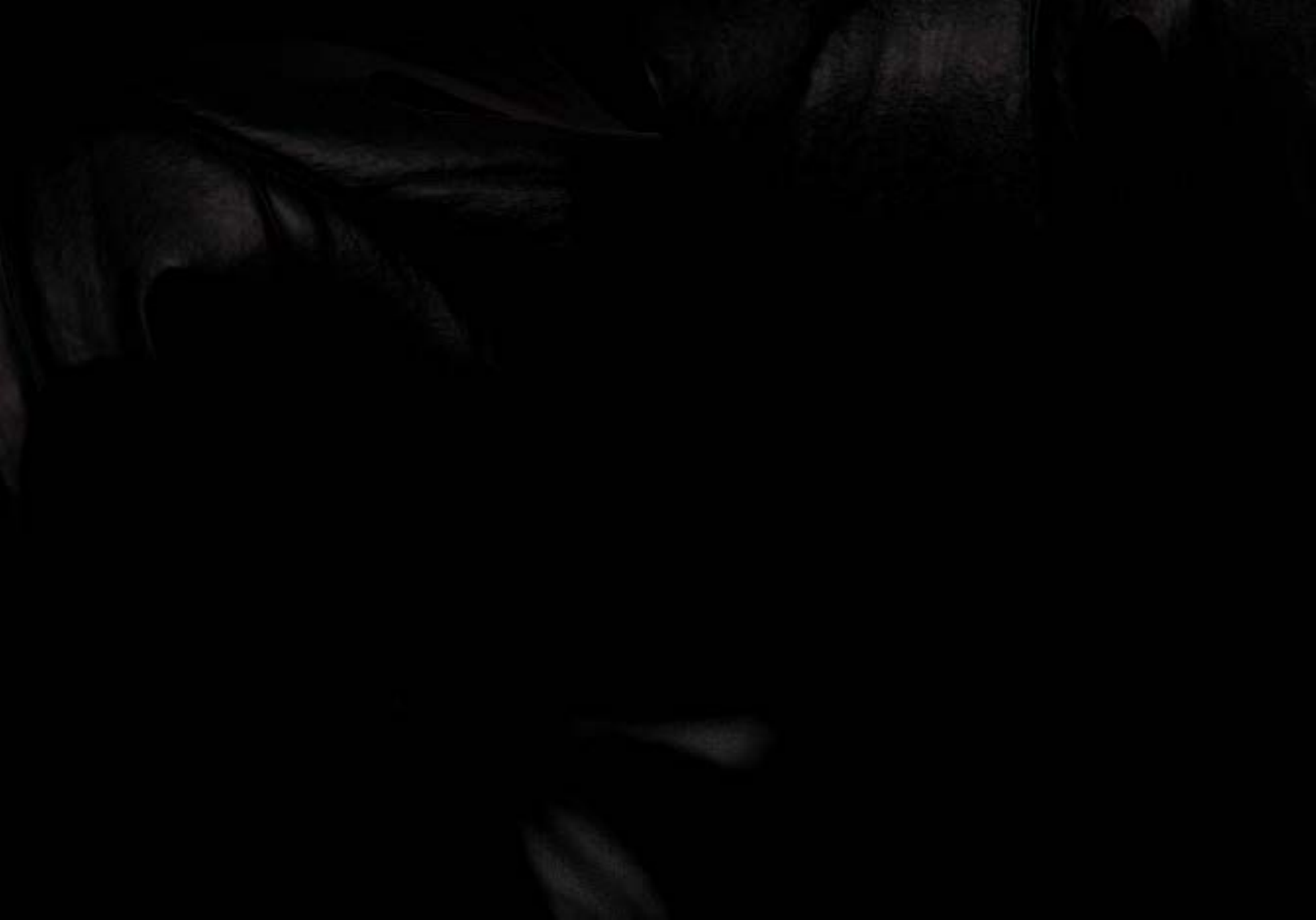}\\
 \end{minipage}}
  \caption{Results on two outdoor natural scenes. (a) Car. (b) Plant. From top to bottom: input image, diffuse and specular component.}
  \label{fig:outdoor}
\end{figure}

Overall, the algorithm produces promising visual results and this validates its effectiveness in the real non-Lambertian scenes. The slight difference with STAR algorithms indicates our superiority in some challenging cases.
For images with slight and small scale specularity, such as the 'Apples' scene in the top row, all the four methods can give good separation results.
In the 'Butterfly' scene, the chromaticity of the pink wing region ($R$=0.7715, $G$=0.3505, $B$=0.5330) is of high similarity to the normalized white illumination ($R$=0.5774, $G$=0.5774, $B$=0.5774) and there is some color bias in \cite{Akashi14}'s, \cite{Shen13}'s and \cite{Yang01}'s results. In contrast, we can still recover the diffuse component correctly.
Besides, our algorithm exhibits superior performance in handling large area highlights on the glossy surface, such as the highlights on the 'Peppers' surface. Akashi and Okatani's approach\cite{Akashi14} also performs well on this example.  In contrast, Shen et al.'s\cite{Shen13} and Yang et al.\cite{Yang01}'s algorithms cannot  handle this situation,  because the assumption in \cite{Shen13} that the number of the specular pixels below a certain threshold is violated in this case, and the propagation strategy adopted in \cite{Yang01} only applies to small local specular regions.
Strong specularity when the surface approaches to mirror is another challenging case for highlight removal, e.g., the color 'Magnets' scene in the bottom row. In this example, the methods in \cite{Akashi14}, \cite{Shen13} and \cite{Yang01} are all incapable of removing the specular component cleanly. Instead, our algorithm can give decent separation if the specular region is not caused by pure specular reflection. We can discriminate the subregions with and without inter-reflection on the same magnet. Besides, we also test our approach on two outdoor scenes to demonstrate its wide availability, as shown in Fig.~\ref{fig:outdoor}.

\subsection{Running Time Analysis and Comparison}
Our highlight removal algorithm does not involve complex calculations, and thus of high efficiency in terms of both storage and computation. We test the efficiency on an Intel Xeon 2.27 GHz CPU workstation with 64 bit Windows 7 system. Roughly, processing a 500$\times$600 pixel image generally takes 0.02s, and there exists slight variation due to the diverse number of materials in different images. The most time consuming module is clustering in our approach. Although the adopted K-Means clustering will slow down at extremely high resolution, we can easily handle such cases via down sampling strategy. Specifically, we first down-sample the original image to a lower version (e.g., $200\times200$ pixels), which contains all the materials in the original one. Then we apply the proposed method to the low resolution image to conduct material clustering, and recover the diffuse chromaticity of each cluster. Finally, for each pixel in the high resolution input image, the cluster label and corresponding diffuse chromaticity is assigned as those of the cluster with the highest correlation in the specular-free space.

In Table \ref{comparison}, we compare our running time with the fast highlight removal approach proposed by Shen et al.\cite{Shen13} and Yang et al.\cite{Yang01}. Since optimization based methods are usually of much higher computation cost, we omit comparison with these methods here. From the data we can see that, even implemented with Matlab, our algorithm is still slightly faster than the other two methods implemented with C++. Benefiting from the down sampling strategy, our efficiency superiority is more prominent at higher resolution.

\section{Conclusions and Discussions}
This paper proposes a new highlight removal approach by defining an $L_2$ normalized dichromatic model and deriving a strict formulation to separate the diffuse and specular components.
Without mathematical approximations and strong assumptions on the scene, the approach can handle a large diversity of cases that fail the previous methods.
Besides, the proposed approach involves few complex calculations and thus can achieve fast processing.

\begin{figure}[t]
\centering
\subfigure[]{
\begin{minipage}[ht]{0.23\textwidth}
 \includegraphics[width=0.95\linewidth]{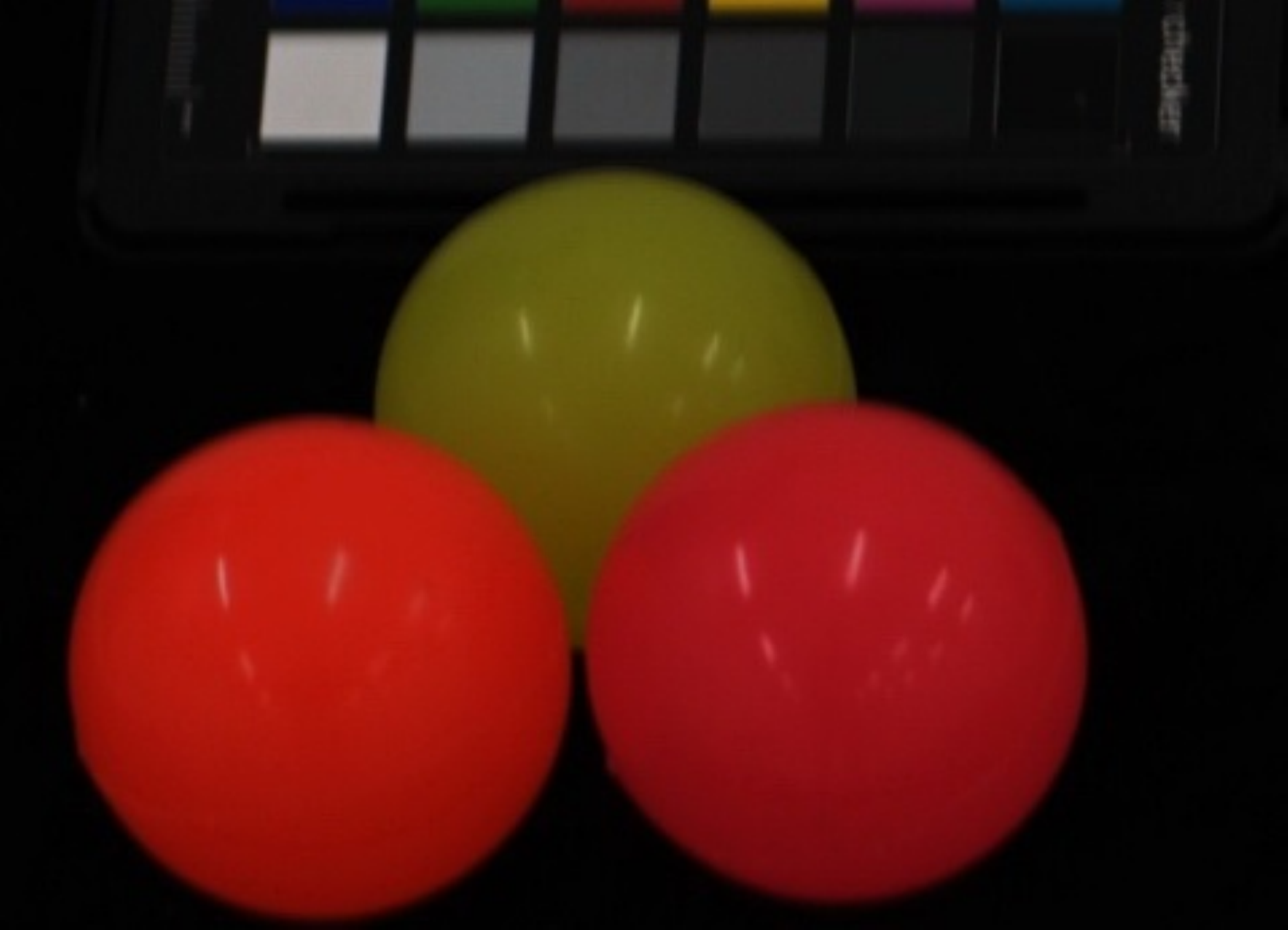}\vspace{1mm}\\
  \includegraphics[width=0.95\linewidth]{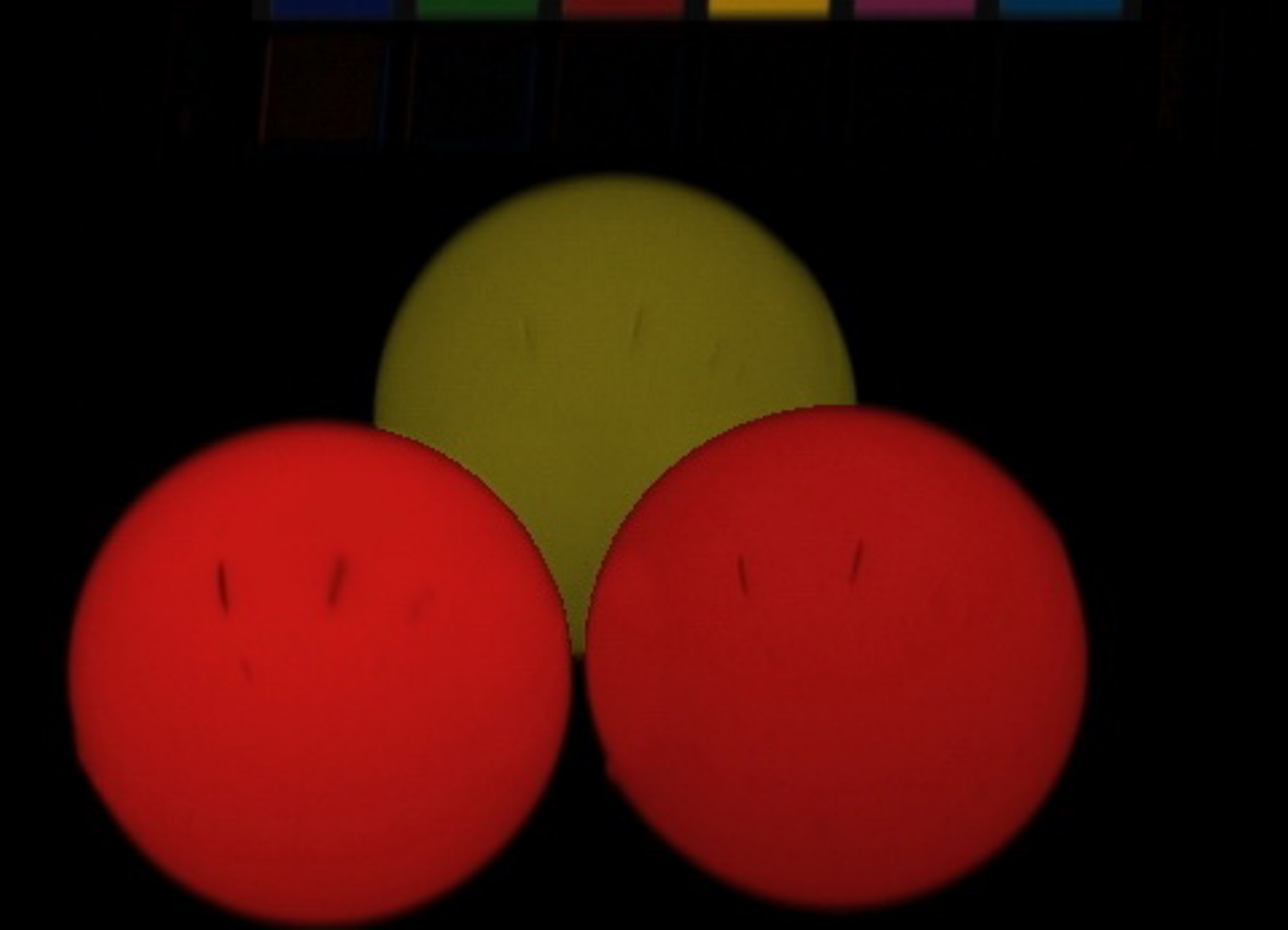}\vspace{1mm}\\
  \includegraphics[width=0.95\linewidth]{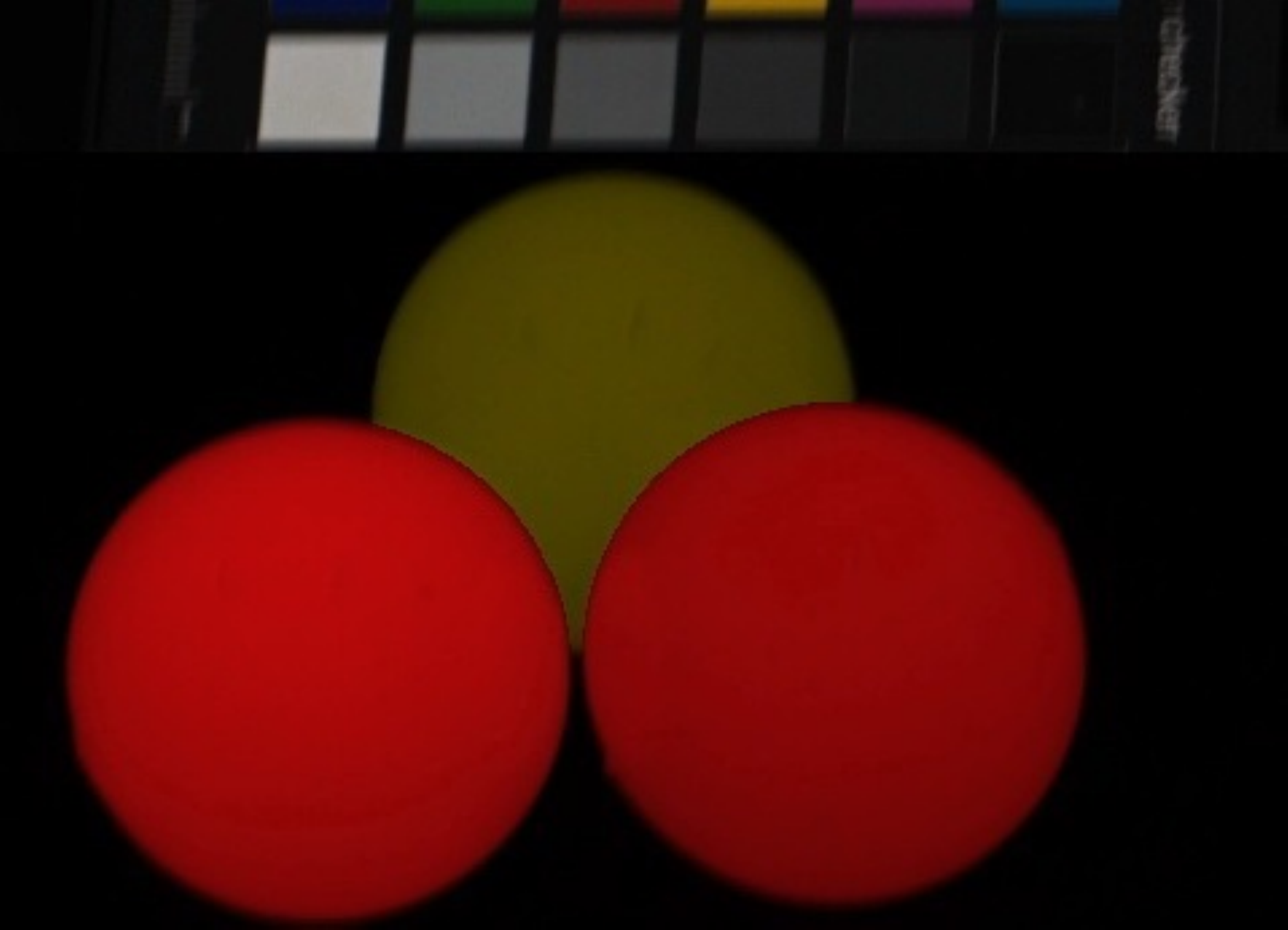}\\
 \end{minipage}}
\subfigure[]{
\begin{minipage}[ht]{0.23\textwidth}
 \includegraphics[width=0.95\linewidth]{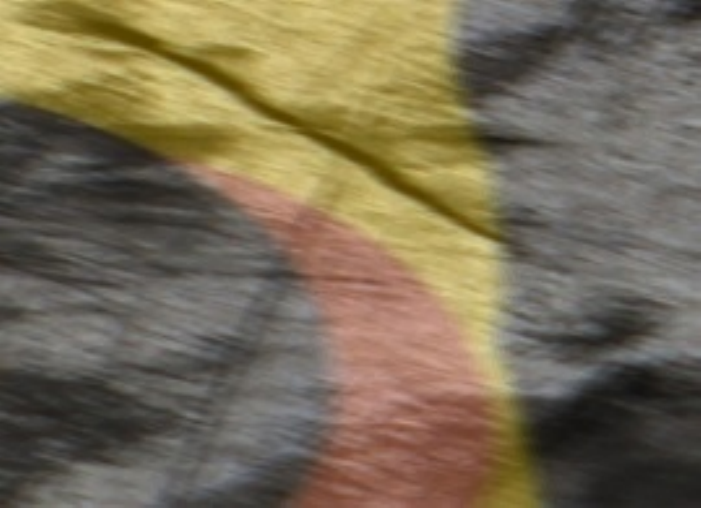}\vspace{1mm}\\
  \includegraphics[width=0.95\linewidth]{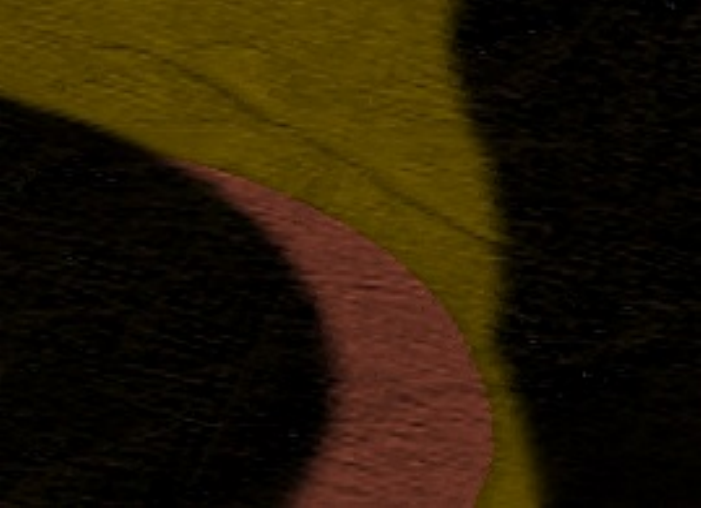}\vspace{1mm}\\
  \includegraphics[width=0.95\linewidth]{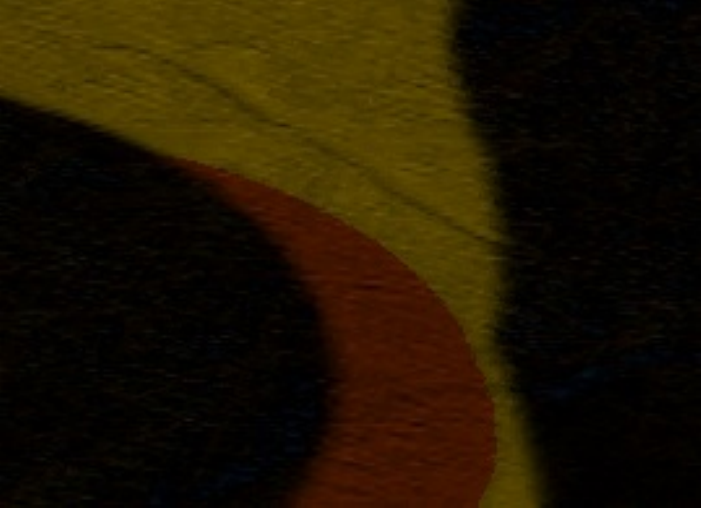}\\
 \end{minipage}}
  \caption{Results in some extreme cases. (a) Gray colors and pure specular reflection. (b) Image wholly covered by highlights. {Top}: input image. {Middle}: highlight removal result. {Bottom}: benchmark diffuse component captured by applying a polarizer.}
  \label{fig:limitation}
\end{figure}

\vspace{2mm}
\noindent{\em Limitations.~~}
Although being widely  applicable, our approach may produce artifacts in some extreme cases. Firstly, the diffuse component of gray (including white) objects or pure specular regions {may lose energy because the coefficients are zero}, such as the mirror reflection on the rubber balls and the gray scale tiles in the color checker in Fig.~\ref{fig:limitation}(a). Such cases are beyond the scope of the dichromatic model and addressing these cases needs user interaction or inpainting processing. {Our method cannot distinguish pixels with the same hue but different saturations either.} Secondly, we assume that for each kind of material, there exist pure diffuse pixels. For a scene totally covered by highlights, such as the example in the first column of Fig.~\ref{fig:limitation}(b), the specularity can be largely reduced but the recovery results is still slightly affected by the illumination. As demonstrated in the comparison between the recovery in the middle column and the benchmark obtained by applying a polarizer in the right column. Highlight removal in such cases is intrinsically illposed for all the methods on dichromatic model, and beyond the scope of this paper. 

\vspace{2mm}
\noindent{\em Future extensions.~~}
We plan to extend current approach to remove specularity in videos. Utilizing the temporal redundancy will further raise the performance and efficiency. For one thing, the sparse abrupt temporal changes will raise robustness to noise. For the other, the slight variation of material types between adjacent frames would accelerate processing, since we do not need to start from 1 cluster. {By taking some priors into consideration, such as the sparsity of materials in the scene, the separation of diffuse and specular component under unknown illumination is also possible.} Besides, highlight removal under multiple illuminations with different colors is also interesting and worth studying.
\ifCLASSOPTIONcaptionsoff
  \newpage
\fi

\bibliographystyle{IEEEtran}
\bibliography{MyBib_short}

\begin{thebibliography}{10}
\providecommand{\url}[1]{#1}
\csname url@samestyle\endcsname
\providecommand{\newblock}{\relax}
\providecommand{\bibinfo}[2]{#2}
\providecommand{\BIBentrySTDinterwordspacing}{\spaceskip=0pt\relax}
\providecommand{\BIBentryALTinterwordstretchfactor}{4}
\providecommand{\BIBentryALTinterwordspacing}{\spaceskip=\fontdimen2\font plus
\BIBentryALTinterwordstretchfactor\fontdimen3\font minus
  \fontdimen4\font\relax}
\providecommand{\BIBforeignlanguage}[2]{{%
\expandafter\ifx\csname l@#1\endcsname\relax
\typeout{** WARNING: IEEEtran.bst: No hyphenation pattern has been}%
\typeout{** loaded for the language `#1'. Using the pattern for}%
\typeout{** the default language instead.}%
\else
\language=\csname l@#1\endcsname
\fi
#2}}
\providecommand{\BIBdecl}{\relax}
\BIBdecl

\bibitem{Lellmann01}
J.~Lellmann, J.~Balzer, A.~Rieder, and J.~Beyerer, ``Shape from specular
  reflection and optical flow,'' \emph{International Journal of Computer
  Vision}, vol.~80, no.~2, pp. 226--241, 2008.

\bibitem{Artusi01}
A.~Artusi, F.~Banterle, and D.~Chetverikov., ``A survey of specularity removal
  methods,'' \emph{Computer Graphics Forum}, vol.~30, no.~8, pp. 2208--2230,
  2011.

\bibitem{Wolff01}
L.~Wolff, ``Polarization-based material classification from specular
  reflection,'' \emph{IEEE Transactions on Pattern Analysis and Machine
  Intelligence}, vol.~12, no.~11, pp. 1059--1071, 1990.

\bibitem{Nayar02}
S.~Nayar, X.~Fang, and T.~Boult, ``Separation of reflection components using
  color and polarization,'' \emph{International Journal of Computer Vision},
  vol.~21, no.~3, pp. 163--186, 1997.

\bibitem{Kim05}
D.~Kim, S.~Lin, K.~Hong, and H.~Shum, ``Variational specular separation using
  color and polarization,'' in \emph{Proceedings of the IAPR Workshop on
  Machine Vision Applications}, 2002.

\bibitem{Umeyama01}
S.~Umeyama and G.~Godin, ``Separation of diffuse and specular components of
  surface reflection by use of polarization and statistical analysis of
  images,'' \emph{IEEE Transactions on Pattern Analysis and Machine
  Intelligence}, vol.~26, no.~5, p. 639¨C647, 2004.

\bibitem{Feris01}
R.~Feris, R.~Raskar, K.-H. Tan, and M.~Turk, ``Specular reflection reduction
  with multi-flash imaging,'' in \emph{Proceedings of the IEEE Brazilian
  Symposium on Computer Graphics and Image Processing}, 2004.

\bibitem{Sato01}
Y.~Sato and K.~Ikeuchi, ``Temporal-color space analysis of reflection,''
  \emph{Journal of the Optical Society of America A}, vol.~11, no.~11, pp.
  2990--3002, 1994.

\bibitem{Lin02}
S.~Lin and H.~Shum, ``Separation of diffuse and specular reflection in color
  images,'' in \emph{Proceedings of International Conference on Computer Vision
  and Pattern Recognition}, 2001.

\bibitem{Lin01}
S.~Lin, Y.~Li, S.~Kang, X.~Tong, and H.~Shum, ``Diffuse-specular separation and
  depth recovery from image sequences,'' in \emph{Proceedings of European
  Conference on Computer Vision}, 2002.

\bibitem{Chen05}
T.~Chen, M.~Goesele, and H.-P. Seidel, ``Mesostructure from specularity,'' in
  \emph{Proceedings of International Conference on Computer Vision and Pattern
  Recognition}, 2006.

\bibitem{Weiss01}
Y.~Weiss, ``Deriving intrinsic images from image sequences,'' in
  \emph{Proceedings of International Conference on Computer Visionn}, 2001.

\bibitem{Qingxiong01}
Q.~Yang, S.~Wang, N.~Ahuja, and R.~Yang, ``A uniform framework for estimating
  illumination chromaticity, correspondence, and specular reflection,''
  \emph{IEEE Transactions on Image Processing}, vol.~20, no.~1, pp. 53--63,
  2011.

\bibitem{Tan01}
R.~Tan, K.~Nishino, and K.~Ikeuchi, ``Illumination chromaticity estimation
  using inverse-intensity chromaticity space,'' in \emph{Proceedings of
  International Conference on Computer Vision and Pattern Recognition}, 2003.

\bibitem{Drew14}
M.~S. Drew, H.~R.~V. Joze, and G.~D. Finlayson, ``The zeta-image, illuminant
  estimation, and specularity manipulation,'' \emph{Computer Vision and Image
  Understanding}, vol. 127, pp. 1--13, 2014.

\bibitem{Lilong01}
L.~Shi and B.~Funt, ``Dichromatic illumination estimation via {H}ough
  transforms in {3D},'' in \emph{Proceedings of International Conference on
  Computer Graphics, Imaging and Visualization}, 2008.

\bibitem{Bajcsy01}
R.~Bajcsy, S.~Lee, and A.~Leonardis, ``Detection of diffuse and specular
  interface reflections and inter-reflections by color image segmentation,''
  \emph{International Journal of Computer Vision}, vol.~17, no.~3, pp.
  241--272, 1996.

\bibitem{Klinker02}
G.~Klinker, S.~Shafer, and T.~Kanade, ``The measurement of highlights in color
  images,'' \emph{International Journal of Computer Vision}, vol.~2, no.~1, pp.
  7--32, 1988.

\bibitem{Pesal01}
P.~Koirala, P.~Pant, M.~Hauta-Kasari, and J.~Parkkinen, ``Highlight detection
  and removal from spectral image,'' \emph{Journal of the Optical Society of
  America A}, vol.~28, no.~11, pp. 2284--2291, 2011.

\bibitem{Tan03}
R.~Tan and K.~Ikeuchi, ``Separating reflection components of textured surfaces
  using a single image,'' \emph{IEEE Transactions on Pattern Analysis and
  Machine Intelligence}, vol.~27, no.~2, pp. 178--193, 2005.

\bibitem{Yang01}
Q.~Yang, S.~Wang, and N.~Ahuja, ``Real-time specular highlight removal using
  bilateral filtering,'' in \emph{Proceedings of European Conference on
  Computer Vision}, 2010.

\bibitem{Mallick01}
S.~Mallick, T.~Zickler, P.~Belhumeur, and D.~Kriegman, ``Specularity removal in
  images and videos: A {PDE} approach,'' in \emph{Proceedings of European
  Conference on Computer Vision}, 2006.

\bibitem{Mallick02}
S.~Mallick, T.~Zickler, D.~Kriegman, and P.~Belhumeur, ``Beyond lambert:
  reconstructing specular surfaces using color,'' in \emph{Proceedings of
  International Conference on Computer Vision and Pattern Recognition}, 2005.

\bibitem{Hyeongwoo01}
H.~Kim, H.~Jin, S.~Hadap, and I.~Kweon, ``Specular reflection separation using
  dark channel prior,'' in \emph{Proceedings of International Conference on
  Computer Vision and Pattern Recognition}, 2013.

\bibitem{Kaiming01}
K.~He, J.~Sun, and X.~Tang, ``Single image haze removal using dark channel
  prior,'' in \emph{Proceedings of International Conference on Computer Vision
  and Pattern Recognition}, 2009.

\bibitem{Akashi14}
Y.~Akashi and T.~Okatani, ``Separation of reflection components by sparse
  non-negative matrix factorization,'' in \emph{Proceedings of Asian Conference
  on Computer Vision}, 2014.

\bibitem{Tan02}
R.~Tan and K.~Ikeuchi, ``Reflection components decomposition of textured
  surfaces using linear basis functions,'' in \emph{Proceedings of
  International Conference on Computer Vision}, 2005.

\bibitem{Shen13}
H.~Shen and Z.~Zheng, ``Real-time highlight removal using intensity ratio,''
  \emph{Applied Optics}, vol.~52, no.~19, pp. 4483--4493, 2013.

\bibitem{Graham01}
G.~D. Finlayson and M.~S. Drew, ``4-sensor camera calibration for image
  representation invariant to shading, shadows, lighting, and specularities,''
  in \emph{Proceedings of International Conference on Computer Vision}, 2001.

\bibitem{Shafer01}
S.~A. Shafer, ``Using color to separate reflection components,'' \emph{Color
  research \& application}, vol.~10, no.~4, pp. 210--218, 1985.

\bibitem{Cong01}
C.~Huynh and A.~Robles-Kelly, ``A solution of the dichromatic model for
  multispectral photometric invariance,'' \emph{International Journal of
  Computer Vision}, vol.~90, no.~1, pp. 1--27, 2010.

\bibitem{Chein01}
C.-I. Chang, ``Orthogonal subspace projection ({OSP}) revisited: a
  comprehensive study and analysis,'' \emph{IEEE Transactions on Geoscience and
  Remote Sensing}, vol.~43, no.~3, pp. 502--518, 2005.

\end{thebibliography}


\begin{thebibliography}{10}
\providecommand{\url}[1]{#1}
\csname url@samestyle\endcsname
\providecommand{\newblock}{\relax}
\providecommand{\bibinfo}[2]{#2}
\providecommand{\BIBentrySTDinterwordspacing}{\spaceskip=0pt\relax}
\providecommand{\BIBentryALTinterwordstretchfactor}{4}
\providecommand{\BIBentryALTinterwordspacing}{\spaceskip=\fontdimen2\font plus
\BIBentryALTinterwordstretchfactor\fontdimen3\font minus
  \fontdimen4\font\relax}
\providecommand{\BIBforeignlanguage}[2]{{%
\expandafter\ifx\csname l@#1\endcsname\relax
\typeout{** WARNING: IEEEtran.bst: No hyphenation pattern has been}%
\typeout{** loaded for the language `#1'. Using the pattern for}%
\typeout{** the default language instead.}%
\else
\language=\csname l@#1\endcsname
\fi
#2}}
\providecommand{\BIBdecl}{\relax}
\BIBdecl

\bibitem{Artusi01}
A.~Artusi, F.~Banterle, and D.~Chetverikov., ``A survey of specularity removal
  methods,'' \emph{Computer Graphics Forum}, vol.~30, no.~8, pp. 2208--2230,
  2011.

\bibitem{Nayar02}
S.~Nayar, X.~Fang, and T.~Boult, ``Separation of reflection components using
  color and polarization,'' \emph{International Journal of Computer Vision},
  vol.~21, no.~3, pp. 163--186, 1997.

\bibitem{Wolff01}
L.~Wolff, ``Polarization-based material classification from specular
  reflection,'' \emph{IEEE Transactions on Pattern Analysis and Machine
  Intelligence}, vol.~12, no.~11, pp. 1059--1071, 1990.

\bibitem{Sato01}
Y.~Sato and K.~Ikeuchi, ``Temporal-color space analysis of reflection,''
  \emph{Journal of the Optical Society of America A}, vol.~11, no.~11, pp.
  2990--3002, 1994.

\bibitem{Lin02}
S.~Lin and H.~Shum, ``Separation of diffuse and specular reflection in color
  images,'' in \emph{Proceedings of International Conference on Computer Vision
  and Pattern Recognition}, 2001.

\bibitem{Lin01}
S.~Lin, Y.~Li, S.~Kang, X.~Tong, and H.~Shum, ``Diffuse-specular separation and
  depth recovery from image sequences,'' in \emph{Proceedings of European
  Conference on Computer Vision}, 2002.

\bibitem{Qingxiong01}
Q.~Yang, S.~Wang, N.~Ahuja, and R.~Yang, ``A uniform framework for estimating
  illumination chromaticity, correspondence, and specular reflection,''
  \emph{IEEE Transactions on Image Processing}, vol.~20, no.~1, pp. 53--63,
  2011.

\bibitem{Tan01}
R.~Tan, K.~Nishino, and K.~Ikeuchi, ``Illumination chromaticity estimation
  using inverse-intensity chromaticity space,'' in \emph{Proceedings of
  International Conference on Computer Vision and Pattern Recognition}, 2003.

\bibitem{Drew14}
M.~S. Drew, H.~R.~V. Joze, and G.~D. Finlayson, ``The zeta-image, illuminant
  estimation, and specularity manipulation,'' \emph{Computer Vision and Image
  Understanding}, vol. 127, pp. 1--13, 2014.

\bibitem{Bajcsy01}
R.~Bajcsy, S.~Lee, and A.~Leonardis, ``Detection of diffuse and specular
  interface reflections and inter-reflections by color image segmentation,''
  \emph{International Journal of Computer Vision}, vol.~17, no.~3, pp.
  241--272, 1996.

\bibitem{Klinker02}
G.~Klinker, S.~Shafer, and T.~Kanade, ``The measurement of highlights in color
  images,'' \emph{International Journal of Computer Vision}, vol.~2, no.~1, pp.
  7--32, 1988.

\bibitem{Pesal01}
P.~Koirala, P.~Pant, M.~Hauta-Kasari, and J.~Parkkinen, ``Highlight detection
  and removal from spectral image,'' \emph{Journal of the Optical Society of
  America A}, vol.~28, no.~11, pp. 2284--2291, 2011.

\bibitem{Tan03}
R.~Tan and K.~Ikeuchi, ``Separating reflection components of textured surfaces
  using a single image,'' \emph{IEEE Transactions on Pattern Analysis and
  Machine Intelligence}, vol.~27, no.~2, pp. 178--193, 2005.

\bibitem{Yang01}
Q.~Yang, S.~Wang, and N.~Ahuja, ``Real-time specular highlight removal using
  bilateral filtering,'' in \emph{Proceedings of European Conference on
  Computer Vision}, 2010.

\bibitem{Mallick01}
S.~Mallick, T.~Zickler, P.~Belhumeur, and D.~Kriegman, ``Specularity removal in
  images and videos: A {PDE} approach,'' in \emph{Proceedings of European
  Conference on Computer Vision}, 2006.

\bibitem{Mallick02}
S.~Mallick, T.~Zickler, D.~Kriegman, and P.~Belhumeur, ``Beyond lambert:
  reconstructing specular surfaces using color,'' in \emph{Proceedings of
  International Conference on Computer Vision and Pattern Recognition}, 2005.

\bibitem{Hyeongwoo01}
H.~Kim, H.~Jin, S.~Hadap, and I.~Kweon, ``Specular reflection separation using
  dark channel prior,'' in \emph{Proceedings of International Conference on
  Computer Vision and Pattern Recognition}, 2013.

\bibitem{Kaiming01}
K.~He, J.~Sun, and X.~Tang, ``Single image haze removal using dark channel
  prior,'' in \emph{Proceedings of International Conference on Computer Vision
  and Pattern Recognition}, 2009.

\bibitem{Akashi14}
Y.~Akashi and T.~Okatani, ``Separation of reflection components by sparse
  non-negative matrix factorization,'' in \emph{Proceedings of Asian Conference
  on Computer Vision}, 2014.

\bibitem{Tan02}
R.~Tan and K.~Ikeuchi, ``Reflection components decomposition of textured
  surfaces using linear basis functions,'' in \emph{Proceedings of
  International Conference on Computer Vision}, 2005.

\bibitem{Shen13}
H.~Shen and Z.~Zheng, ``Real-time highlight removal using intensity ratio,''
  \emph{Applied Optics}, vol.~52, no.~19, pp. 4483--4493, 2013.

\bibitem{Graham01}
G.~D. Finlayson and M.~S. Drew, ``4-sensor camera calibration for image
  representation invariant to shading, shadows, lighting, and specularities,''
  in \emph{Proceedings of International Conference on Computer Vision}, 2001.

\bibitem{Shafer01}
S.~A. Shafer, ``Using color to separate reflection components,'' \emph{Color
  research \& application}, vol.~10, no.~4, pp. 210--218, 1985.

\bibitem{Cong01}
C.~Huynh and A.~Robles-Kelly, ``A solution of the dichromatic model for
  multispectral photometric invariance,'' \emph{International Journal of
  Computer Vision}, vol.~90, no.~1, pp. 1--27, 2010.

\end{thebibliography}
\end{document}